\documentclass[preprint,3p,twocolumn]{elsarticle}

\usepackage{custom}
\usepackage{lineno,hyperref}
\usepackage{times}
\usepackage{epsfig}
\usepackage{graphicx}
\usepackage{amsmath}
\usepackage{amssymb}
\usepackage{algpseudocode,algorithm}
\usepackage{cases}
\usepackage{comment}
\usepackage[flushleft]{threeparttable}
\usepackage{booktabs}
\usepackage{multirow}
\usepackage{here}
\usepackage{custom_neural_networks}
\usepackage[font=normalsize]{caption}
\usepackage[normalsize]{subfigure}

\newcommand*{\eg}{\emph{e.g.,}}
\newcommand*{\ie}{\emph{i.e.,}}
\newcommand*{\etal}{\emph{et al.}}

\newcommand*{\id}{InD}
\newcommand*{\ood}{OoD}

\usepackage{subfigure}

\usepackage{amsmath,amsfonts,bm}

\def\eqref#1{equation~\ref{#1}}

\def\1{\bm{1}}

\def\rvx{{\mathbf{x}}}
\def\rvy{{\mathbf{y}}}

\def\vtheta{{\bm{\theta}}}

\def\vf{{\bm{f}}}
\def\vg{{\bm{g}}}

\def\vp{{\bm{p}}}

\def\vv{{\bm{v}}}

\DeclareMathAlphabet{\mathsfit}{\encodingdefault}{\sfdefault}{m}{sl}
\SetMathAlphabet{\mathsfit}{bold}{\encodingdefault}{\sfdefault}{bx}{n}

\def\gC{{\mathcal{C}}}
\def\gD{{\mathcal{D}}}

\def\gI{{\mathcal{I}}}

\def\gN{{\mathcal{N}}}

\date{}
\journal{Neural Networks}
\begin{document}

\begin{frontmatter}

\title{Three approaches to facilitate DNN generalization to objects in \\ out-of-distribution orientations and illuminations}
\author[1]{Akira Sakai\corref{cor1}}
\ead{akira.sakai@fujitsu.com}
\author[1]{Taro Sunagawa} \author[2,4]{Spandan Madan} \author[1]{Kanata Suzuki} \author[1]{Takashi Katoh} \author[1]{Hiromichi Kobashi} \author[2]{\\Hanspeter Pfister} \author[3]{Pawan Sinha} \author[3,4]{Xavier Boix\corref{cor1}\fnref{fn1}} \ead{xboix@mit.edu}
\author[1,4]{Tomotake Sasaki\corref{cor1}\fnref{fn1}} \ead{tomotake.sasaki@fujitsu.com}

\cortext[cor1]{Corresponding author}
\fntext[fn1]{Equal contribution}
\address[1]{Artificial Intelligence Laboratory, Fujitsu Limited, 4-1-1 Kamikodanaka, Nakahara-Ku, Kawasaki, Kanagawa 211-8588, Japan.}
\address[2]{School of Engineering and Applied Sciences, Harvard University, 29 Oxford Street, Cambridge, MA 02138, USA}
\address[3]{Department of Brain and Cognitive Sciences, Massachusetts Institute of Technology, 77 Massachusetts Avenue, Cambridge, MA 02139, USA.}
\address[4]{Center for Brains, Minds and Machines, 77 Massachusetts Avenue, Cambridge, MA 02139, USA.}

\begin{abstract}

The training data distribution is often biased towards objects in certain orientations and illumination conditions. While humans have a remarkable capability of recognizing objects in out-of-distribution (OoD) orientations and illuminations, Deep Neural Networks (DNNs) severely suffer in this case, even when large amounts of training examples are available. In this paper, we investigate three different approaches to improve DNNs in recognizing objects in OoD orientations and illuminations. Namely, these are (i) training much longer after convergence of the in-distribution (InD) validation accuracy, i.e., late-stopping, (ii) tuning the momentum parameter of the batch normalization layers, and (iii) enforcing invariance of the neural activity in an intermediate layer to orientation and illumination conditions. Each of these approaches substantially improves the DNN's OoD accuracy (more than 20\% in some cases). We report results in four datasets: two datasets are modified from the MNIST and iLab datasets, and the other two are novel (one of 3D rendered cars and another of objects taken from various controlled orientations and illumination conditions). These datasets allow to study the effects of different amounts of bias and are challenging as DNNs perform poorly in OoD conditions. Finally, we demonstrate that even though the three approaches focus on different aspects of DNNs, they all tend to lead to the same underlying neural mechanism to enable OoD accuracy gains -- individual neurons in the intermediate layers become more selective to a category and also invariant to OoD orientations and illuminations. We anticipate this study to be a basis for further improvement of deep neural networks’ \ood~generalization performance, which is highly demanded to achieve safe and fair AI applications.

\end{abstract}

\begin{keyword}
Out-of-distribution Generalization; Object Recognition in Novel Conditions; Neural Invariance;Neural Selectivity; Neural Activity Analysis
\end{keyword}
\end{frontmatter}

\section{Introduction}

The object recognition performance of Deep Neural Networks (DNNs) dramatically degrades when the train and test distributions are not identical due to dataset bias~\cite{torralba2011unbiased}, \ie~when tested in out-of-distribution (\ood) conditions.
There is a big gap between DNNs and humans when evaluated in \ood~conditions. This issue has been getting much interest in recent years~\cite{geirhos2018imagenettrained,BeeryHP18,hendrycks2020many,recht2018cifar10.1,recht2019imagenet}, as it severely compromises the safety and fairness of AI applications.

One of the most prominent factors of dataset bias is that objects may appear in a constrained range of orientation and illumination conditions~\cite{alcorn2019strike,barbu2019objectnet}. While generalization to \ood~orientations and illumination conditions has been long studied in both biological and artificial neural networks,~\eg~\cite{sinha1996role,ullman1996high,Anselmi2016invarianceselectivity}, the computational mechanisms that facilitate such generalization remain as a key outstanding question. Recently,~\cite{zaidi2020robustness,madan2020capability} have shown that  DNNs are capable to overcome bias by transferring the generalization ability obtained from objects seen in a richer set of conditions to the objects seen in biased conditions. Also, the emergence of representations at the individual neuron level in the intermediate layers of the DNN that are selective to categories and invariant to the \ood~conditions  has been identified as a mechanism that may facilitate such \ood~generalization. Invariant neural representations have been studied during decades,~\eg~\cite{Anselmi2016invarianceselectivity}, and here they appear as the mechanism that allows \ood~generalization. This begs the question whether we can further encourage the emergence of invariant neural representations in DNNs in order to further improve \ood~generalization.

\begin{figure*}[pht]
\centering
    \begin{tabular}{cc}
     \normalsize {\bf Late-stopping} & \normalsize {\bf Tuning batch normalization} \\[-0.5em]
        \subfigure[]{\label{fig:daiso_late-stopping}
        \includegraphics[width=0.45\textwidth]{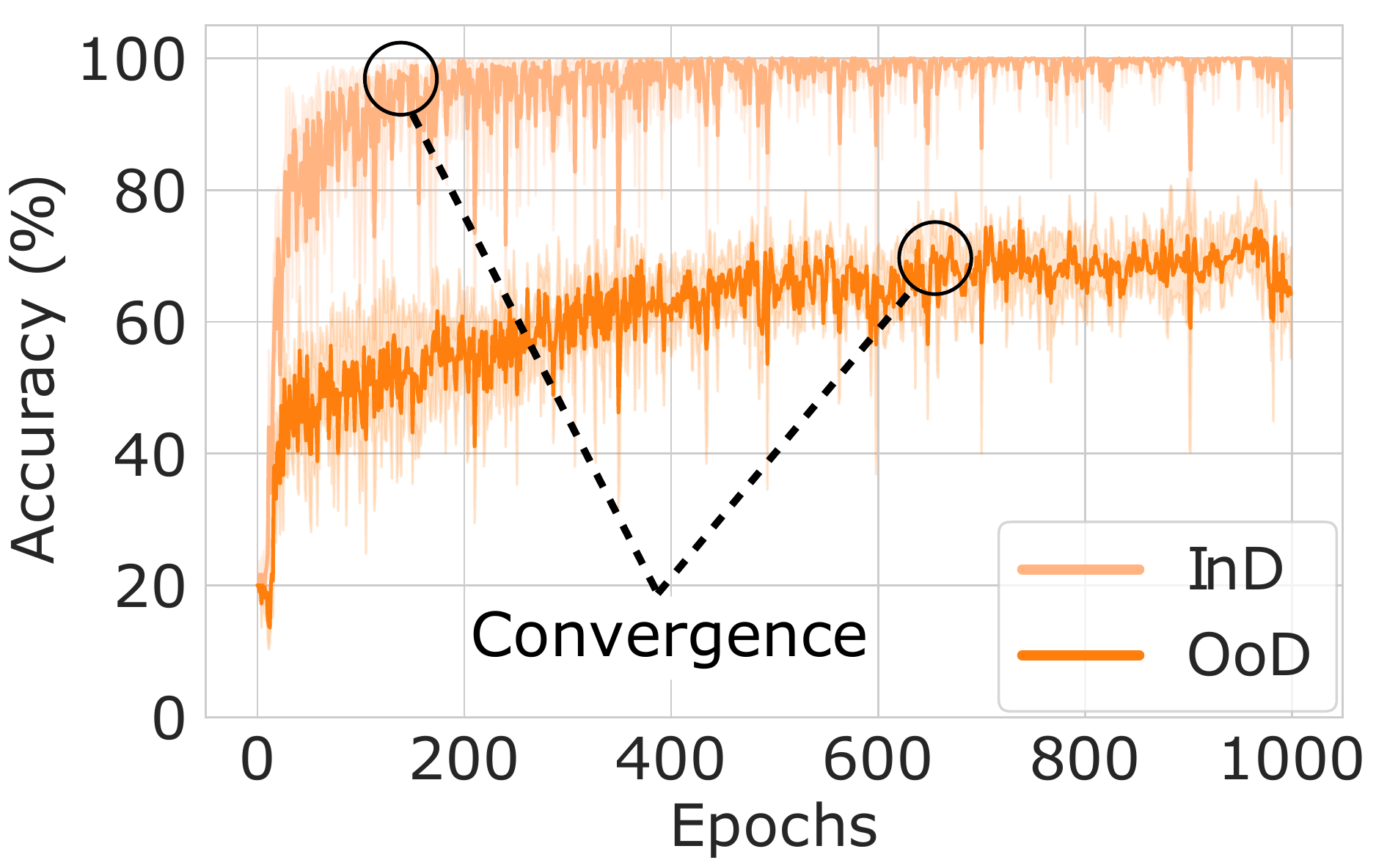}} &
         \subfigure[]{\label{fig:carscg_bnmomentum}
        \includegraphics[width=0.45\textwidth]{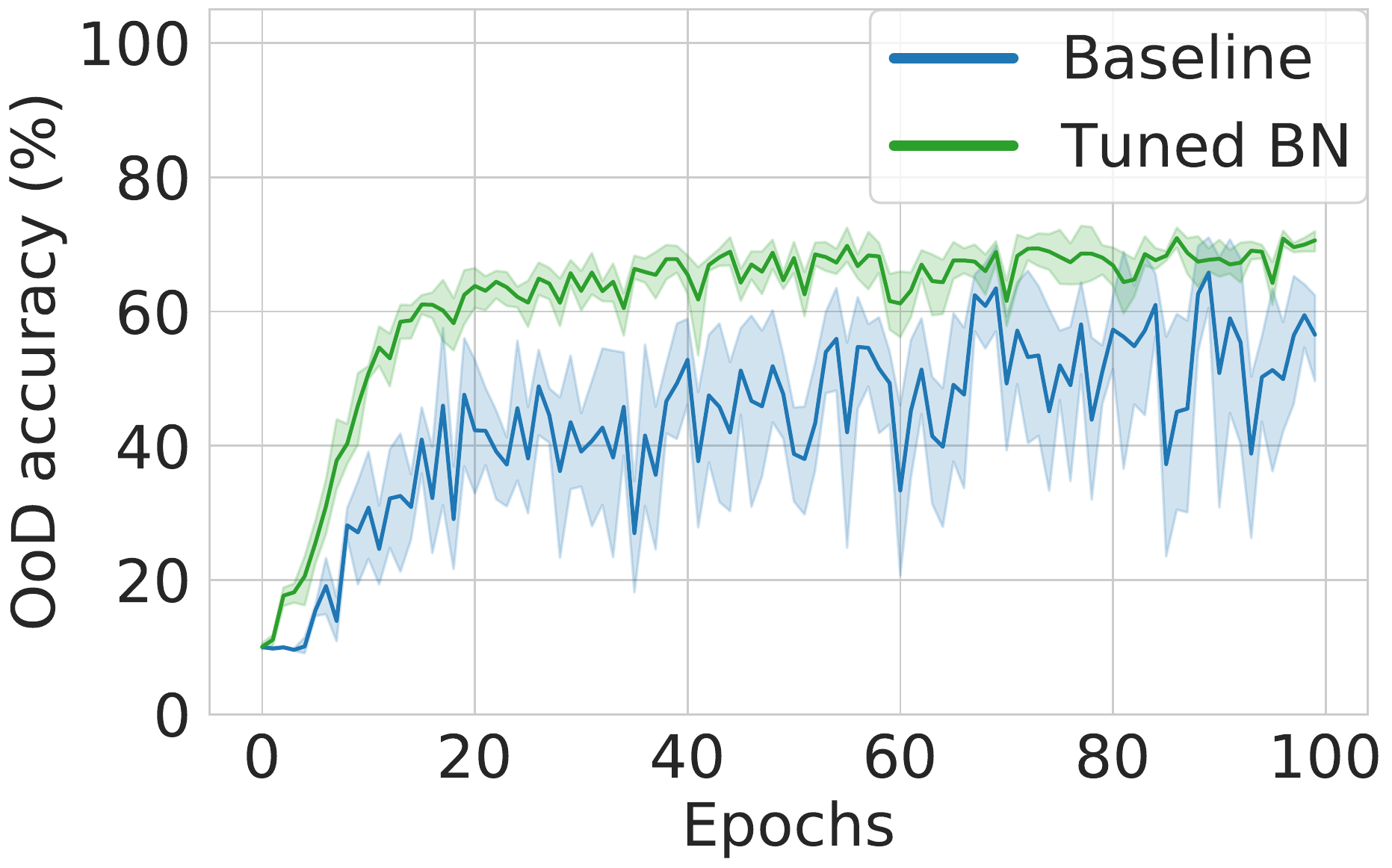}}\\
    \end{tabular}\\
   {\bf  Invariance loss}\\
    \subfigure[]{\label{fig:algo}
    \includegraphics[width=0.99\textwidth]{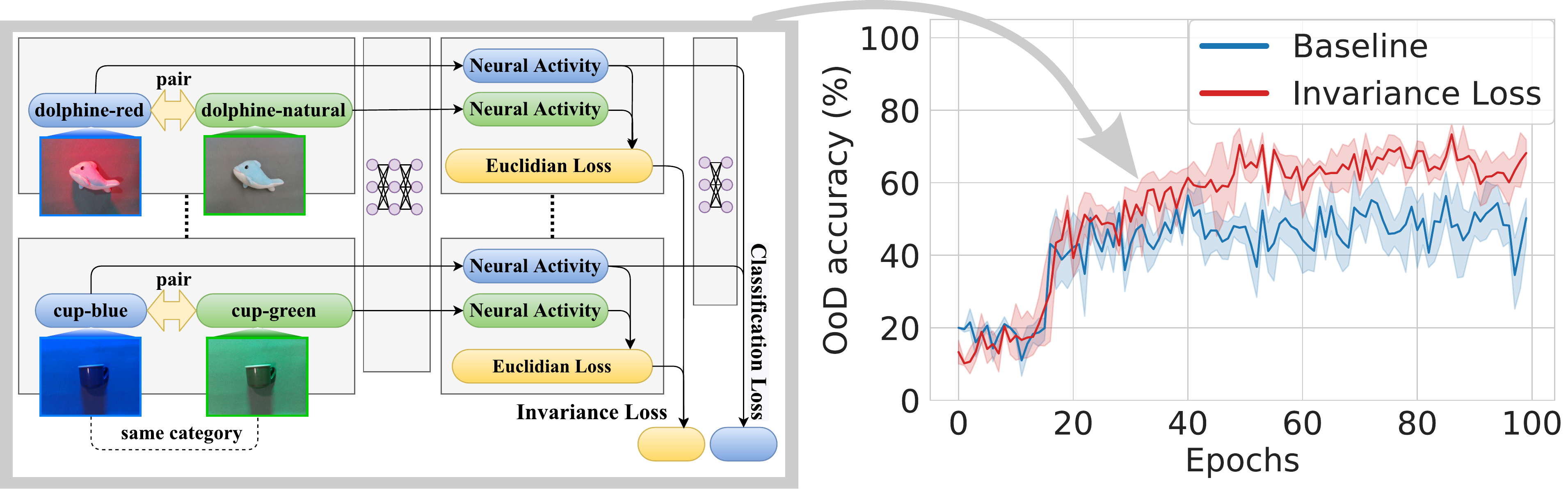}}
\caption{
\emph{Three approaches to facilitate generalization to objects in out-of-distribution (\ood) orientations and illuminations.} (a) Learning curves of in-distribution (\id) test accuracy and \ood~accuracy for late-stopping applied to the MiscGoods-illuminations dataset (medium \id~data diversity). 
OoD accuracy converges much later than InD accuracy.  (b) Learning curves of the OoD accuracy with and without tuning batch normalization momentum (tuned BN) in the CarsCG-Orientations, dataset (medium \id~data diversity).
It can be seen that tuning the momentum reduces the oscillation of the \ood~accuracy and improves the performance. 
(c) Left: Conceptual diagram of the invariance loss. Pairs of images that belong to the same  category are fed into the DNN. The invariance loss is based on the Euclidean distance between the pairs of the last ReLU activity. The classification loss is calculated with the network output as usual. The total loss is the weighted sum of the invariance and classification losses.
    Right: Learning curve of \ood~accuracy in MiscGoods-Illuminations dataset (medium \id~data diversity) when the invariance loss is applied. The \ood~accuracy increases by about $20\%$ compared to the baseline. The solid lines in the plots are the mean value.  The lighter semitransparent colors surrounding the the solid lines indicate $95\%$ confidence interval.   }
\label{fig:three_approaches}
\end{figure*}

In this paper, we investigate factors that can substantially boost the DNN ability to recognize objects in \ood~orientations and illuminations. In particular, we discover that the following factors, summarized in Fig.~\ref{fig:three_approaches}, have a remarkable impact:
\begin{enumerate}
    \item \emph{Late-stopping:} DNNs are usually trained until the validation recognition accuracy (which is in-distribution) converges. We found that in many cases the \ood~recognition accuracy improves slowly, yet consistently, after the validation (in-distribution) accuracy has converged. This finding is surprising as classic machine learning theory suggests early-stopping as a regularization mechanism~\cite{Yao07onearly}, and we found that the opposite is beneficial to improve \ood~generalization in DNNs. We call this approach ``late-stopping''.
    \item \emph{Tuning the batch normalization parameter:} Batch normalization (BN) is known to have an impact in \ood~recognition accuracy~\cite{SchneiderRE0BB20}. We found that tuning the only hyperparameter of BN, \ie~the momentum, yields substantial gains of \ood~recognition accuracy. This approach is denoted as ``tuned BN''.
    \item \emph{Neural activity invariance loss:} Motivated by the aforementioned finding in previous works that invariant neural representations leads to improvements of  the \ood~recognition accuracy, we include an additional term in the loss function to encourage this phenomenon. This loss term takes the Euclidean distance between neural activity corresponding to pairs of images from the same category on an intermediate layer. By minimizing this loss term, the neural activity tends to be invariant for objects of the same category even in different viewing conditions. We do not consider that pairs of images from different categories should have distinguishable neural activity, since the classification loss term already encourages this. We call this approach ``invariance loss'' in short.
\end{enumerate}

Our results demonstrate that each of these three approaches alone leads to substantial improvements of  object recognition in \ood~orientations and illumination conditions. Results also corroborate that when any of the three  approaches leads to an increase of selectivity and invariance at the individual neuron level,  \ood~recognition accuracy improves in the majority of trials. 
Experiments are performed in four challenging benchmarks, namely modifications of the MNIST dataset~\cite{Lecun1998mnist} and iLab dataset~\cite{Borji2016iLab} and two novel datasets we introduce, which are the CarsCG and the MiscGoods datasets.  CarsCG contains 3D rendered cars from different orientations, and the MiscGoods dataset consists of images of objects taken with a robotic arm from different viewpoints and controlled illumination conditions. These datasets allow to evaluate the DNN generalization ability to recognize objects in \ood~orientations and illumination conditions. Also, they allow to analyze the effects of different amounts of bias and are challenging as DNNs perform poorly in  \ood~conditions.

\begin{figure*}[t]

    \begin{tabular}{@{\hspace{-0.1cm}}cc}
    MNIST-Positions & iLab-Orientations\\[-0.5em]
        \subfigure[]{\label{fig:mnist} \includegraphics[width=0.47\textwidth]{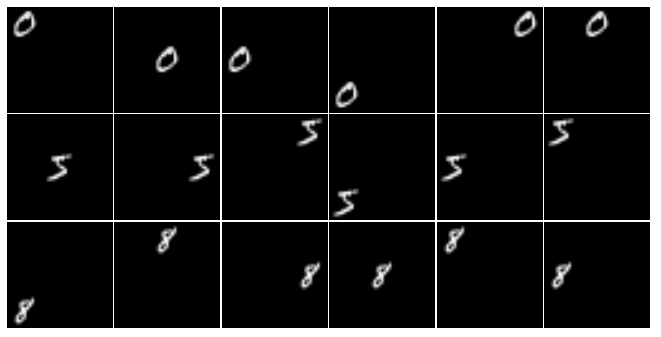}} &
        \subfigure[]{\label{fig:ilab} 
           \includegraphics[width=0.47\textwidth]{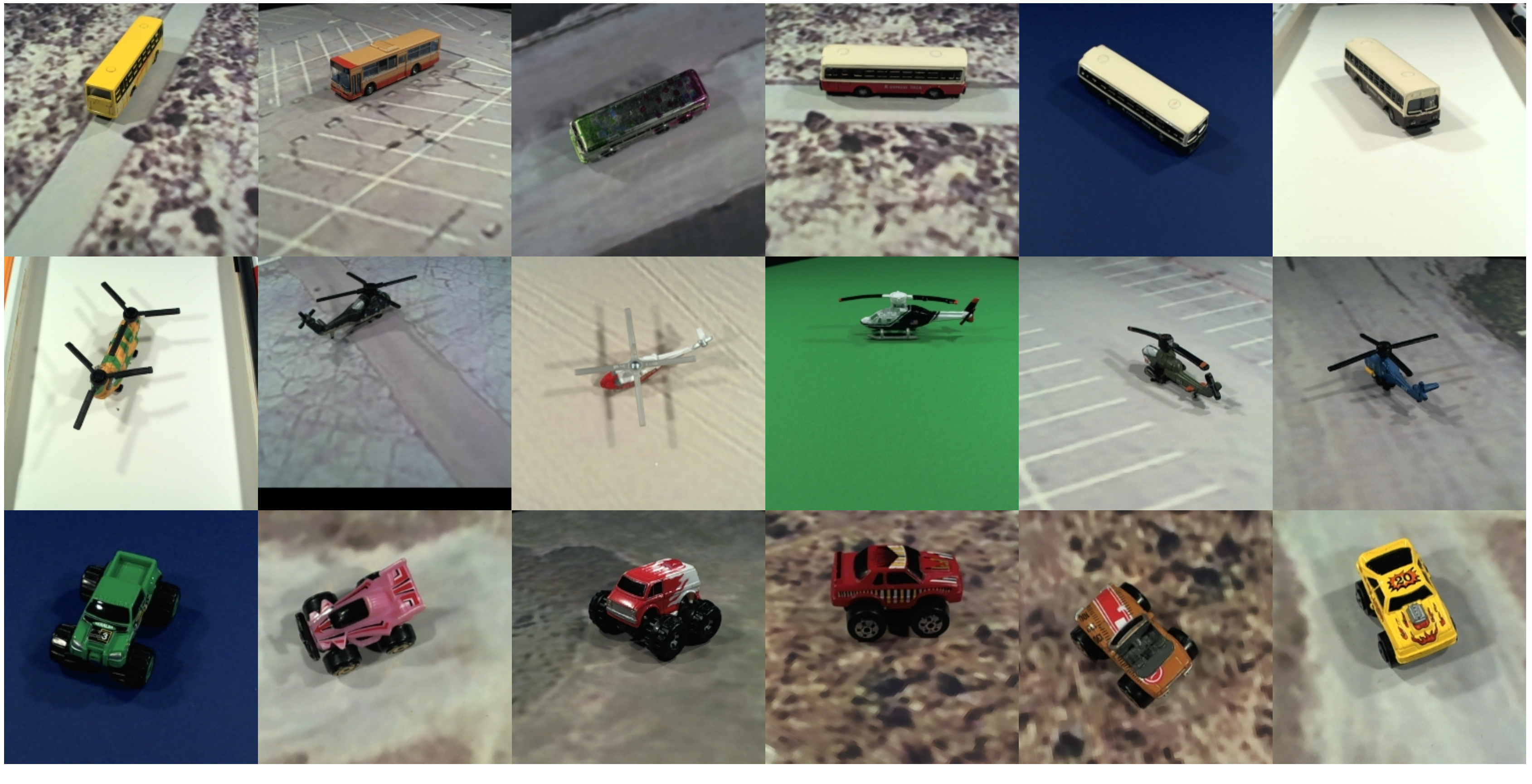}}\\
           CarsCG-Orientations & MiscGoods-Illuminations\\[-0.5em]
        \subfigure[]{\label{fig:carcgs}
           \includegraphics*[width=0.47\textwidth]{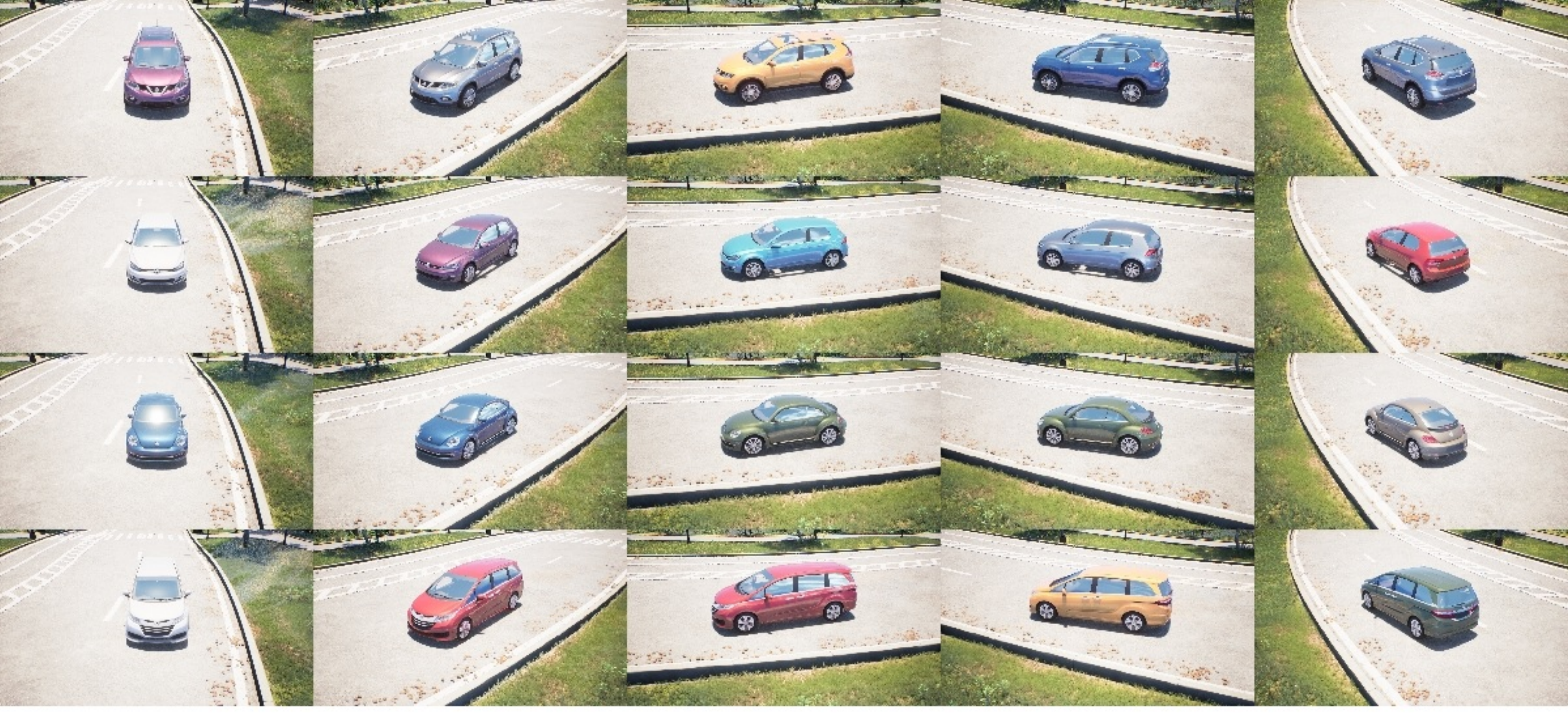}} &
        \subfigure[]{\label{fig:daiso}
           \includegraphics*[width=0.47\textwidth]{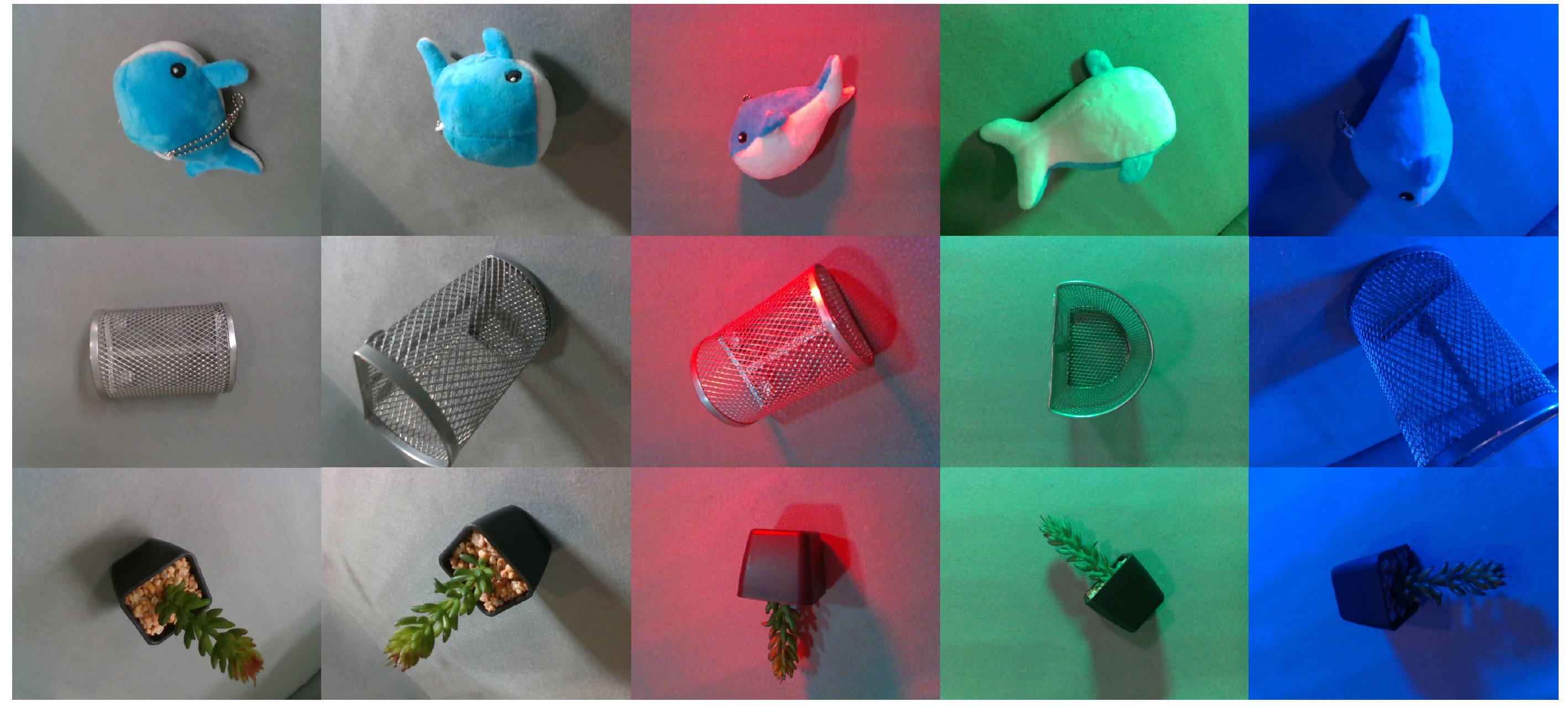}}
    \end{tabular}
    \caption{\emph{Sample images from four datasets.} (a)~MNIST-Positions, (b)~iLab-Orientations , (c)~CarsCG-Orientations, and (d)~MiscGoods-Illuminations are shown in each subfigure. Samples from each dataset are arranged in a grid pattern. Each row indicates categories and each column indicates either an orientation or an illumination condition.}
\end{figure*}

\begin{figure}[t]
\begin{center}
\includegraphics[width=0.47\textwidth]{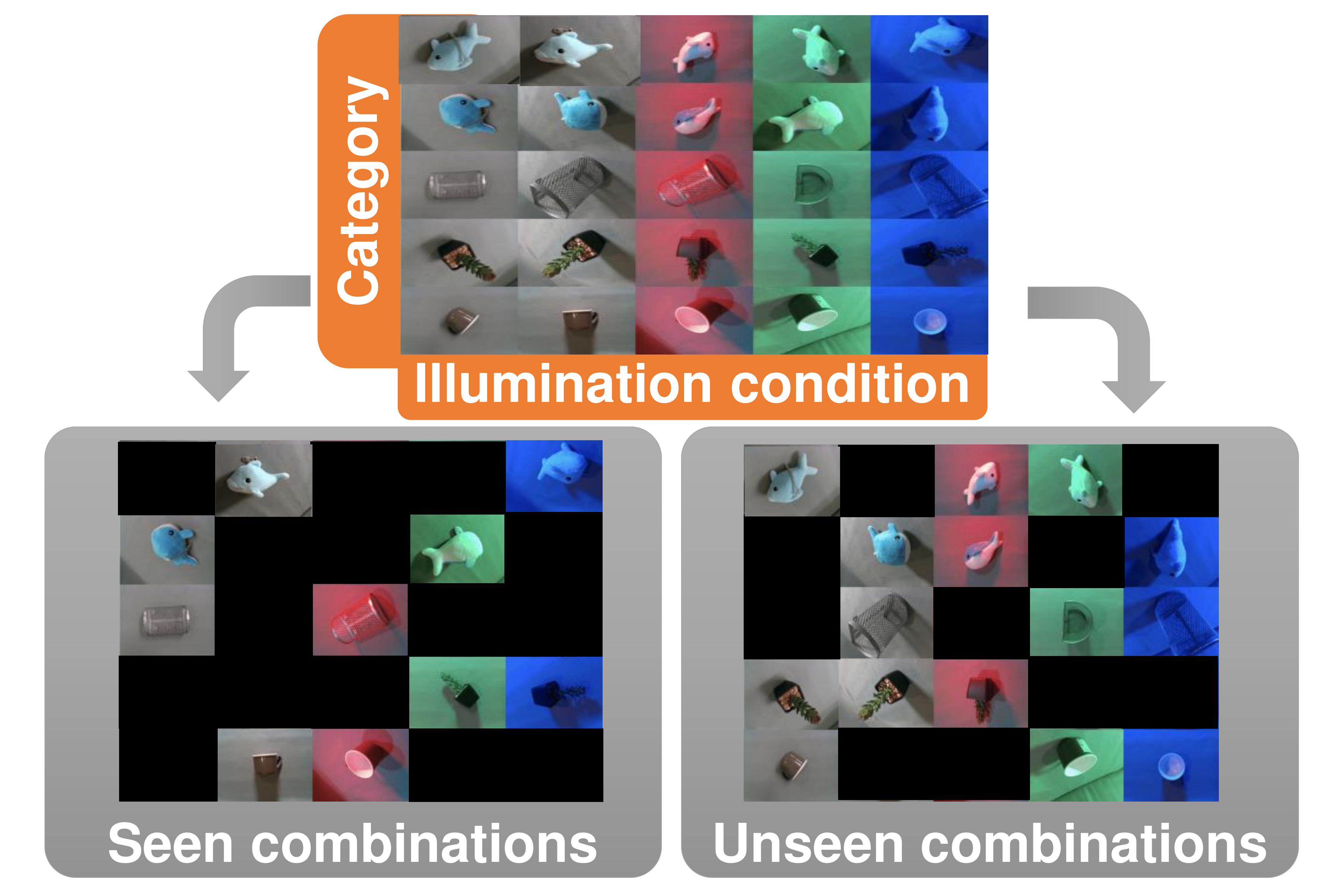}
\end{center}
\caption{\emph{InD and OoD combinations for bias-controlled experiments.} 
Each sample is a combination of a category  and an orientation or illumination condition. We create a set of  combinations called ``InD combinations'' and a set of  combinations called ``OoD combinations''. 
The ratio of InD combinations to all combinations is called InD data diversity.
In addition, we create a train dataset ($\gD^{(\rm{InD)}}_{\rm train}$) and an InD validation dataset ($\gD^{(\rm{InD})}_{\rm val}$) from samples included in the InD combinations, and an OoD test dataset ($\gD^{(\rm{OoD})}$)  from the samples included in the OoD combinations.
}
\label{fig_setting}
\end{figure}

\begin{figure*}[t]
    \begin{tabular}{c|c|c}
    Low InD data diversity & $\scriptstyle \mbox{Medium InD data diversity}$ &$\scriptstyle \mbox{High InD data diversity}$\\[-0.6em]
            \subfigure{
            \includegraphics[width=0.25\textwidth]{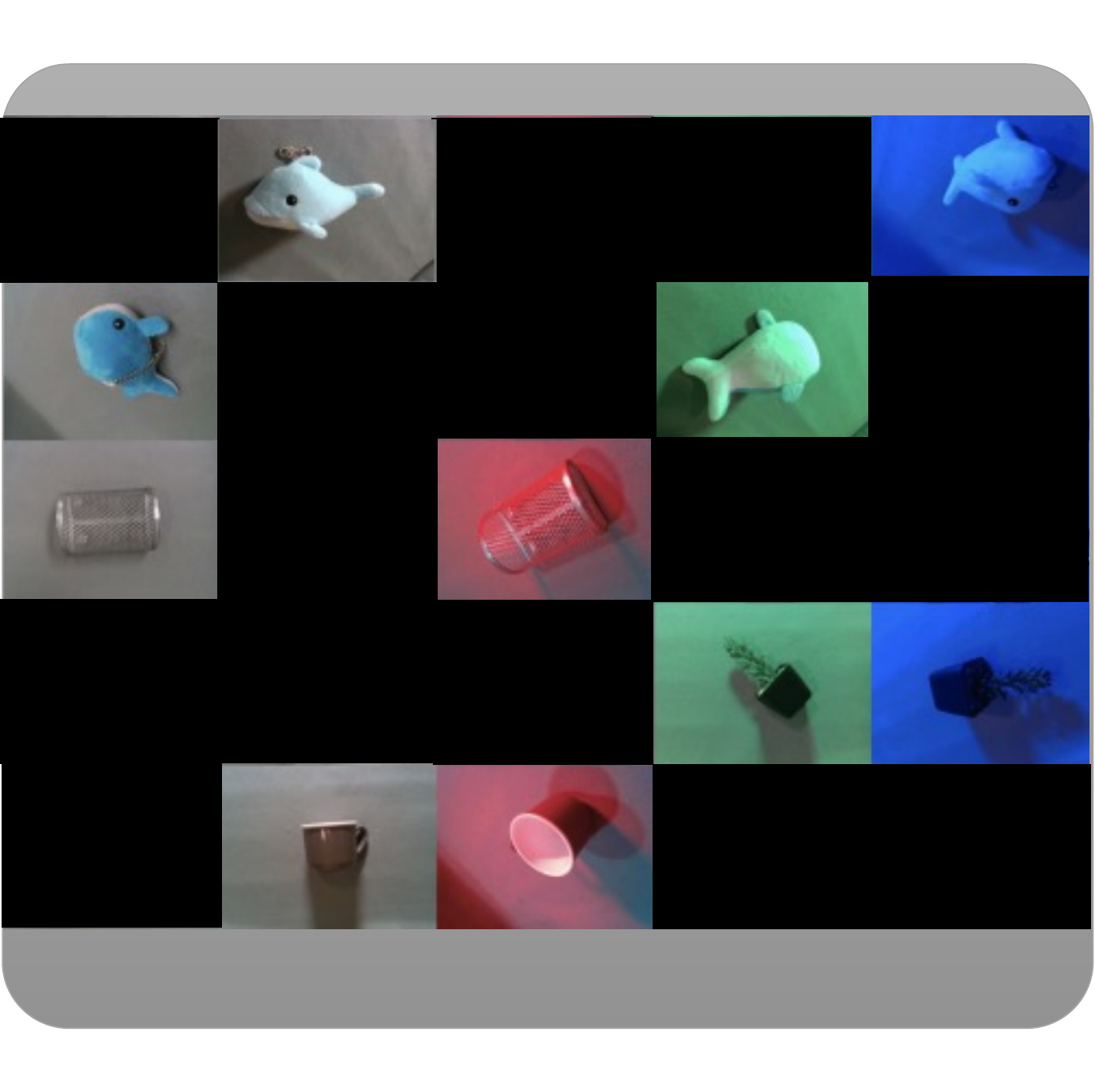}}&
        \subfigure{{
            \includegraphics[width=0.25\textwidth]{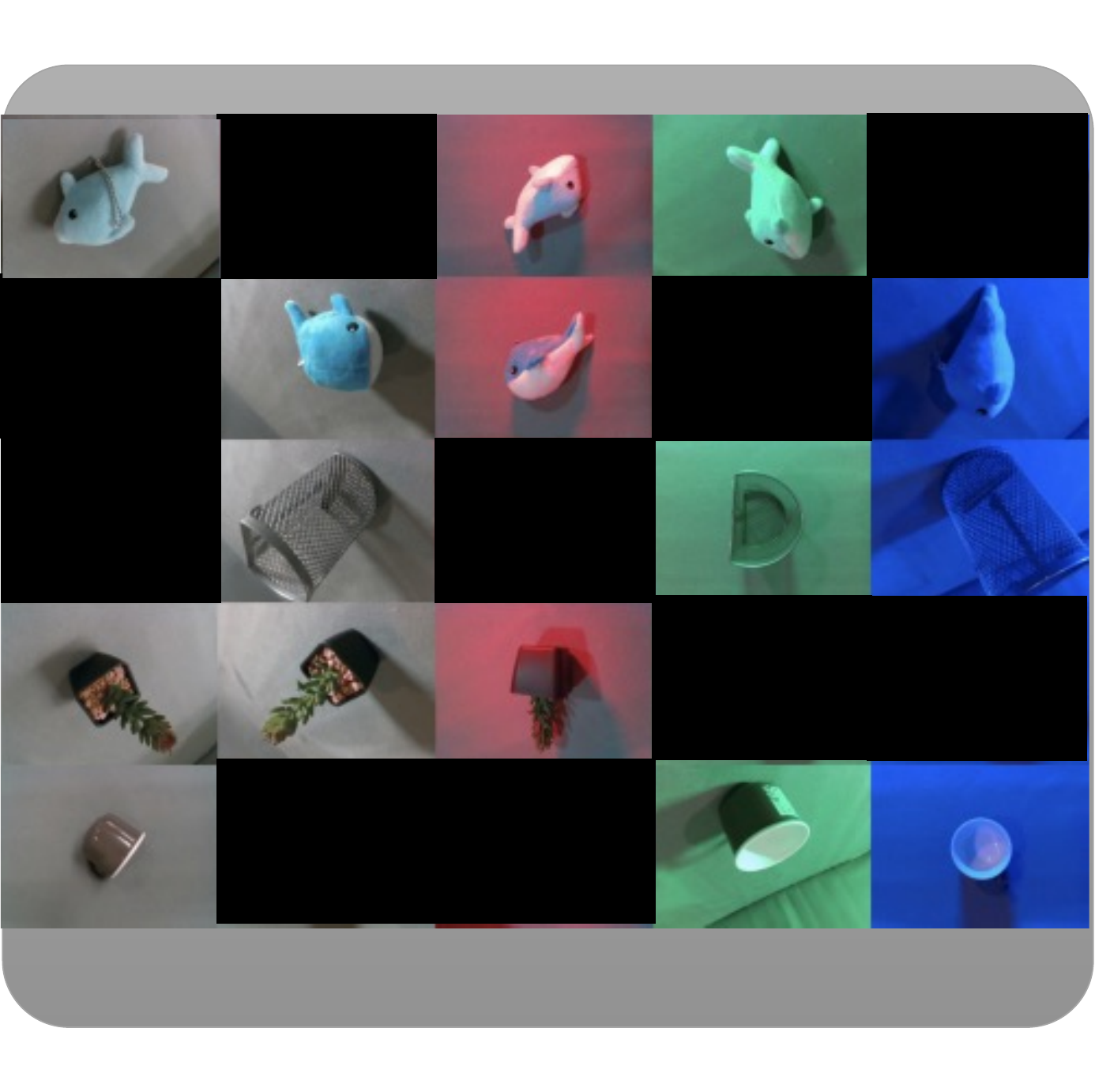}}}&
        \subfigure{
            \includegraphics*[width=0.25\textwidth]{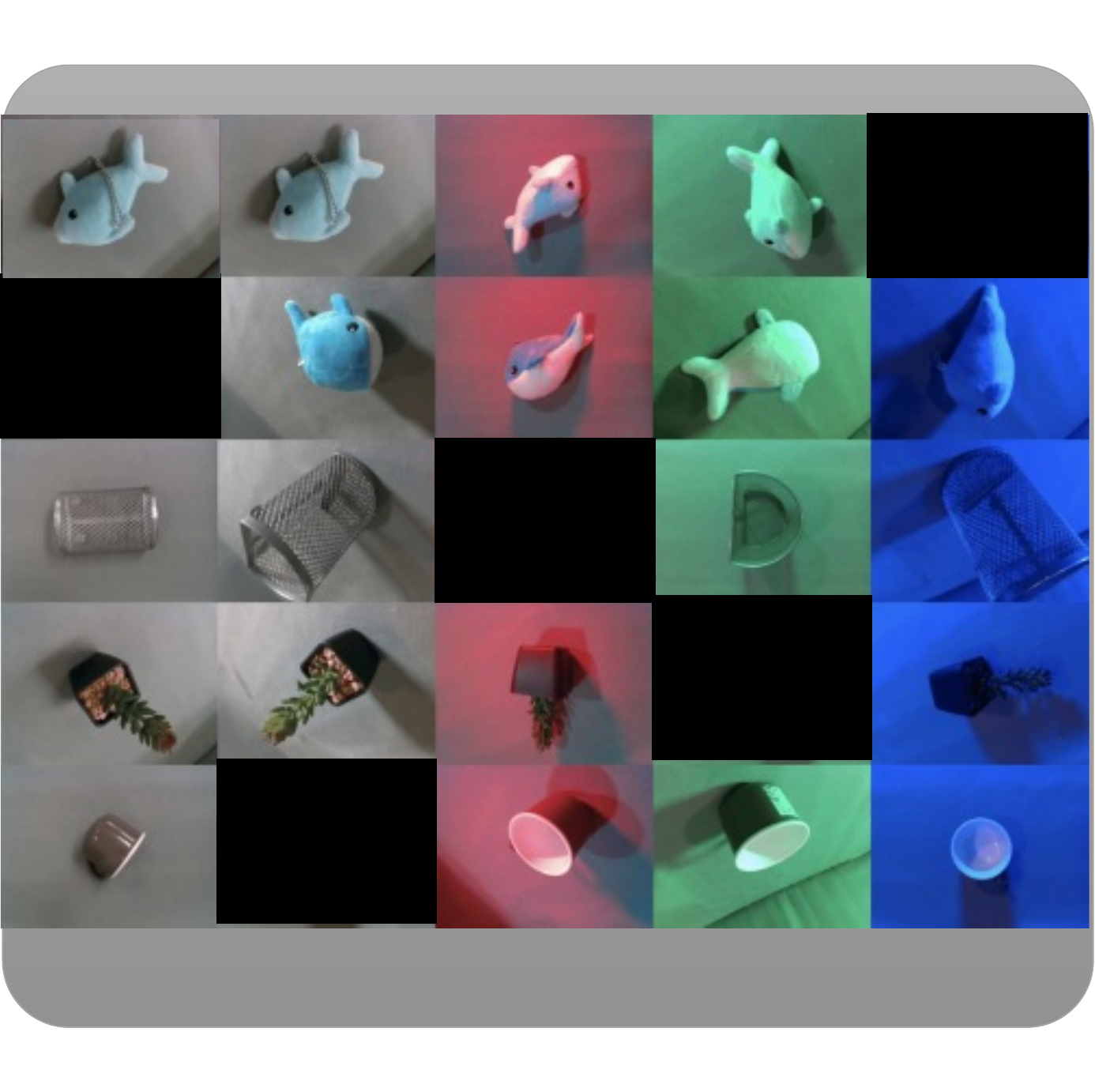}} \\
        \setcounter{subfigure}{0}
        \subfigure[]{\label{fig:vanilla_degradate_low}
            \includegraphics[width=0.3\textwidth]{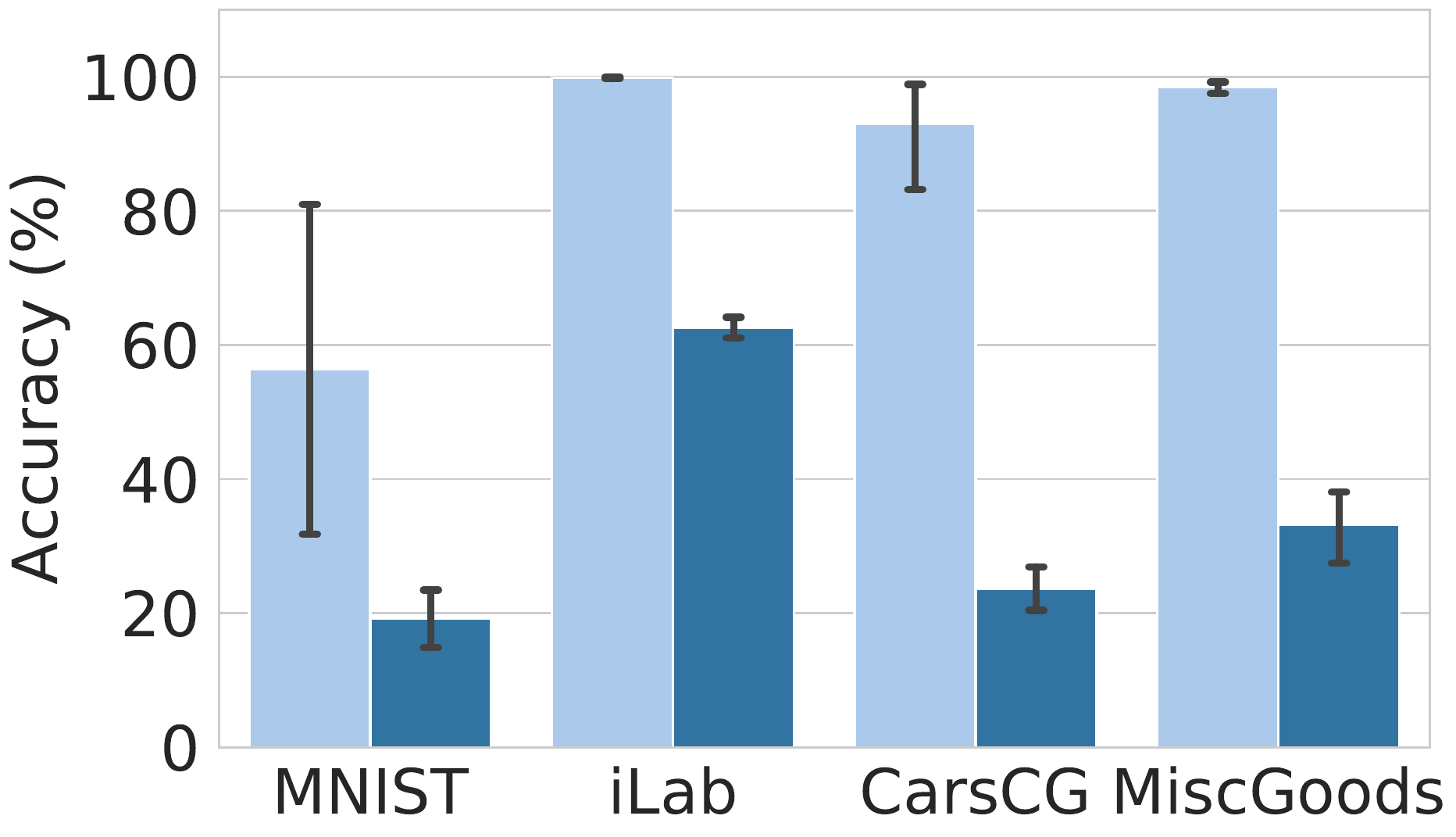}}&
        \subfigure[]{\label{fig:vanilla_degradate_medium}
            \includegraphics[width=0.3\textwidth]{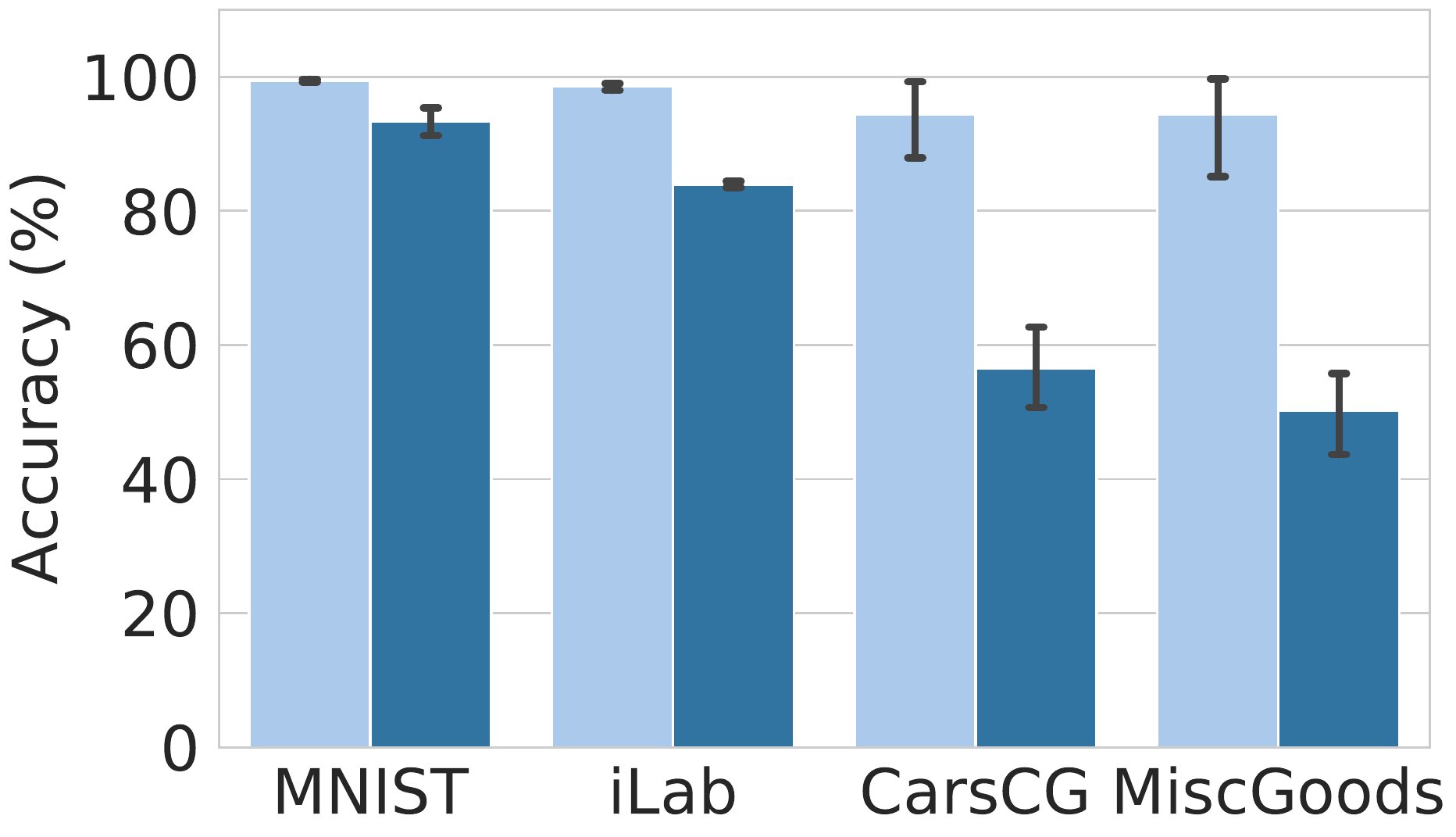}}&
        \subfigure[]{\label{fig:vanilla_degradate_high}
            \includegraphics*[width=0.3\textwidth]{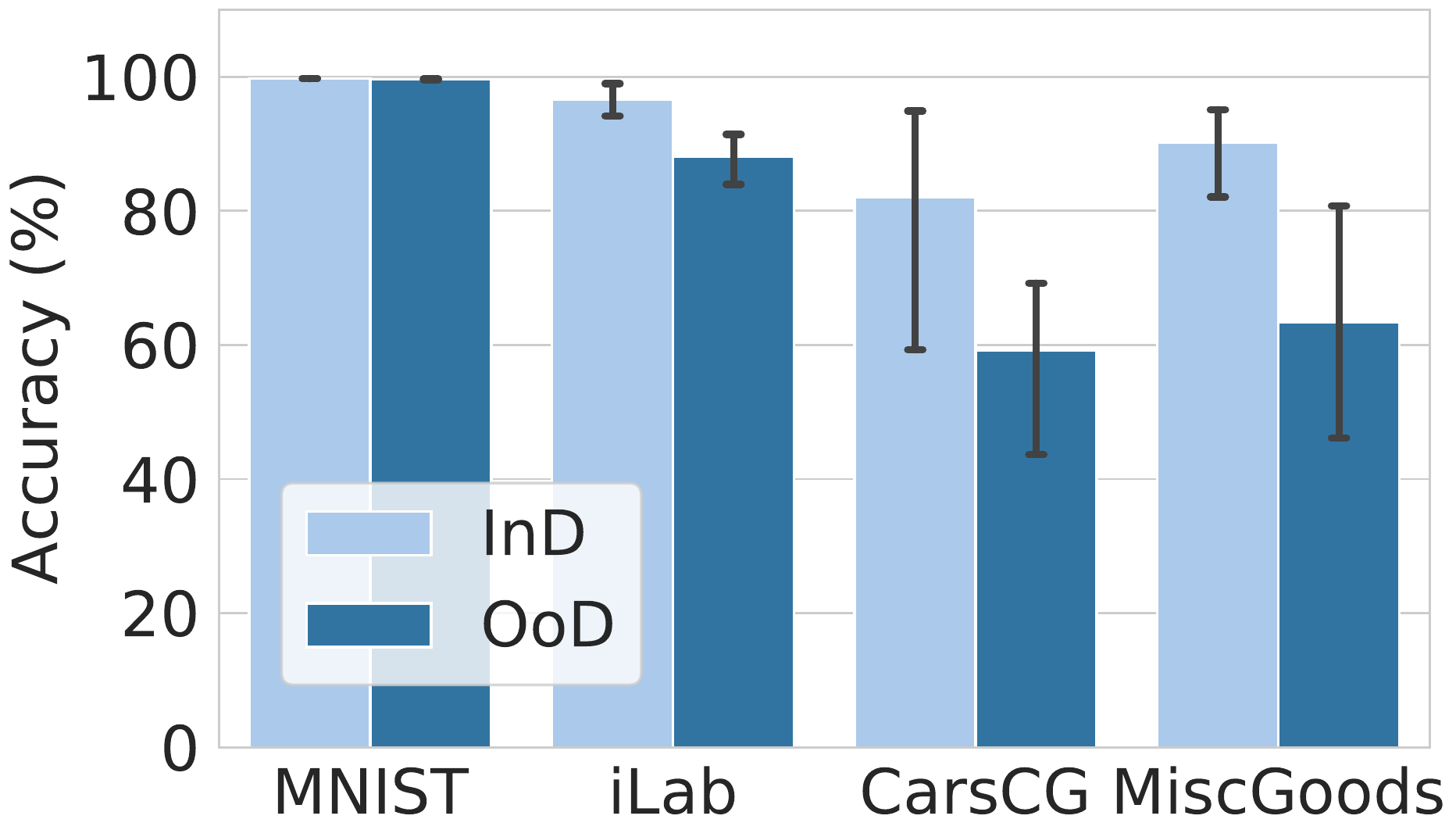}} \\

    \end{tabular}
\caption{\emph{Performance degradation in the \ood~conditions.}  Upper figures show the examples of \id~combination in MiscGoods-Illuminations dataset to depict the \id~data diversity of each experiment. Lower figures show InD or OoD accuracy of ResNet-18 in (a)~low \id~data diversity, (b)~medium \id~data diversity and (c)~high \id~data diversity performed on four datasets. 
Each experiment is conducted five times, and the mean and $95\%$ confidence interval are reported.
Sharp performance degradation in \ood~accuracy is observed (\eg~between $40\%$ to almost $80\%$ is observed when the \id~data diversity is low). 
These result shows the impact of a distribution shift from InD to OoD to the performance of a DNN.
}
\label{fig:performance_degrade}
\end{figure*}

\section{Previous works}

Our results add to the growing body of literature to improve the generalization ability of DNNs to \ood~orientations and illumination conditions. Prior efforts leverage synthesized sources of training data~\cite{cubuk2019randaugment,Halder2019rainAugment,Kim2021LocalAug,qiao2020learning}, 3D models of objects~\cite{wang2020NeMo}, specific characteristics of the target domain~\cite{Chides2019RotationRobust,Qi2018Concentricpooling,Sabour2017capsule}, or  sensing approaches such as omnidirectional imaging~\cite{cohen2018spherical}. These approaches add preconceived components to the DNN that need to be  adjusted at hand for new objects and conditions. Here, we focus on pure learning-based strategies as these are not constrained to specific objects and conditions and can be automatically adjusted to new datasets. 

Other strands of research that live in neighbouring areas investigate generalization to new domains and also, overcoming spurious correlations between image features and categories. While domain generalization does not tackle dataset bias and overcoming spurious correlations does not address recognition of objects in \ood~orientations and illuminations, these two research areas use related techniques and concepts to our work. In the following we review both of them.

\paragraph{Domain generalization}
There is a plethora of works that consists on learning representations in several domains that can be easily transferred to new domains,~\eg~\cite{carlucci2019domain,dou2019domain,Ghifary2015DomainGF,guo2020learning,Jia2020SSDG,Li2018DBLP,Li2018DGAdversarial,Volpi2018unseendomains}. The problem of domain generalization is similar to the problem overcoming dataset bias in our study in the sense that representations that facilitate generalization to novel conditions should be learned. However, in domain generalization the learner has access to multiple domains during training that can be leveraged for generalization, while in the problem of overcoming dataset bias only one training set is available. Recently, several works in domain generalization~\cite{Chatto2020DMG,ilse2020diva,rame2021ishr,xiao2021a} highlighted the need of invariant representations to obtain further improvements in generalization, which further motivates investigating the invariance loss in our study.

\paragraph{Overcoming spurious correlations between image features and categories}
Many datasets are biased in a way that a specific image feature consistently appears in images of the same category. DNNs tend to learn that those features are informative of the  category~\cite{geirhos2020shortcut}. This form of dataset bias is different from the bias in the object orientation and illumination conditions, which do not necessarily lead to spurious correlations.
Recently, there have been several works that address spurious correlations. These are based on automatically detecting the features that spuriously correlate with the category, and encourage the DNN not to rely on those features~\cite{Sagawa*2020Distributionally,arjovsky2019invariant}.  Ahmed~\emph{et al.}~\cite{ahmed2021systematic} introduced a method that effectively alleviates the effect of spurious correlation caused by biased object background. This work exploits the assumption that the training distribution also contains examples without spurious correlations. 
It employs EIIL~\cite{creager2021environment} to classify the images of an  category with the features that spuriously correlate with the  category and without them. Then, invariance is encouraged across these two groups of images. Thus, invariance appears once more as a facilitator of generalization.

\section{Performance degradation on \ood~Conditions}

In this section, we introduce the methodology to evaluate the accuracy of the DNN in \ood~conditions. First, we describe the procedure of the bias-controlled experiment. Next, we introduce the four datasets used in this study and finally, we evaluate the performance degradation that occurs in \ood~conditions in these four datasets.

\subsection{Bias-controlled experiments}\label{subsec:Creating spurious correlation}

In a dataset there could be multiple biasing factors at the same time that can cause performance degradation. In the datasets in this study, we analyze either the orientation or illumination condition, as it allows to more clearly understand the effect of each individual factor. 
Thus, the datasets that we use contains several combinations of  categories and conditions. We use $\gC$ to denote the set of all  categories and $\gN$ the set of all  orientation or illuminations conditions. 
Let $ \rvx^{(k)}$ be an image of the dataset and let $\rvy^{(k)} := (c^{(k)}, n^{(k)})$ be a tuple representing the  ground-truth  category (\ie~$c^{(k)}\in \gC$), and the orientation or illuminations condition (\ie~$n^{(k)}\in \gN$).

In order to evaluate the DNN's \ood~generalization capabilities, we train them in a dataset that follows a distribution that only contains a subset of all possible combinations,~\ie~a subset of $\gC \times \gN$. Then, the DNN is evaluated with images from combinations that were not included in the training distribution. Let $\gI \subset \gC \times \gN$ be the set of combinations used to generate the \id~combinations. We ensure that $\gI$ contains all categories and all conditions at least once (but not all combinations), such that we have images from all image categories and conditions in a balanced manner.

We use $\gD^{(\rm InD)}$ to denote the set of images that are \id,~\ie~images whose label is in $\gI$, $\rvy^{(k)}\in \gI$. Namely, the \id~images dataset, $\gD^{(\rm InD)}$, is defined as in the following: 
\begin{equation}
\gD^{(\rm InD)} := \{(\rvx,\rvy)| \rvy \in \gI \}.
\end{equation}
 $\gD^{(\rm InD)}$ is further divided into train dataset and validation dataset, which we denote as $\gD^{\rm{(InD)}}_{\rm train}$ and $\gD^{(\rm InD)}_{\rm val}$, respectively.
The term \id~accuracy refers to the DNN's accuracy on $\gD^{(\rm InD)}_{\rm val}$. 
 The \ood~dataset $\gD^{(\rm OoD)}$ is defined as 
\begin{equation} 
\gD^{(\rm OoD)} := \{(\rvx,\rvy)| \rvy \in (\gC\times\gN)  \setminus \gI \}.
\end{equation}
The term OoD accuracy refers to the accuracy on the \ood~dataset $\gD^{(\rm OoD)}$. Figure~\ref{fig_setting} also illustrates how to split all dataset to \ood~dataset and \id~dataset.

We also define the \id~data diversity of a dataset as $\#(\gI) /\#(\gC \times \gN)$, where $\#(\cdot)$ denotes a number of elements. Thus, the data diversity measures the portion of combinations included in the training distribution.
To directly compare the effect of the \id~data diversity on the \ood~accuracy, we vary the \id~data diversity such that the combinations in the distributions of lower \id~data diversity are included in the combinations of higher \id~data diversity,
while keeping the training set size constant,~\ie~$\#(\gD^{(\rm InD)}_{\rm train}(\gI))$ is constant for all \id~data diversity. 
These restrictions allows us to evaluate the performance of the DNN only by the difference in \id~data diversity, not by the difference in the amount of combinations or training examples.

\subsection{Datasets}
We use the following four datasets. These datasets have labels for both  category and  either orientation or illumination condition, in order to evaluate \ood~generalization. 
See~\ref{app:dataset} for further details than the ones provided in the following.

\paragraph{MNIST-Positions}
It is based on the MNIST dataset~\cite{Lecun1998mnist}. We created a dataset of $42\times 42$ pixels with nine numbers by resizing images to $14\times 14$ and placing them in one of nine possible positions in a $3\times 3$ empty grid. We call this dataset the \emph{MNIST-Positions} dataset. In our experiments, the digits are considered to be the category set, and the positions  where the digits are placed is considered as the orientation. We use nine digits and nine positions.
Samples are shown in Fig.~\ref{fig:mnist}. 
We used $54$K images for $\gD^{\rm{(InD)}}_{\rm train}$, $8$K images for $\gD^{\rm{(InD)}}_{\rm val}$ and $8$K images for $\gD^{\rm{(OoD)}}$.
Low, medium, and high \id~data diversity are set to be $2/9$, $4/9$, and $8/9$, respectively.

\paragraph{iLab-Orientations}
iLab-2M is a dataset created from iLab-20M dataset~\cite{Borji2016iLab}. The dataset consists of images of $15$ categories of physical toy vehicles photographed in various orientations, elevations, lighting conditions, camera focus settings and backgrounds. The image size is $256\times 256$ pixels. 
From the original iLab-2M dataset, we chose  
six categories (bus, car, helicopter, monster truck, plane, and tank) 
and six orientations.
We call it iLab-Orientations. 
Samples are shown in Fig.~\ref{fig:ilab}.  
We resized each image to $64\times 64$ pixels. We used $18$K images for $\gD^{\rm{(InD)}}_{\rm train}$, $8$K images for $\gD^{\rm{(InD)}}_{\rm val}$ and $8$K images for $\gD^{\rm{(OoD)}}$.
Low, medium, and high \id~data diversity are set to be $2/6$, $3/6$, and $5/6$, respectively.

\paragraph{CarsCG-Orientations}
CarsCG-Orientations 
is a new dataset 
that consists of images of ten types of cars 
in various conditions rendered by Unreal Engine.
It includes ten orientations, three elevations, ten body colors, five locations and three time frames (daytime, twilight, night). We synthesize images with $1920\times 1080$ pixels and resize them as $224\times 224$ pixels for our experiment.
We chose ten types of cars as  categories and ten orientations for each of them. 
Samples are shown in Fig.~\ref{fig:carcgs}. More samples are provided in~\ref{app:dataset}. 
In the experiment, we used $3400$ images for $\gD^{\rm{(InD)}}_{\rm train}$, $450$ images for $\gD^{\rm{(InD)}}_{\rm val}$ and $800$ images for  $\gD^{\rm{(OoD)}}$. Low, medium, and high \id~data diversity are set to be $2/10$, $5/10$, and $9/10$, respectively.

\paragraph{MiscGoods-Illuminations}
MiscGoods-Illuminations is a subset of DAISO-10, a novel dataset collected for this study. The dataset consists of ten physical miscellaneous goods photographed using a robotic arm  with five controlled  illumination conditions, two ways of object placement, twenty object orientations, and five camera angles. Each image is $640\times 480$ pixels in size.
We chose five categories (stuffed dolphin, stuffed whale, metal basket, imitation plant and cup) and five illumination conditions as shown in Fig.~\ref{fig:daiso}. More samples are displayed in~\ref{app:dataset}. We resize the images to $224\times 224$ pixels for our experiments. 
We used $800$ images for train $\gD^{\rm{(InD)}}_{\rm train}$, $200$ images for $\gD^{\rm{(InD)}}_{\rm val}$ and $400$ images for $\gD^{\rm{(OoD)}}$. 
Low, medium, and high \id~data diversity are set to be $2/5$, $3/5$ and $4/5$, respectively.

\subsection{\ood~accuracy results}\label{sec:perform_degrad}

We now demonstrate that these four datasets are extremely challenging for DNNs as these achieve low accuracy in \ood~conditions.
We examine the performance degradation in three \id~data diversity: low, medium, and high. 
Recall that we evaluate \id~accuracy in $\gD_{\rm val}^{\rm (InD)}$ and the \ood~accuracy in $\gD^{\rm (OoD)}$. We use ResNet-18~\cite{He2016resnet} trained with $\gD_{\rm train}^{\rm (InD)}$.
The experimental setup is introduced in Section~\ref{sec:set_of_exp}.

Figure~\ref{fig:performance_degrade} shows the \ood~accuracy degradation regarding the four datasets ranging low to high \id~data diversity.
While the \id~accuracy is more than $80\%$ for all four datasets at almost all data diversities (except for MNIST-positions), the \ood~accuracy showed a substantial degradation when the DNN was trained with low and medium \id~data diversities.
Between $20\%$ to $70\%$ performance degradation is observed in low \id~data diversity in all four datasets.
In medium \id~data diversity, large performance degradation ranging from $10\%$ to $50\%$ is observed, and for high \id~data diversity, there is more than $10\%$ performance degradation in CarsCG-Orientations and MiscGoods-Illuminations datasets.
Thus, dramatic drops of accuracy are observed in \ood~conditions, which confirms that these benchmarks are very challenging for DNNs.

\ood~accuracy is often overlooked in standard computer vision benchmarks and only \id~is usually reported. This is usually due to the difficulty of measuring \ood~accuracy. Our datasets enable evaluating \ood~accuracy in a controlled way that facilitates understanding the different factors that may affect the \ood~accuracy. The performance degradation in \ood~conditions is expected when deploying application of deep learning.
Recently, it has been reported that even a small amount of data bias can cause major performance degradation~\cite{recht2018cifar10.1}, and this is reconfirmed for our four datasets. Also, the drop of accuracy in our datasets is dramatic, specially for low \id~data diversity. Our datasets allow to gain an understanding of the specific biasing factors in the dataset,~\ie~orientation and illumination conditions, and analyze aspects such as the \id~data diversity.

\section{Three approaches to improve \ood~accuracy}\label{sec:three_app}

We now introduce the three approaches to address the performance drop of accuracy in \ood~conditions, which are ``late-stopping'', ``tuning the batch normalization momentum'' and ``invariance los''. These three approaches are independent on each other and tackle different aspects of the DNN training.

\subsection{Late-stopping}

The stopping criteria for training   is known to have an impact on the DNNs performance~\cite{cataltepe1999,caruana2001,yao2007}. In particular, stopping the training before convergence of the training accuracy,~\ie~early stopping, is known to prevent overfitting in shallow classifiers~\cite{prechelt1998}. However, these results are with respect to \id~accuracy, and  little is known regarding the relation between the stopping critaria and \ood~accuracy.
We therefore run experiments with a large number of training epochs (up to $1000$ epochs) in order to investigate any patterns. Figure~\ref{fig:three_approaches}(a) shows the change of \id~and \ood~ accuracy when ResNet-18 is trained with the medium \id~data diversity. Surprisingly, the \ood~accuracy, unlike the \id~accuracy, continued to increase in performance after training during a large number of epochs.
This phenomenon was not known because usually only the \id~accuracy is analyzed. We denote the approach of continuing the training of a DNN after the convergence of \id~validation accuracy as ``late stopping".

\subsection{Tuning batch normalization}
Batch normalization (BN)~\cite{Ioffe2015} is a method used to speed-up and stabilize the training of DNN networks through normalization of the layers' inputs by re-centering and re-scaling them.  Batch normalization has also been reported to act as a regularizer and improve generalization~\cite{luo2018,SchneiderRE0BB20}. Thus, it is reasonable that batch normalization could help improving \ood~generalization but this has not been studied so far.

Batch normalization uses the so called moving average to recenter the layer's input. Let  $\vv_{\text{ma}}(t)$ be the moving average at training step $t$. The moving average is updated at each training step in the following way: 
\begin{align}
\vv_{\text{ma}}(t) = (\beta - 1)\vv_{\text{mean}}(t) + \beta\vv_{\text{ma}}(t - 1),
\end{align}
where $\vv_{\text{mean}}(t)$ is the mean activity over the batch of the $t$-th training step, and $\beta \in [0,1]$ is called momentum and balances the update of the moving average between $\vv_{\text{mean}}(t)$ and itself. Note that the only hyperparameter available for batch normalization is $\beta$, and we use this to adjust it.
Usually, $\beta$ is set to $0.9$ or $0.99$, which is the default value  in standard deep learning frameworks.
We use the default value $0.99$ that is employed by the TensorFlow library~\cite{abadi2016tensorflow}.

We investigated how the OoD generalization performance behaves depending on the value of the batch normalization momentum, $\beta$. Figure~\ref{fig:three_approaches}(b) shows the learning curves of ResNet-18 trained on MiscGoods-illuminations with the medium \id~data diversity. Experimentally, we found that the tuning momentum parameter, $\beta$, can have a significant positive impact on the \ood~generalization performance.
Generally, the default value of $\beta = 0.99$ was too large for almost all cases in our experiments.
We call this approach as tuning batch normalization or ``tuning BN''.

\subsection{Invariance loss}
The ``invariance loss'' approach is intended to increase the invariance score that is introduced in Madan~\etal~\cite{madan2020capability}, which we explain in Section~\ref{sec:invsel}. 
This invariance score measures the degree of invariance in the neural activity of intermediate layers, and previous works have shown that DNNs that generalize better to \ood~conditions have developed larger degrees of invariance in the intermediate layers.

Concretely, we encourage the emergence of invariant representations by taking pairs  of images that belong to the same category and enforce that the neural activity is as similar as possible. To do so, we use the Euclidean distance between the activities of neurons in an intermediate layer caused by the pairs of images, and add this as an additional loss term to the classification loss.
Figure~\ref{fig:three_approaches}(c) shows the scheme of this approach. 
Let $\vg(\cdot \ ;\vtheta_{g})$ be the neural activity of a DNN's intermediate layer, where $\vtheta_{g}$ are the parameters of the DNN before the intermediate layer. Let $\vf(\cdot \ ; \vtheta_{f})$ be the output of the DNN given as input the intermediate layer, $\vg(\cdot \ ;\vtheta_{g})$, where $\vtheta_{f}$ are the DNN parameters from the intermediate layer to the output of the network. 
Thus, the neural activity of the intermediate  layer for an image $\rvx$ is $\vg(\rvx;\vtheta_{g})$ and the  output of the whole network  is $\vf(\vg(\rvx;\vtheta_{g});\vtheta_{f}) $. 
Let $\rvx$ be a training image,  and  let $\rvx^\prime$ be another image that belongs to the same  category as $\rvx$, and is sampled from the training data $D_{\rm train}^{\rm (InD)}$ according to some sampling strategy (in our experiments, we use  random sampling with uniform distribution across the training images of the same category).  
Thus, the invariance loss is expressed as 
\begin{align}\label{equ:inv_def}
\|\vg(\rvx; \boldsymbol{\theta}_{ g}) - \vg(\rvx^\prime; \boldsymbol{\theta}_{ g})\|_{2}. 
\end{align}
This term is added to the categorical cross entropy loss
weighted with a hyperparameter that we call $\lambda$, such that the invariance loss term acts as a regularization term.
Note that the invariance loss is equivalent to the contrastive loss~\cite{LeCun2006} for positive examples in the context of metric learning, but it has not been used so far to improve generalization to \ood~orientations and illumination conditions. 
\section{Selectivity and invariance for \ood~generalization}\label{sec:invsel}

We now revisit the mechanism at the individual neuron level of intermediate layers that previous works have suggested that facilitates \ood~generalization, \ie~individual neurons being  selective to a category and invariant \ood~conditions. 
This mechanism has been shown to explain the improvement in OoD accuracy with increased InD data diversity~\cite{zaidi2020robustness,madan2020capability}.

For a given intermediate layer of the DNN, let ${\alpha}^{j}_{cn}$ be the average activity for the $j$-th neuron over all images with the $c$-th category and the $n$-th orientation or illumination condition. For neuron $j$, the activity is $0$-$1$ normalized.
Let $c^{*j}$ be  the category that a neuron $j$ is most active on average, \ie~$c^{*j} := {\rm arg max}_{c}\sum_{n}{\alpha}^{j}_{cn}$.
This is called preferred category.
The selectivity score $S^{j}$ is defined as
\begin{align}
    S^{j} := \frac{\hat{\alpha}^{j} - \bar{\alpha}^{j}}{\hat{\alpha}^{j} + \bar{\alpha}^{j}},
\end{align}  
where,
    $\hat{\alpha}^{j} := \frac{1}{\#(\gN)} \sum_{n} {\alpha}_{c^{*j}n}^{j}$ and
    $\bar{\alpha}^{j} := \frac{\sum_{c \neq c^{*d}}\sum_N {\alpha}_{cn}^{j}}{\#(\gC)(\#(\gC)-1)}$
denote the average activity for the preferred category  
and for the remaining categories, respectively.
This selectivity score ranges from zero to one and takes its maximum value in the case that the neuron average activity,  ${\alpha}^{j}_{cn}$, is $0$ for all categories except for the preferred category, \ie~the neuron is only active for the preferred category.
The invariance score $I^{j}$ is defined as
\begin{equation}\label{eq:inv}
    I^{j} := 1 - ({\rm max}_{n} {\alpha}_{c^{*j}n}^{j} - {\rm min}_n {\alpha}_{c^{*j}n}^{j}), 
\end{equation}
and it also ranges from zero to one and takes the maximum 
in the case that the average activity, ${\alpha}^{j}_{cn}$, takes the same value for the preferred category regardless of the orientation and illumination conditions.

Finally, we define the SI score of a neuron as the geometric mean of the selectivity and invariance scores,~\ie  $\sqrt{S^{j}I^{j}}$.
Neurons that have a larger SI score are active for specific categories independently on the orientation and illumination conditions. Networks with neurons that have larger SI scores  have been observed to generalize better in~\ood~conditions. In order to provide a score that summarizes the SI score across all neurons in the layer, we use the upper $20$ percentile of the scores among all neurons. This is because  not all neurons are required to have larger SI to improve \ood~generalization, and we just take into account a portion of  neurons with the highest SI score. In the experiments, we use this summary of the SI score across neurons to assess whether the three approaches we introduce yield improved \ood~accuracy through improving selectivity and invariance.

\section{Experiments and analysis} \label{SecExperiments}

We first introduce the experimental setup, and then report the \ood~accuracy facilitated by the three approaches explained in Section~\ref{sec:three_app}. Finally, we analyze whether this boost of \ood~accuracy is driven by selective and invariance mechanism revisited in Section~\ref{sec:invsel}.

\subsection{Experimental setting}\label{sec:set_of_exp}

We apply the three approaches to improve \ood~accuracy to ResNet-18~\cite{He2016resnet} and evaluate its effectiveness in the aforementioned datasets (MNIST-Positions, iLab-Orientations, CarsCG-Orientations, and MiscGoods-Illuminations). Standard ResNet-18 is adopted as the network for all experiments and we trained it in the standard manner. 
Namely, all neurons 
employ the ReLU activation function $g(z) = \max \{0, z  \}$~\cite{dahl2013improving} and Glorot uniform initializer~\cite{Glorot10understandingthe} is adopted for the network weights initialization for all experiments.
Adam~\cite{Diederik2015Adam} is employed as the optimization algorithm. 
The pixels of images are normalized within $0$ to $1$ as a preprocessing for all datasets.

We run five trials in all cases and report mean accuracy and its $95\%$ confidence interval.
In each trial, the \id~combinations  are chosen randomly as long as they satisfy the conditions explained in Section~\ref{subsec:Creating spurious correlation}, and the \ood~combinations are created accordingly.  Each of the four approaches, including baseline, is subjected to a hyper-parameter search before performing the five trials. We select the hyper-parameters in a different trial from the ones used to report \ood~accuracy. In this reserved trial, we select the hyper-parameters with the highest \ood~accuracy by grid search. 
For all tested approaches, we selected a learning rate  in $ \{0.1, 0.01, 0.001, 0.0001, 0.00001\}$, and other hyper-parameters depending in the approach. In the following we detail the experimental setting of the different approaches.

\paragraph{Late-stopping} The epoch size is set to $1,000$ epochs for late stopping, and $100$ epochs for the other approaches, including baseline. 
We confirmed that $100$ epochs are sufficient for convergence in InD accuracy by the preliminary experiments.
For late stopping, we run as many epochs as computing resources allow (about a week of training).

\paragraph{Tuning batch normalization} For tuning batch normalization, we perform a grid-search for  $\beta = \{0.01, 0.1, 0.5, 0.9, 0.99\}$ in addition to the learning rate.
For the other approaches, we use $0.99$ as a momentum parameter $\beta$ for batch normalization layer, which is the default value in TensorFlow.

\paragraph{Invariance loss} Invariance loss is applied to the last ReLU activation layer “activation\_17" which has $512$ neurons. We keep fixed the pairs of images in which invariance is enforced, and we randomize the pairs from time to time. We perform a grid search to determine how frequently we randomize the pairs of images  (the choices are randomizing every  $\{10, 20, 50, 100\}$ epochs). The weight of the invariance loss term, $\lambda$, is also selected via a grid search among the following values: 
$\lambda = \{1.0, 0.1, 0.01, 0.001, 0.0001\}$.
For more details we refer the reader to \ref{app:networks}.

\subsection{Improvement of OoD accuracy}

\begin{figure*}[t]
    \begin{tabular}{c|c|c}
    $\scriptstyle \mbox{Low InD data diversity}$ & $\scriptstyle \mbox{Medium InD data diversity}$ & $\scriptstyle \mbox{High InD data diversity}$ \\[-0.5em]
        \subfigure{
            \includegraphics[width=0.23\textwidth]{./figs/result/top3-2-1.pdf}} &
        \subfigure{
            \includegraphics[width=0.23\textwidth]{./figs/result/top3-2-3.pdf}} &
        \subfigure{
            \includegraphics*[width=0.23\textwidth]{./figs/result/top3-2-2.pdf}} \\
                \setcounter{subfigure}{0}
        \subfigure[
]{\label{fig:low_unseen}
            \includegraphics[width=0.30\textwidth]{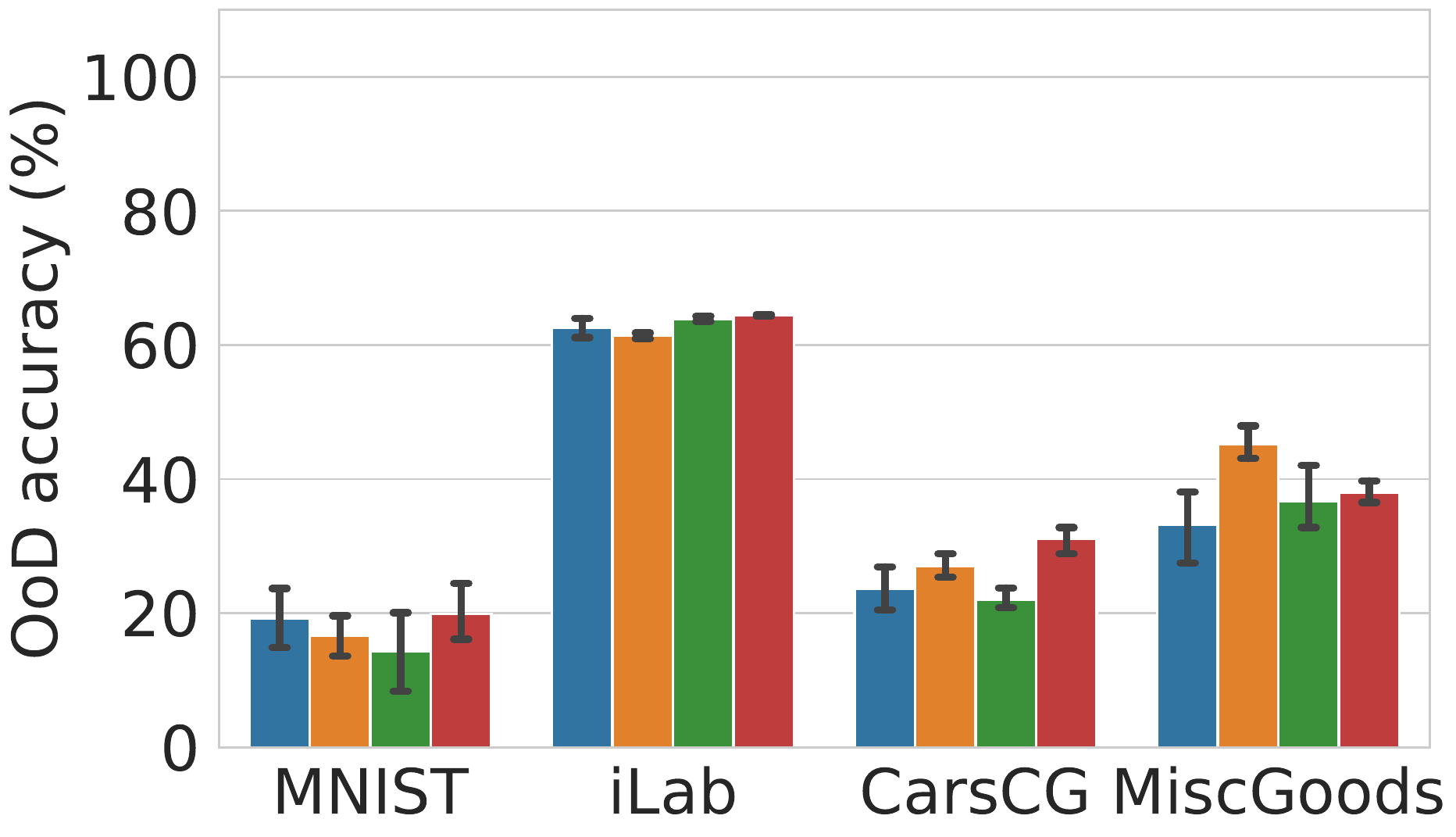}} &
        \subfigure[]{\label{fig:medium_unseen}
            \includegraphics[width=0.30\textwidth]{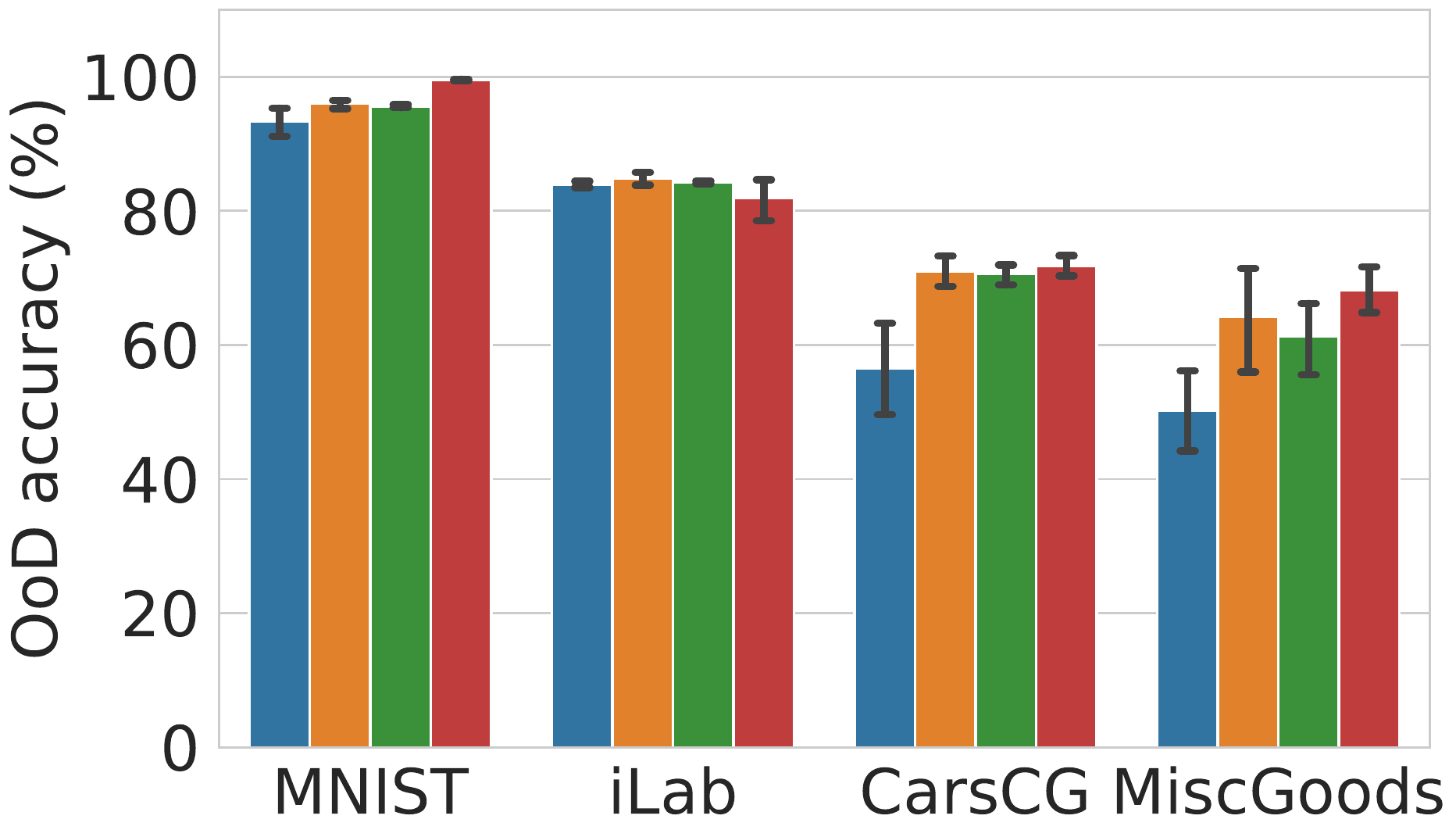}} &
        \subfigure[]{\label{fig:high_unseen}
            \includegraphics*[width=0.30\textwidth]{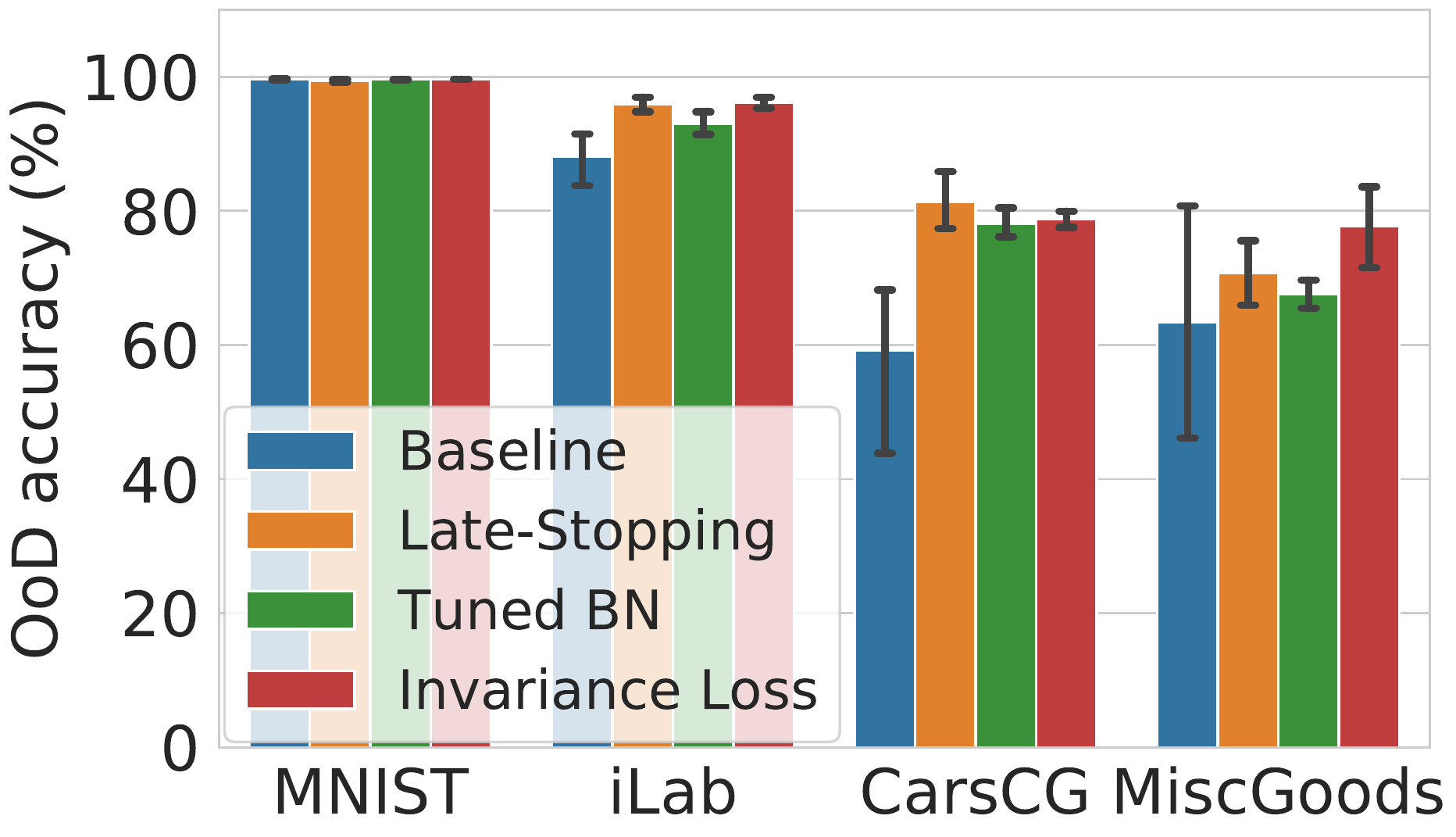}} \\

\subfigure[
]{\label{fig:low_unseen_diff}
            \includegraphics[width=0.30\textwidth]{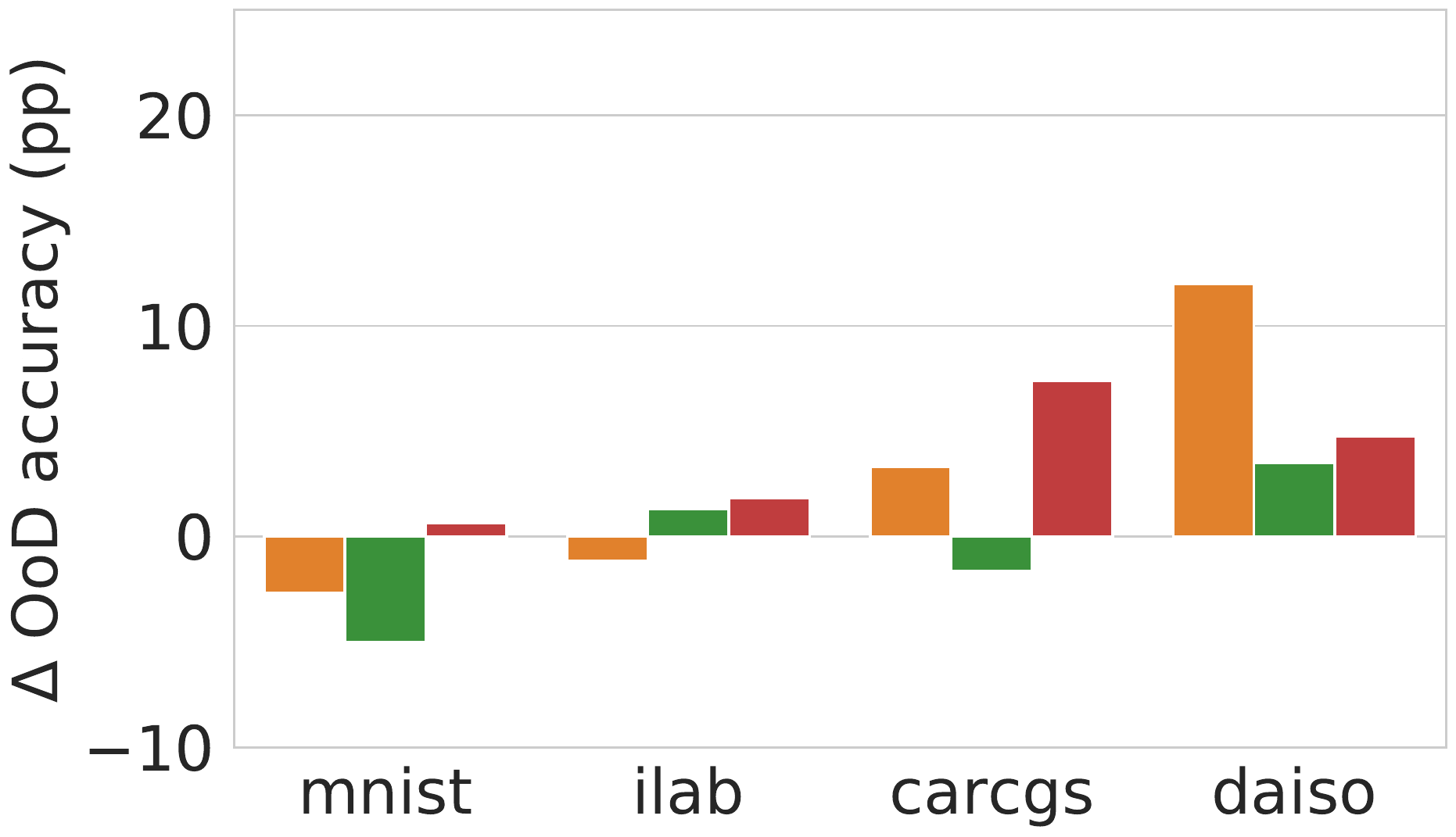}} &
        \subfigure[]{\label{fig:medium_unseen_diff}
            \includegraphics[width=0.30\textwidth]{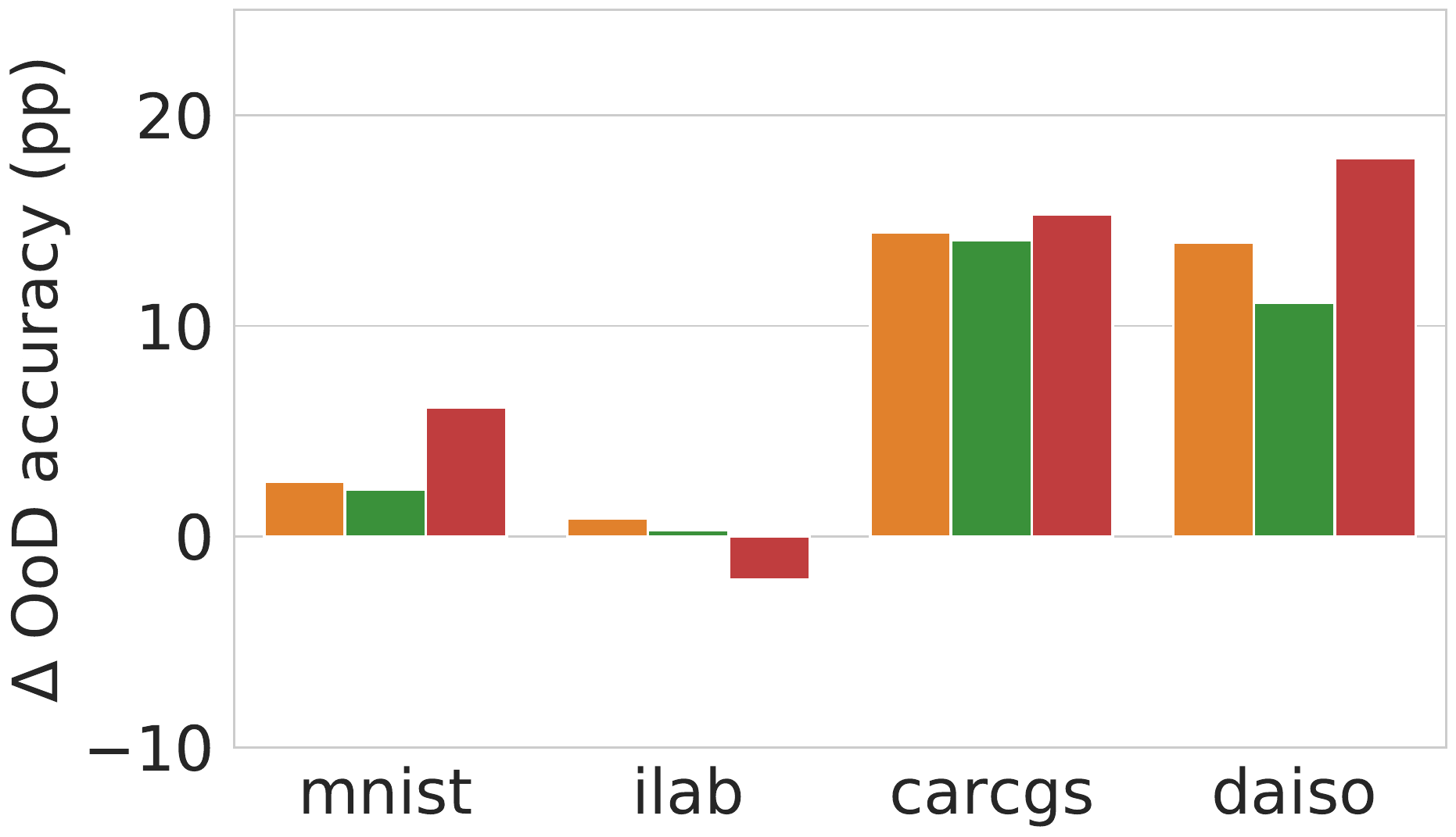}} &
        \subfigure[]{\label{fig:high_unseen_diff}
            \includegraphics*[width=0.30\textwidth]{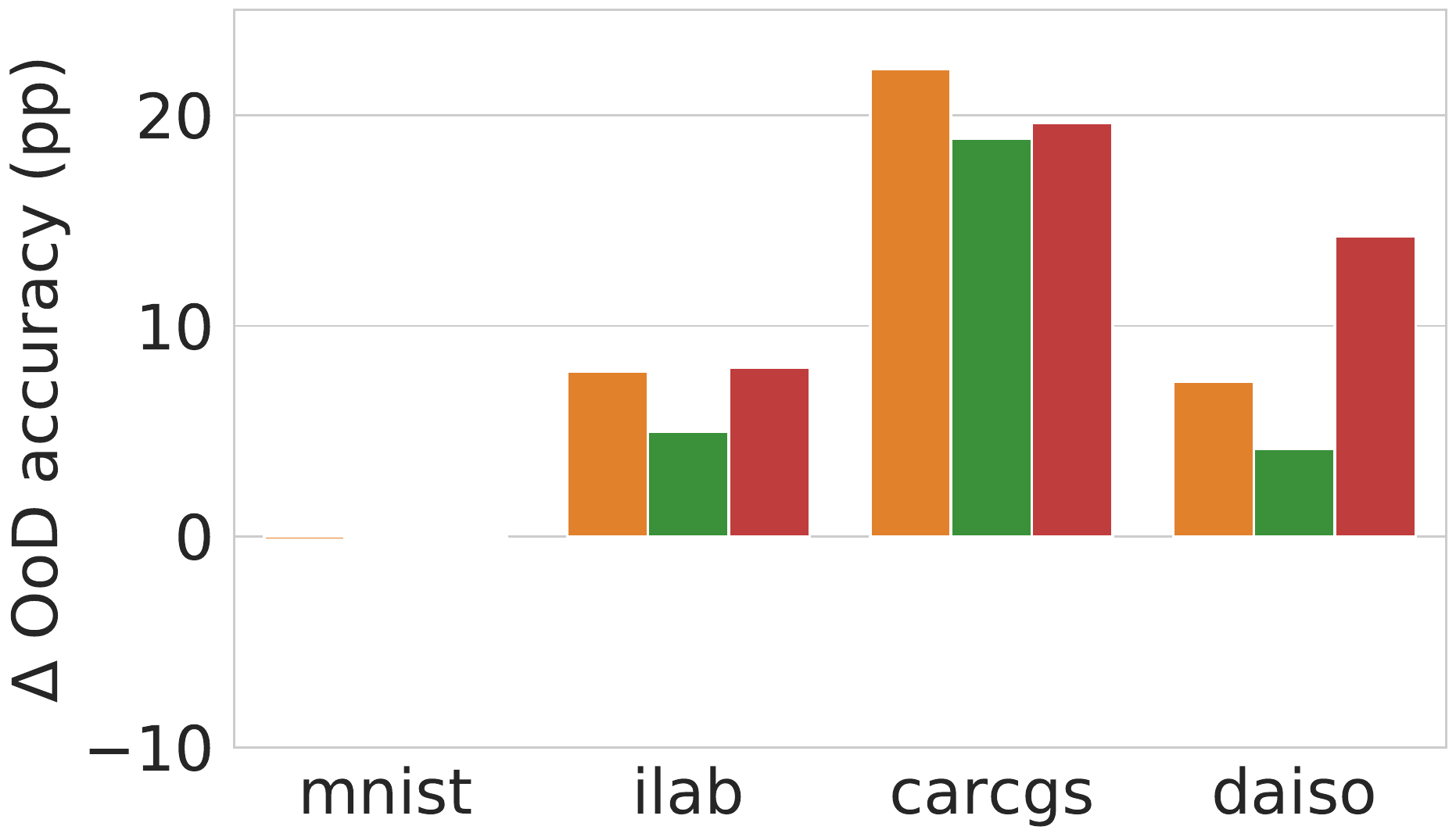}}
    \end{tabular}
\caption{
\emph{Performance improvement of the mean \ood~accuracy.} 
Top figures show the examples of \id~combinations, \ie~InD data diversity, in MiscGoods-Illuminations dataset.
(a), (b), and (c) show the mean \ood~accuracy of the three approaches and the baseline for different InD data diversities. Error bars show $95\%$ confidence interval.
(d), (e), and (f) show the increase of the mean OoD accuracy by the three approaches over the baseline.  
The unit ``pp" in figures denotes percentage points, \ie~the unit for the arithmetic difference of two percentages.
}
\label{fig:compare_perform}
\end{figure*}

Figure~\ref{fig:compare_perform} compares the mean OoD accuracy between the baseline and  the three approaches for all tested data sets and all tested \id~data diversities.
Looking at the case of the CarsCG-Orientations and MiscGoods-Illuminations datasets, we can see that the three approaches increase the mean OoD accuracy at almost all the data diversities. Comparing the three approaches, late stopping and invariance loss both achieves the best improvement rate in some combinations, and batch norm momentum does not achieve the best improvement in any combination. The highest improvement  of $22.2\%$ is achieved by late stopping with a high InD data diversity. 
 The performance improvement across datasets and data diversities is remarkable. Only in  iLab-Orientations dataset is relatively small, but for high InD data diversity in this dataset, all three approaches achieve better OoD accuracy than the baseline approach.
For MNIST-Positions, all three approaches showed an improvement in performance with medium InD diversity. In~\ref{app:results} we report the learning curves and the \id~accuracy for a more detailed depiction of the effects of the three approaches during training.

\begin{figure}[ht]
\centering          \includegraphics[width=0.4\textwidth]{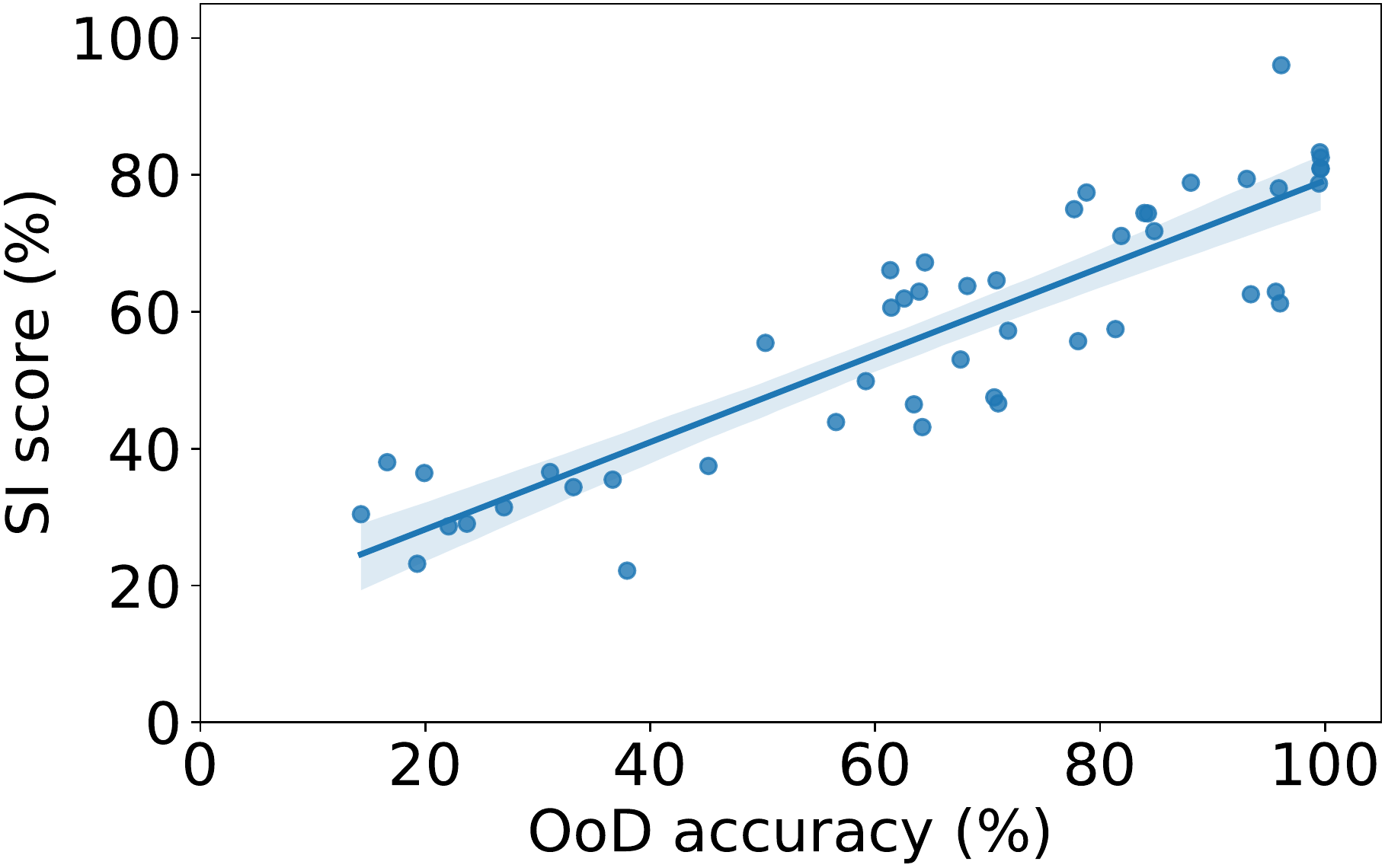}
\caption{\emph{Correlation analysis.}
This figure shows the correlation between \ood~test accuracy and SI score. The Pearson's correlation coefficient is $0.891$.
}
\label{fig:col_analysis}
\end{figure}

We also investigated whether the three approaches combined together are more effective than the best of three approaches applied alone. Thus, we trained networks using late-stopping, tuned BN and invariance loss together. We call this approach ``three approaches together''. Another way of combining the three approaches is training networks with each approach alone and then selecting the best of the approaches using a validation set. We call this approach ``best of three approaches alone''.
The hyper-parameter tuning method of these combined approaches is detailed in~\ref{app:combined}.
Table~\ref{tbl:combination} shows the comparison between these two combination approaches and also the baseline,~\ie~the network trained without any approach to improve the \ood~accuracy. The table reports the number times a method outperformed another method across all datasets and \id~data diversity.
The results show that using the best of the three approaches alone obtains the best results in the vast majority of experiments.
Interestingly, the three approaches together performs worse than the baseline for more than half of the experiments. 
This indicates that the three approaches together interfere with each other and should not be used.

\begin{table*}[t!]
\caption{\emph{Comparison of ways to combine the three approaches.} We compare the the best of the three approaches alone (\ie~training a network different times each with one of the three approaches alone, and then selecting the best of the three in a validation set), training with the three approaches together (\ie~training a network using the three approaches together), and the baseline (\ie~training the network without using any approach). Results compare how many times each of these strategies outperformed another strategy,  across  \id~data diversities in each of the four datasets.}
\begin{center}
\begin{tabular}{llllll}
\hline
 Comparison of OoD accuracy            &MNIST  &iLab  &CarsCG  & MiscGoods & Total  \\ \hline\hline
 Best of 3 approaches alone vs  Baseline        & $3$ vs $0$ & $2$ vs $1$  & $3$ vs $0$ & $3$ vs $0$& $11$ vs $1$   \\
 3 approaches together vs Baseline        & $1$ vs $2$ & $1$ vs $2$ &  $1$ vs $2$ & $2$ vs $1$ & $5$  vs $7$ \\
 Best of 3 approaches alone vs  3 approaches together  & $3$ vs $0$ & $2$ vs $1$ &  $3$ vs $0$ & $1$ vs $2$ & $9$ vs $3$\\ \hline
\end{tabular}
\label{tbl:combination}
\end{center}
\end{table*}

\subsection{Analysis for selectivity and invariance mechanism}

\begin{table*}[th]
\caption{\emph{Analysis of the dependency between  improvements of \ood~accuracy and SI score.}
 This table shows the relative frequency of improvement ($+$) or degradation ($-$) of the mean OoD accuracy $\Delta_{\rm acc}$ or mean SI score $\Delta_{\rm SI}$.
Relative frequency $P(x)$ is calculated by counting the number of cases that satisfy the condition $x \in \{\Delta_{\rm acc}^{+},\Delta_{\rm SI}^{+}\}$, and normalize it by total number of cases (\ie~$12$, $3$ possible \id~data diversity $\times$ $4$ datasets).
Conditional relative frequency $P(y|x)$ is also calculated by counting the number combinations satisfying $y \in \{\Delta_{\rm acc}^{+},\Delta_{\rm acc}^{-} \}$ in the condition of $x \in  \{\Delta_{\rm SI}^{+},\Delta_{\rm SI}^{-} \}$, and divide it by the number of combinations satisfying $x$.
The first and second columns show the proportion of cases where the mean OoD accuracy  $\Delta_{\rm acc}^{+}$ and the mean SI score  $\Delta_{\rm SI}^{+}$ increased, respectively.
The third column shows the proportion of  cases where the mean OoD accuracy increased $\Delta_{\rm acc}^{+}$  when the mean SI score increased $\Delta_{\rm SI}^{+}$.
The fourth column shows the proportion of cases where the mean OoD accuracy increased $\Delta_{\rm acc}^{-}$  when the mean SI score increased $\Delta_{\rm SI}^{-}$.
}
\begin{center}
\begin{tabular}{lllll}
\hline
 Approach   &  $P(\Delta_{\rm acc}^{+})$ & $P(\Delta_{\rm SI}^{+})$ & $P(\Delta_{\rm acc}^{+}|\Delta_{\rm SI}^{+})$ & $P(\Delta_{\rm acc}^{+}|\Delta_{\rm SI}^{-})$   \\ \hline\hline
Late-Stopping ($\%$) &   $75.0~(9/12)$  & $50.0~( 6/12)$  & $83.3~(  5/6 )$& $66.6~( 4/6 )$   \\ \hline
Tuned BN ($\%$) &$75.0~(9/12)$& $83.3~(10/12)$ & $80.0~( 8/10 )$ & $50.0~( 1/2 )$   \\ \hline
Invariance Loss ($\%$)&$91.7~(11/12)$ &$83.3~(10/12)$& $100.0~( 10/10 )$ & $50.0~( 1/2 )$\\\hline\hline
Total ($\%$) &$80.6~(29/36)$&$72.2~(26/36)$ & $88.4~( 23/26 )$& $60.0~( 6/10 )$ \\\hline
\end{tabular}
\end{center}
 
\label{tbl:causual}
\end{table*}

Figure~\ref{fig:col_analysis} shows the relationship between the SI score of the last ReLU layer and the OoD accuracy for all combinations of dataset, InD data diversity, and approach (details are provided in Fig.~\ref{fig:compare_col_analysis}). 
We can see that there is a large correlation between SI score and OoD accuracy (Pearson’s correlation coefficient is $0.891$). While it has already been shown in Madan \etal~\cite{madan2020capability} that increasing the InD data diversity improves the \ood~accuracy and the SI score, here we show for the first time that approaches that targets improving the \ood~accuracy also yield increases of the SI score.

Next, we analyze the relationship between improvements of \ood~accuracy and increases of the SI score. We investigate whether increases of the SI score always precede improvement of \ood~accuracy, which serves to assess whether invariant representations drive \ood~generalization in a more stringent way than the correlational analysis presented before. Let $P(\Delta_{\rm acc}^{+})$  be the probability that the \ood~accuracy increases when using one of the three approaches to train the network, compared to not using it. Also, let $P(\Delta_{\rm SI}^{+})$ be the probability that the SI scores increases when using one of the three approaches, compared to not using it. The conditional probabilities between these two events provides insights regarding whether increases of the the SI score precedes the improvements of the \ood~accuracy. We calculate the probabilities by evaluating the frequency that the events happen across datasets and \id~data diversities. We report them in Table~\ref{tbl:causual}.

We observe by analyzing  $P(\Delta_{\rm acc}^{+})$ that the \ood~accuracy increases very often with the three approaches, at least $75\%$ of the cases. In particular,  the OoD accuracy increased $91.7\%$ of the cases for the invariance loss.
The analysis of $P(\Delta_{\rm SI}^{+})$ shows that tuned BN and invariance loss increase the SI score $83.3\%$ of the cases.
This suggests that these two approaches tend to improve the SI score. For late-stopping this trend is not as strong. Yet, when analyzing $P(\Delta_{\rm acc}^{+}|\Delta_{\rm SI}^{+})$, we observe that for the three approaches, increases of the SI score precede the improvements of \ood~accuracy (this is in $83.3\%~(5/6)$, $80.0\%~(8/10)$ and $100\%~(10/10)$ of the cases for late-stopping, tuned BN and invariance loss, respectively).  Note that the invariance loss directly encourages to increase the SI score, and when the SI score in fact increases, the  OoD accuracy always has improved. Late stopping and tuning batch normalization momentum do not directly encourage to increase the SI score, but we observe that they do increase the SI score most of the cases, and when this happens, the OoD accuracy is also improved in more than $80.0~\%$ of the cases. Thus, these results suggest that the improvement of \ood~accuracy is strongly driven by the increase of the SI score.

Finally, we observe by analyzing $P(\Delta_{\rm acc}^{+}|\Delta_{\rm SI}^{-})$, that when the SI score has not increased after applying one of the three approaches, the \ood~accuracy still improves in a non-negligible number of cases.
This suggests the existence of another mechanism that can improve the \ood~accuracy even if the selectivity and invariance mechanisms did not emerge.
However, one possible limitation of this interpretation  is that selectivity and invariance may have emerged but have not been captured by the SI score, because the SI score may not quantify the emergence of these mechanisms  in the most precise way.  Thus, we can not make any assertion beyond the fact that it is unclear what are the neural mechanisms that facilitate \ood~generalization when the three approaches do not manage to increase the SI score. This result motivates follow-up investigations.  

In summary, in this study we provided evidence that the invariance and selectivity mechanism drives \ood~generalization. Also, we found cases in which improvements of \ood~generalization may not be preceded by the strengthening of the selectivity and invariance mechanism in the neural representations, which requires future work proposing novel mechanisms to explain these cases. We believe our experimental framework will facilitate such future discoveries.

\section{Conclusion}

We have shown that late-stopping, tuning the batch
normalization momentum parameter, and optimizing the invariance loss during learning lead to substantial improvements of the DNN recognition accuracy of objects in \ood~orientations and illuminations (in some cases more than $20\%$). These improvements are consistent across four datasets, and different degrees of dataset bias.  We also corroborated that the neural mechanisms of selectivity to a category and invariance to orientations and illuminations, at the individual neuron level, lead to the aforementioned improvements of \ood~recognition accuracy. Namely, we found that in the majority of trials where any of the three approaches yield an increase of selectiviy and invariance, resulted in improvements of the \ood~recognition accuracy. Nonetheless, our analysis also revealed that other mechanisms different from selectivity and invariance may also exists, as we observed that gains of \ood~recognition accuracy were not preceded by an increase of the SI score in some trials. What are the neural mechanisms that drive \ood~generalization in these cases remains as an open question for future work. Furthermore, there are also other novel questions derived from our results that motivate future works: Is there any effective way of combining the three approaches investigated in this paper that leads to even more improvements of \ood~generalization? Are these approaches applicable to other factors beyond orientations and illumination conditions? How these approaches relate to biological learning systems? We hope that the substantial improvements of \ood~recognition accuracy that we demonstrated  in this paper motivate new research to address the fascinating questions that have cropped up ahead of us.

Finally, we would like to highlight that poor \ood~generalization is one of the issues of machine learning that needs to be urgently addressed in order to allow for safe and fair AI applications.
 We hope that this research serves as a basis for further improvements of \ood~generalization.  
 
 \section*{Code and data availability statement} 
The source code used in this study is publicly available in the following GitHub repository: \url{https://github.com/FujitsuResearch/three-approaches-ood}. All datasets used in this study are publicly available (see \ref{app:dataset}).

\section*{Acknowledgments}
We are grateful to Tomaso Poggio and Hisanao Akima for their insightful advice and warm encouragement. We thank Shinichi Matsumoto and Shioe Kuramochi for their assistance to create CarsCG and DAISO-10 datasets, respectively. This work was supported by Fujitsu Limited (Contract No. 40008819 and 40009105) and by the Center for Brains, Minds and Machines (funded by NSF STC award CCF-1231216). 
PS and XB are partially supported by the R01EY020517 grant from the National Eye Institute (NIH), and SM and HP are partially funded by NSF grant III-190103.

\section*{Conflicts of Interests Statement}
The authors declare that the research was conducted in the absence of any commercial or financial relationships that could be construed as a potential conflict of interest.

{\small
\bibliographystyle{elsarticle-num}
\bibliography{egbib}
}

\appendix

\onecolumn
\section{Details of datasets}
\label{app:dataset}

\subsection{MNIST-Positions}

\begin{figure*}[ht]
    \centering
    \includegraphics[width=\columnwidth]{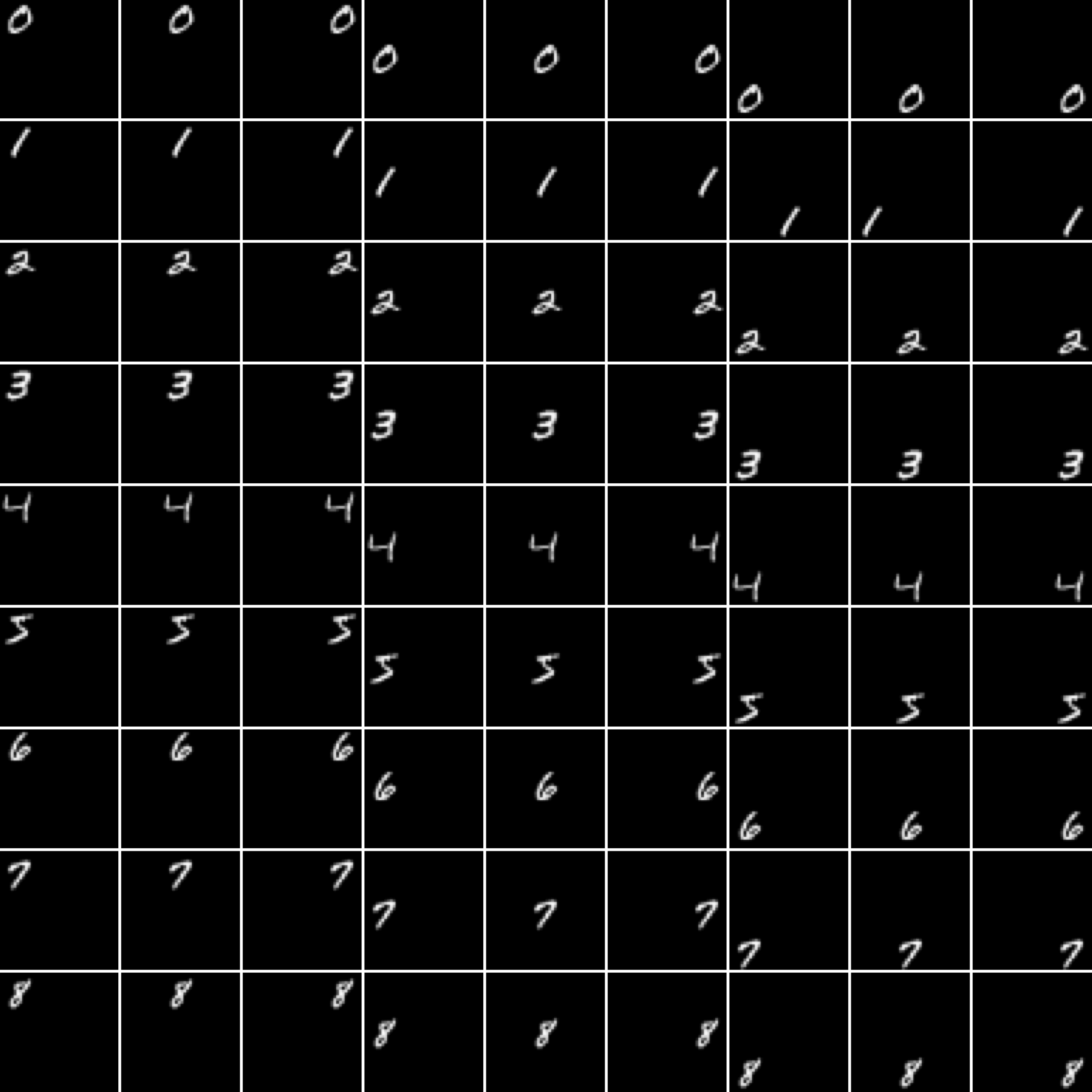}
    \caption{Sample images of MNIST-Positions dataset arranged in a grid pattern. Each row indicates the number as the object category. MNIST-Poitions include nine numbers from 0 to 8. Each column indicates the positions as the condition category. There are 9 positions in this dataset.}
    \label{fig:mnist_grid}
\end{figure*}

Starting with the MNIST dataset~\cite{Lecun1998mnist},  which is available at \url{http://yann.lecun.com/exdb/mnist/} (Last access: Dec.~15, 2021) under the CC0 license, we created a dataset of 42$\times$42 pixels with nine numbers (0 to 8) by resizing images to 14$\times$14 and placing them in one of 9 possible positions in a 3$\times$3 empty grid. We call it MNIST-Positions. 
Fig.~\ref{fig:mnist_grid} shows the all categories and positions of MNIST-Positions. 
In our experiments, the numbers are considered to be the object category set $\gC$ and the positions where the numbers are placed is considered as the condition set $\gN$.
Thus, it is written as $\#(\gC)=\#(\gN)=9$.

\begin{figure*}[p]
    \centering
    \includegraphics[width=\columnwidth]{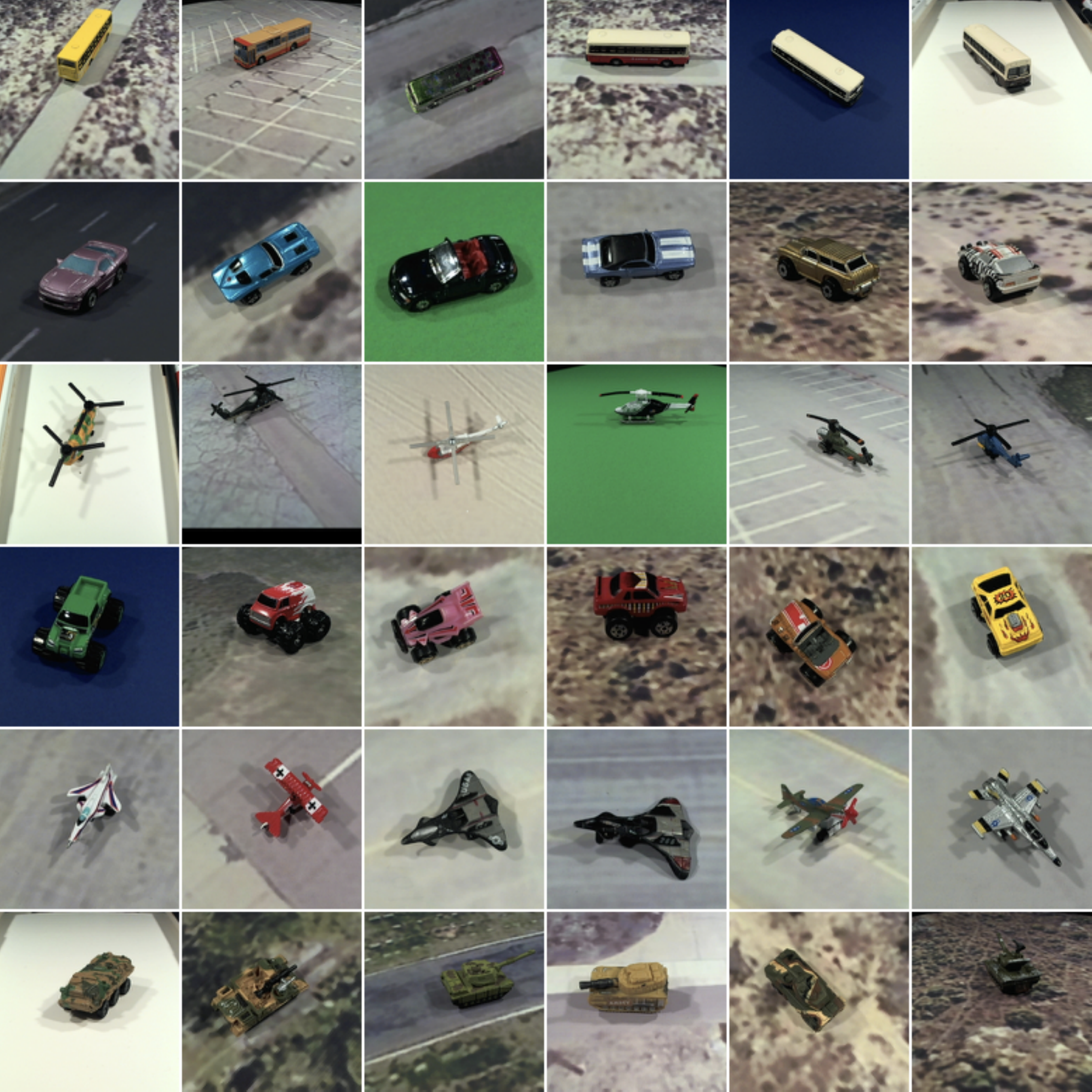}
    \caption{Sample images of iLab--Orientations dataset arranged in a grid pattern. Each row indicates the object category. iLab-Orientations include six object categories --- bus, car, helicopter, monster truck, plane, and tank. Each column indicates the orientations as the condition category. There are 6 orientations in this dataset.}
    \label{fig:ilab_grid}
\end{figure*}

\subsection{iLab-Orientations}

iLab-2M is a dataset created from iLab-20M dataset~\cite{Borji2016iLab}:  freely and publicly available at \url{https://bmobear.github.io/projects/viva/} (Last access: Dec.~15, 2021).
The dataset consists of images of 15 categories of physical toy vehicles photographed in various orientations, elevations, lighting conditions, camera focus settings and backgrounds. It has 1.2M training images, 270K validation images, 270K test images, and each image is 256$\times$256 pixels. 
We chose from the original iLab-2M dataset 
six categories --- bus, car, helicopter, monster truck, plane, and tank 
as $\gC$ and six orientations as $\gN$.
We call it iLab-Orientations. 
Fig.~\ref{fig:ilab_grid} shows samples of the all categories and orientations of  iLab-Orientations dataset.

 \begin{figure*}[p]
    \centering
    \includegraphics[width=\columnwidth]{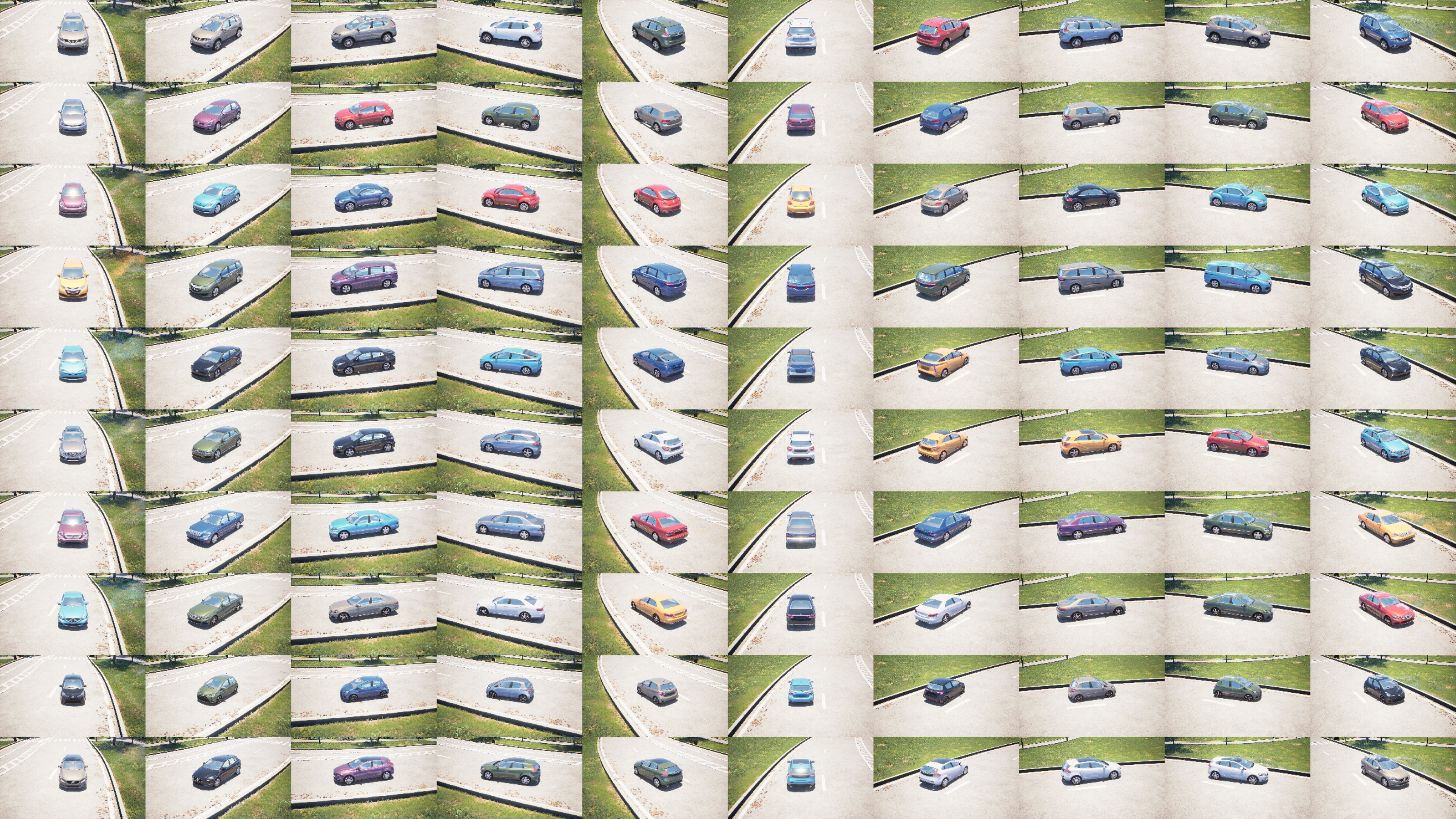}
\caption{Sample images of each object category and orientation of CarCGs-Orientations. Each row indicates object categories --- Nissan Rogue\textsuperscript \textregistered, Volkswagen\textsuperscript \textregistered~Golf, Volkswagen\textsuperscript \textregistered~Beetle, Honda Odyssey\textsuperscript \textregistered, Toyota Prius\textsuperscript \textregistered, Mercedes Benz\textsuperscript \textregistered~ A-Class, Lexus\textsuperscript \textregistered~LS, Mercedes Benz\textsuperscript \textregistered~ E-Class, Toyota Yaris\textsuperscript \textregistered and Volvo\textsuperscript \textregistered~V40. Each column indicates the nuisance attribute categories, orientations from 0 to 324 degrees. These categories and orientations in this figure are used in our experiments.}
    \label{fig:carcgs_grid}
\end{figure*}
\begin{figure*}[p]
    \centering
    \includegraphics[width=\columnwidth]{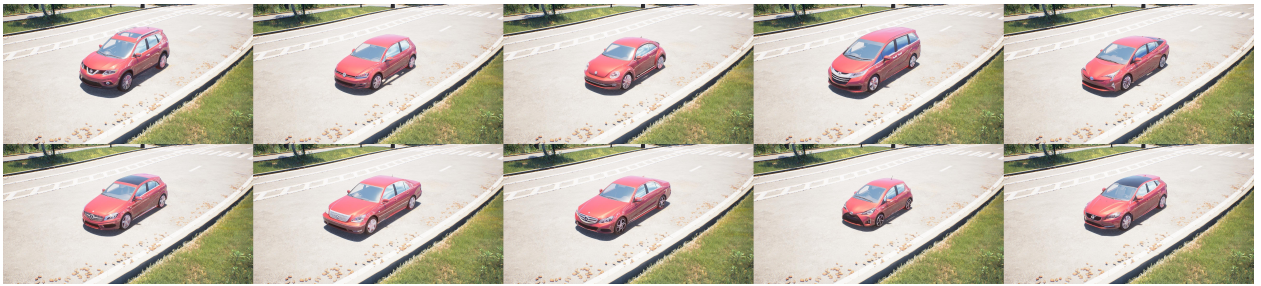}
    \caption{Sample images of ten object categories of CarsCG-Orientatoins --- Nissan Rouge\textsuperscript \textregistered, Volkswagen\textsuperscript \textregistered~Golf, 
    Volkswagen\textsuperscript \textregistered~ Beetle, Honda Odyssey\textsuperscript \textregistered, Toyota Prius\textsuperscript \textregistered, Mercedes Benz\textsuperscript \textregistered~A-Class, Lexus\textsuperscript \textregistered~LS, Mercedes Benz\textsuperscript \textregistered~E-Class, Toyota Yaris\textsuperscript \textregistered~and Volvo\textsuperscript \textregistered~V40 are shown in this figure from the top left to the bottom right. All conditions except object category are fixed  in this figure.}
    \label{fig:carcgs_category}
\end{figure*}
\begin{figure*}[p]
    \centering
    \includegraphics[width=\columnwidth]{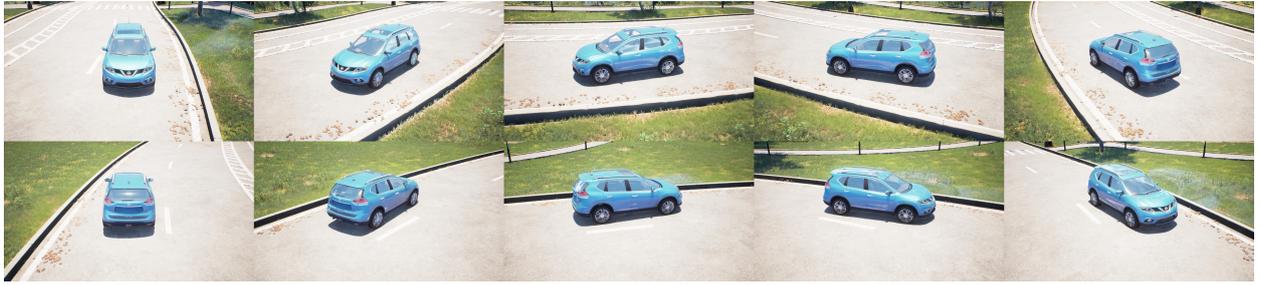}
    \caption{Sample images of ten orientations (condition categories) of CarsCG-Orientations.  Ten orientations from 0 to 324 degrees are displayed from the top left to the bottom right. 
    All conditions except orientation are fixed in this figure.}
    \label{fig:carcgs_orientation}
\end{figure*}
\begin{figure*}[p]
    \centering
    \includegraphics[width=0.6\columnwidth]{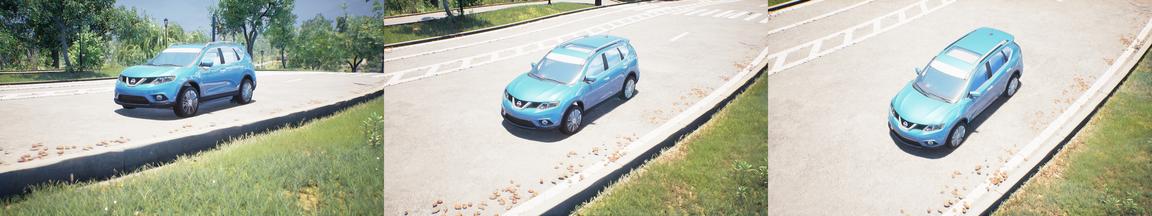}
    \caption{Sample images of the three elevation angles (10, 15, 30 degrees) of CarsCG-Orientations. Left figure is the image whose elevation angle is 10 degrees. Middle figure is the image whose elevation angle is 15 degrees. Right figure is the image whose elevation angle is 30 degrees.}
    \label{fig:carcgs_elevation}
\end{figure*}
\begin{figure*}[p]
    \centering
    \includegraphics[width=\columnwidth]{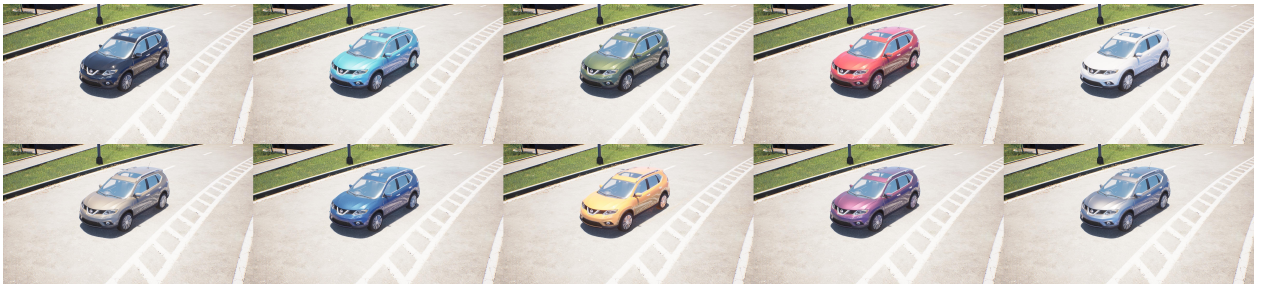}
    \caption{Sample images of cars painted in ten colors of CarsCG-Orientations. There are images painted in black, light blue, green, red, white, beige, dark blue, orange, plum, and silver from the top left to the bottom right.}
    \label{fig:carcgs_color}
\end{figure*}
\begin{figure*}[p]
    \begin{tabular}{c}
        \subfigure[\vspace{2mm}Sample images of each location from elevation angle of 10 degrees that CarsCG-Orientations has.]{\label{fig:carcgs_location}
            \includegraphics[width=\columnwidth]{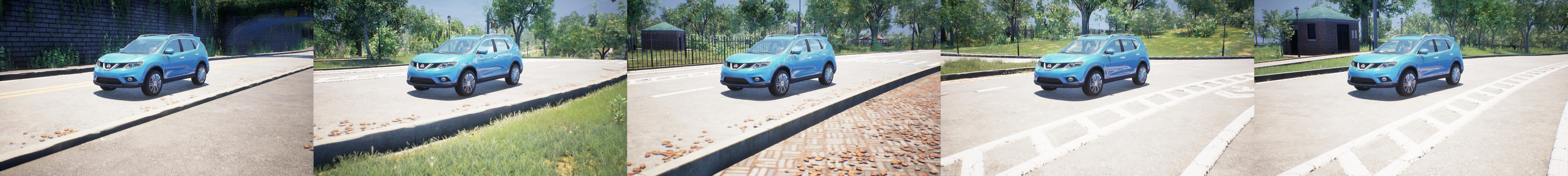}}\\
        \subfigure[\vspace{2mm}Sample images of each location from elevation angle of 15 degrees that CarsCG-Orientations has.]{\label{fig:carcgs_location2}            \includegraphics[width=\columnwidth]{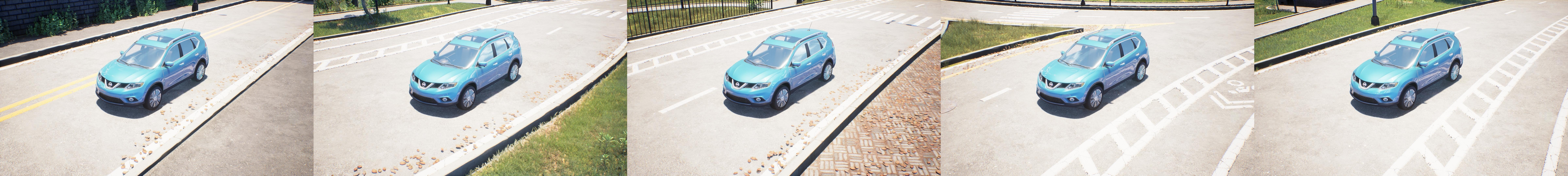}}\\
    \end{tabular}
\caption{(a)~is images of a car placed in five different locations taken from elevation angle of 10 degrees. (b)~is images taken from elevation angle of 15 degrees. There are differences in texture of road and background.}
\label{fig:Location}
\end{figure*}
\begin{figure*}[p]
    \centering
    \includegraphics[width=0.6\columnwidth]{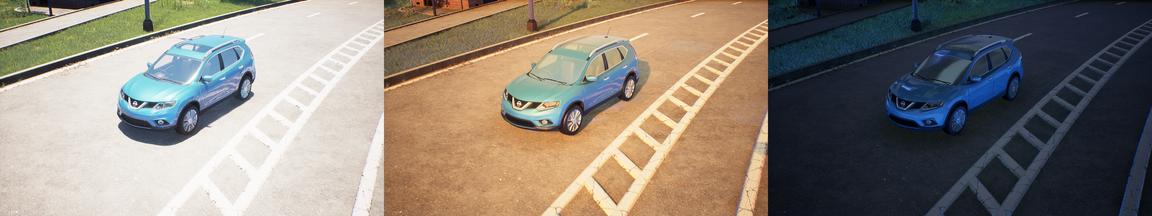}
    \caption{Sample images of each time slot of CarsCG-Orientations. Left figure is an image of car taken in the daytime. Middle figure shows an image of a car taken in the twilight. The color of the car is different from that in the left and right ones. It is caused by the twilight sunlight condition. Right figure shows an image of car taken at midnight. The color of the car is also different from left and middle ones. } \label{fig:carcgs_time}
\end{figure*}

\subsection{CarsCG-Orientations}
CarsCG-Orientations is a new dataset that consists of images of ten models of cars in various conditions rendered by Unreal Engine version 4.25.3; this dataset is publicly available at \url{http://dataset.jp.fujitsu.com/data/carscg/index.html}.
The conditions consist of ten orientations, three elevations, ten body colors, five locations and three time slots. 
Fig.~\ref{fig:carcgs_grid} shows the all  car models (categories) and orientations (conditions) in the grid form. The details of these are as follows. 
\begin{itemize}
\item Categories: CarsCG-Orientations dataset consists of  images of the following cars --- Nissan Rouge\textsuperscript \textregistered, Volkswagen\textsuperscript \textregistered~Golf, Volkswagen\textsuperscript \textregistered~Beetle, Honda Odyssey\textsuperscript \textregistered, Toyota Prius\textsuperscript \textregistered, Mercedes Benz\textsuperscript \textregistered~A-Class, Lexus\textsuperscript \textregistered~LS, Mercedes Benz\textsuperscript \textregistered~E-Class, Toyota Yaris\textsuperscript \textregistered and Volvo\textsuperscript \textregistered~V40 (See Fig.~\ref{fig:carcgs_category}). 
We used the whole car models as  categories $\gC$.
Therefore the number of  categories is $\#(\gC)=10$ in the experiments 
conducted in this study. 
\item Orientations: During the rendering process, the virtual camera (camera actor) was rotated around the yaw axis of each car from 0 to 324 degrees in units of 36 degrees. 
Therefore, each car model appears in the images with ten different azimuth orientations. 
All orientations are shown in Fig.~\ref{fig:carcgs_orientation}. 
We used the whole orientations as conditions $\gN$. 
Thus the number of the conditions  is $\#(\gN)=10$ in the experiments 
conducted in this study.
\end{itemize}
To create variety of samples for each combination of the  categories (car models) and conditions (orientations), we added other conditions as follows.
\begin{itemize}
\item Elevations: The virtual camera was located at three elevation angles, namely, 10, 15, and 30 degrees, during the rendering process. 
Sample images taken from each angle are shown in Fig.~\ref{fig:carcgs_elevation}. 
\item Body colors: Each car model is rendered with ten colors, namely, black, light blue, green, red, white, beige, dark blue, orange, plum, and silver by using  Automotive Materials (a library for Unreal Engine). 
Fig.~\ref{fig:carcgs_color} shows sample images of Nissan Rouge rendered with these colors. 
\item Locations: We used a sample environment of an urban park contained in City Park Environment Collection. 
We chose five locations from the sample environment 
and modified them for our experiments. Sample images taken at each location are shown in Fig.~\ref{fig:Location}.
\item Time slots: We used Ultra Dynamic Sky 3D model set to synthesize the three different times slots, namely, daytime, twilight, and night. 
Fig.~\ref{fig:carcgs_time} shows the samples of these three time slots.
\end{itemize}
The number of images and the image size are as follows. 
\begin{itemize}
    \item Number of images and image size: The total number of images of this dataset is 45K $=$ 10 (categories) $\times$ 10 (orientations) $\times$ 3 (elevations) $\times$ 10 (body colors) $\times$ 5 (locations) $\times$ 3 (time slots). 
The images are rendered in 3840$\times$2160 pixels and then resized to 1920$\times$1080 pixels for the sake of anti-aliasing.
\end{itemize}

 \begin{figure*}[p]
    \centering
    \includegraphics[width=0.8\columnwidth]{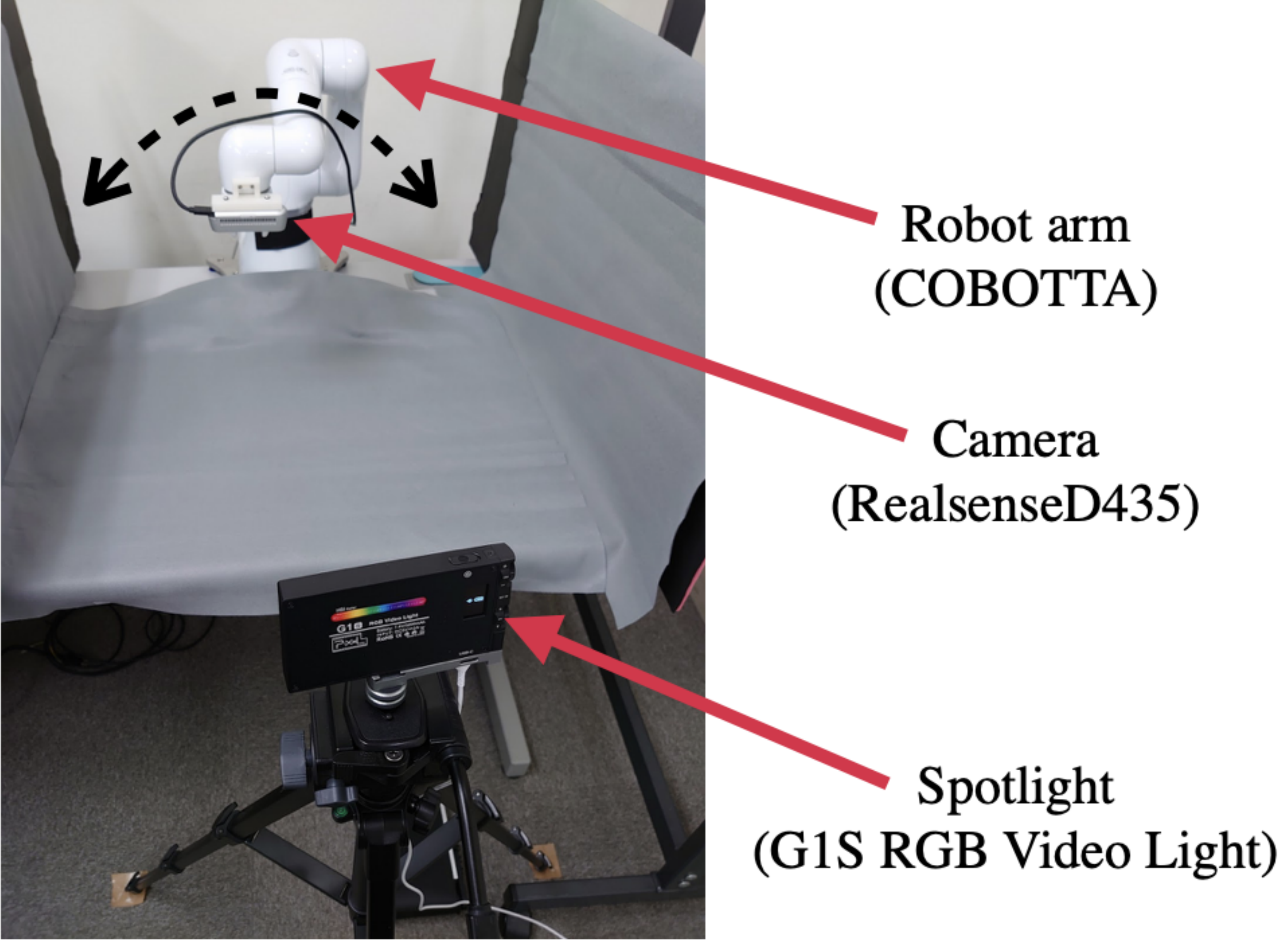}
    \caption{Robotic image capture system for MiscGoods-Illuminations. Dashed bidirectional arrow indicates the robot motion.}
    \label{fig:robot}
\end{figure*}
\begin{figure*}[p]
    \centering
    \includegraphics[width=0.5\columnwidth]{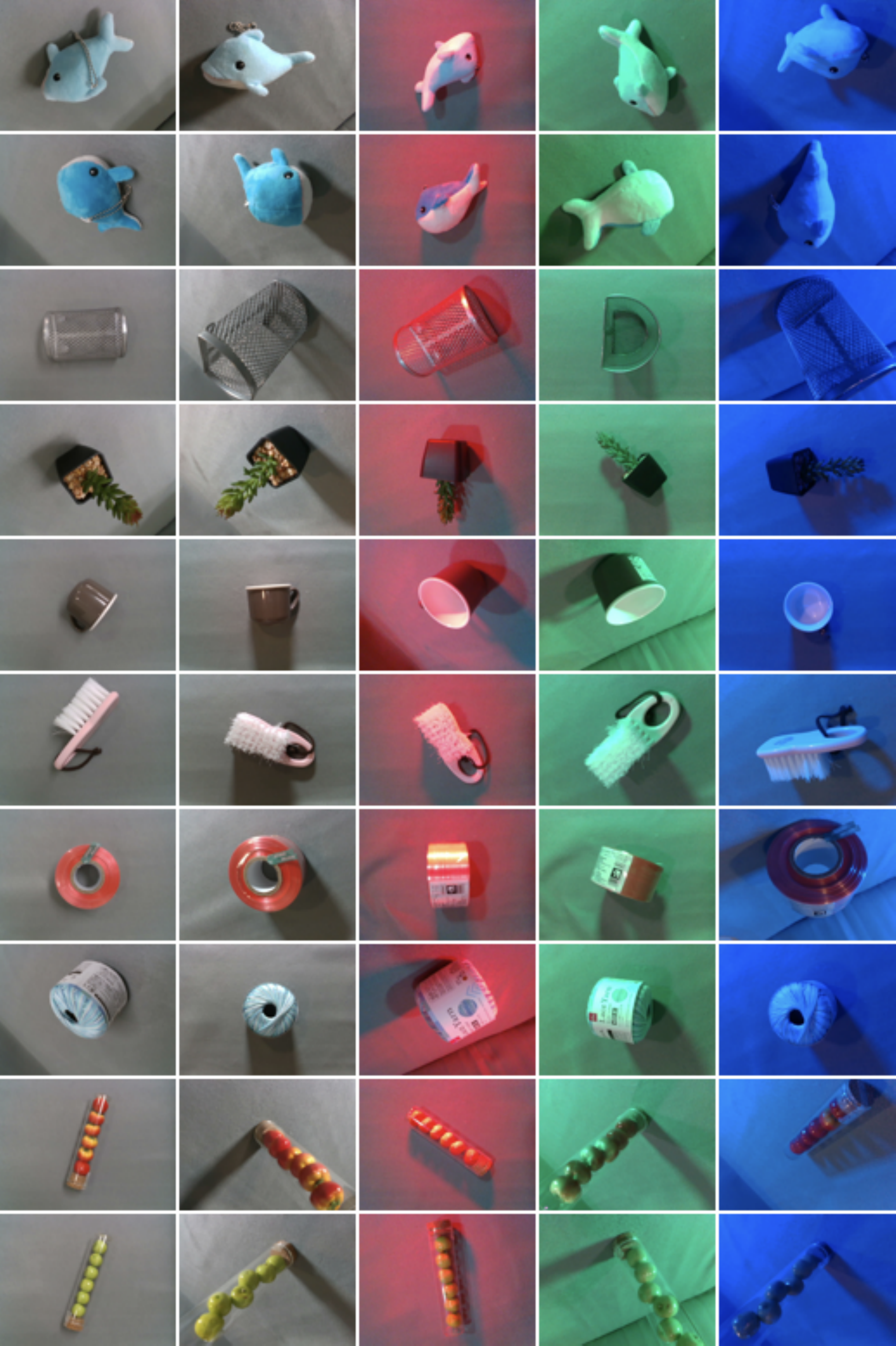}
    \caption{Sample images of each object category and illumination condition of MiscGoods-Illuminations are shown in this figure. Each row indicates object categories --- stuffed dolphin, stuffed whale, metal basket, imitation plant, cup, cleaning brush, winding tape, lace yarn, bottled imitation tomatoes, and bottled imitation green apples. Each column indicates the condition categories, illumination conditions --- ceiling light, white spotlight, red spotlight, green spotlight and blue spotlight. These five categories from the top and five illumination conditions are used as object categories and condition categories in our experiments.}
    \label{fig:daiso_grid}
\end{figure*}
\begin{figure*}[p]
    \centering
    \includegraphics[width=0.3\columnwidth]{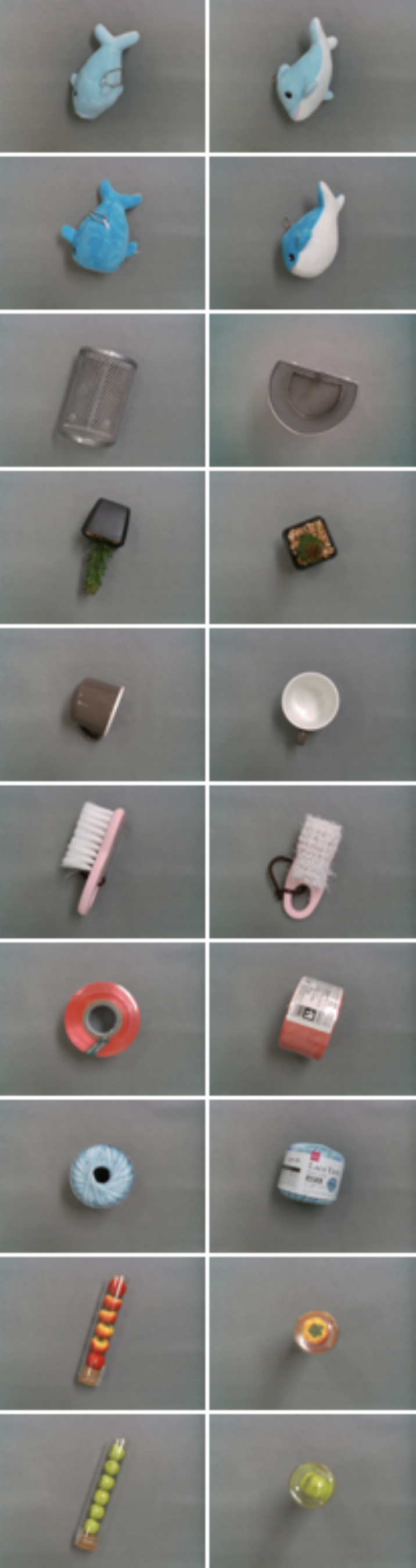}
    \caption{Sample images of MiscGoods-Illuminations with two ways of object placement. Each object has these two aspects as condition. The shapes of these objects in an image are changed by the way of object placement.}
    \label{fig:daiso_aspect}
\end{figure*}
\begin{figure*}[p]
    \centering
    \includegraphics[width=0.98\columnwidth]{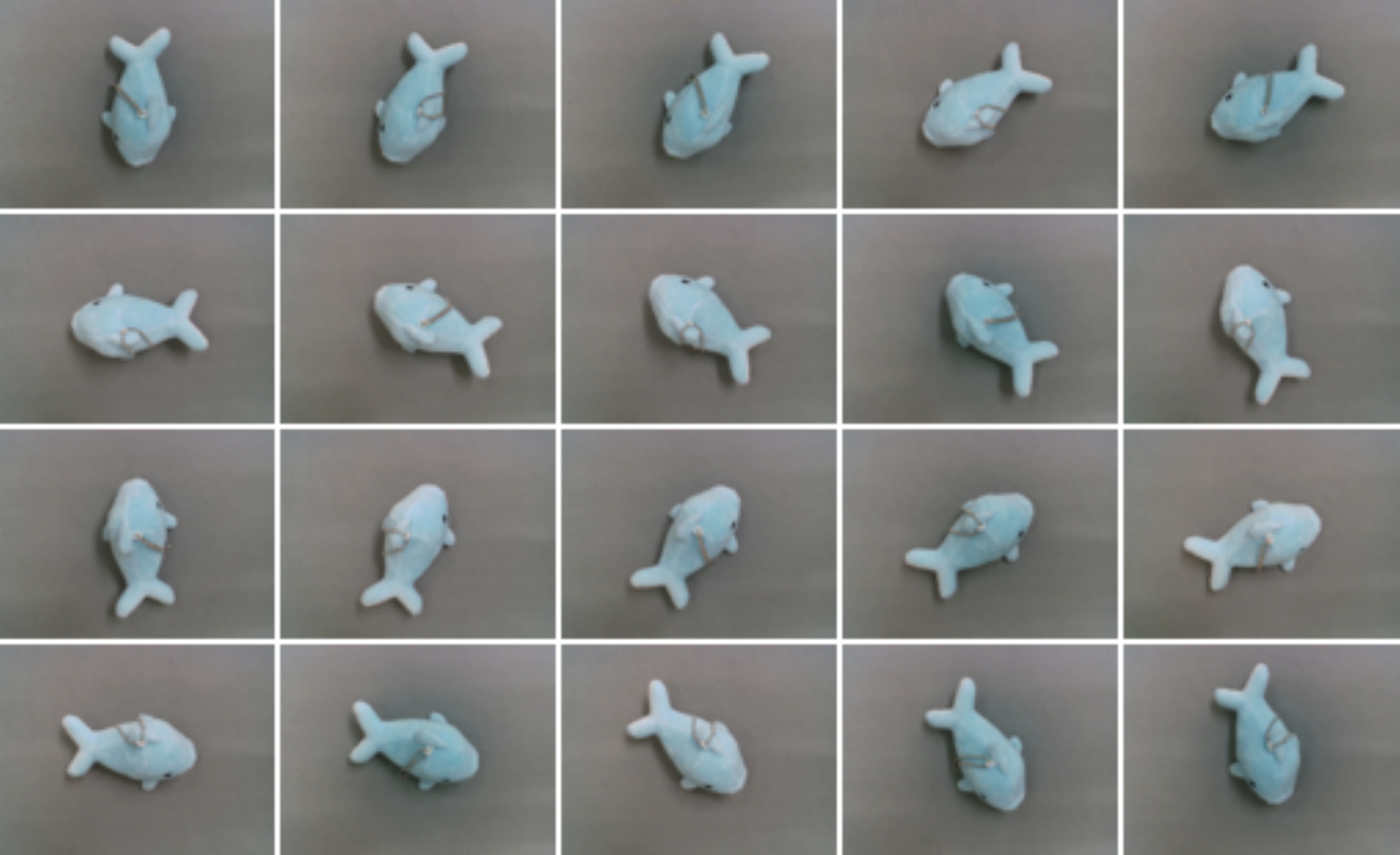}
    \caption{Sample images of each orientation of MiscGoods-Illuminations. 20 orientations from 0 to 342 degrees that the dataset has are shown in this figure from the top left to the bottom right.}
    \label{fig:daiso_orientation}
\end{figure*}
\begin{figure*}[p]
    \centering
    \includegraphics[width=0.98\columnwidth]{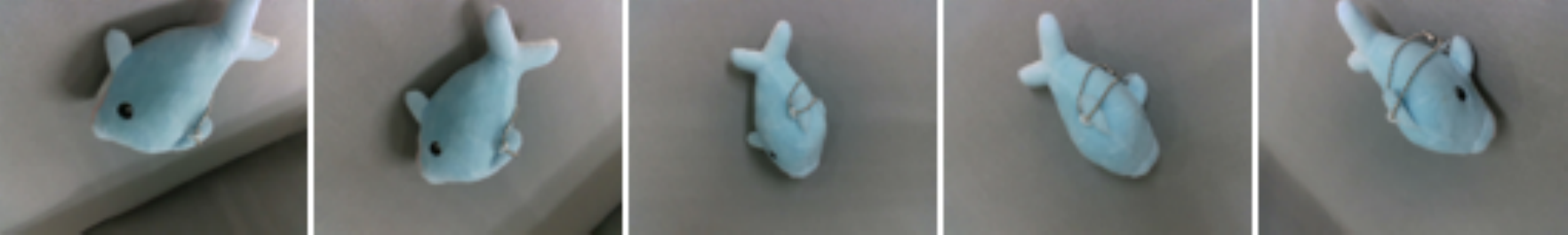}
    \caption{Sample images from each camera angles of MiscGoods-Illuminations. There are five angles in the dataset. The postures were defined so that the acquired image shows the entire object pose. These five camera angles are related to postures of robot arm that the camera is connected.}
    \label{fig:daiso_angle}
\end{figure*}

\subsection{MiscGoods-Illuminations}
MiscGoods-Illuminations is a subset of DAISO-10, a novel dataset constructed for this study;  this dataset is publicly available at \url{http://dataset.jp.fujitsu.com/data/daiso10/index.html}. 
The dataset consists of images of ten physical miscellaneous goods taken with five illumination conditions, two ways of object placement, twenty object orientations, five camera angles. 
Images were taken with a robot arm (Fig.~\ref{fig:robot}).
Fig.~\ref{fig:daiso_grid} shows the all miscellaneous goods (categories) and illumination conditions  in the grid form. The details of these  are as follows.
\begin{itemize}
\item Categories: As shown in Fig.~\ref{fig:daiso_grid}, DAISO-10 has ten types of miscellaneous goods --- stuffed dolphin, stuffed whale, metal basket, imitation plant, cup, cleaning brush, winding tape, lace yarn, bottled imitation tomatoes, and bottled imitation green apples. 
In this study, we selected the following five miscellaneous goods from DAISO-10 as the  categories $\gC$ --- stuffed dolphin, stuffed whale, metal basket, imitation plant and cup. Therefore the number of  categories is $\#(\gC)=5$ in the experiments conducted in this study.
\item Illumination conditions: As the conditions, we created five illumination conditions (lighting conditions); 
one is created with ceiling lights, 
and the rest are with a colored spotlight. 
All illumination conditions are shown in Fig.~\ref{fig:daiso_grid}. 
For spotlight conditions, 
the light source (PIXEL G1S\textsuperscript \texttrademark~RGB Video Light) was placed 23 cm in front of the object (See Fig.~\ref{fig:robot}). 
The parameters of the light source were H217/S141$=$8500k (white light), H0/S100 (red light), H120/S100 (green light), and H240/S100 (blue light). 
These parameters were set so that the condition of the illumination makes a sufficient difference in the learning experiments. 
We used whole illumination conditions $\gN$. Thus the number of the conditions is $\#(\gN)=5$ in the experiments conducted in this study.
\end{itemize}
As we did for CarCGs-Orientations, 
we added other conditions to 
create variety of samples for each combination of the  categories and illumination conditions as follows. 
\begin{itemize}
\item Object poses (ways of object placement and orientations): In this dataset,we placed each object in two representative ways of object placement for each lighting condition. 
Fig.~\ref{fig:daiso_aspect} shows the two ways of object placement of all objects.
For additional diversity, we rotated the object every 18 degrees from 0 to 342 degrees (Fig.~\ref{fig:daiso_orientation}). 
In total, there are 40 patterns in object pose conditions. 
\item Camera angles: To capture 
the images automatically, we created a 
robotic image capture system (see Fig.~\ref{fig:robot}). 
A camera (Intel\textsuperscript \textregistered~Realsense D435) was attached  to a robot arm (COBOTTA\textsuperscript \textregistered),  
and the system captured images from five 
camera angles for each lighting and object pose condition (Fig.~\ref{fig:daiso_angle}). 
The postures were defined so that the acquired image shows the entire object pose. 
The series of operations from robot control to image acquisition is automated by utilizing ROS kinetic.
\end{itemize}
The number of images and the image size are as follows. 
\begin{itemize}
\item Number of images and image size: The number of images of whole dataset is 10K $=$ 10 (categories) $\times$ 5 (illuminations) $\times$ 2 (ways of object placement) $\times$ 20 (orientations) $\times$ 5 (camera angles), and each image size is 640 $\times$ 480 pixels.
\end{itemize}

\setcounter{figure}{0}
\setcounter{table}{0}
\setcounter{equation}{0}

\section{Details of experiments} \label{sec:details_exp}
\begin{figure}[p]
    \centering
    \includegraphics[width=0.3\columnwidth]{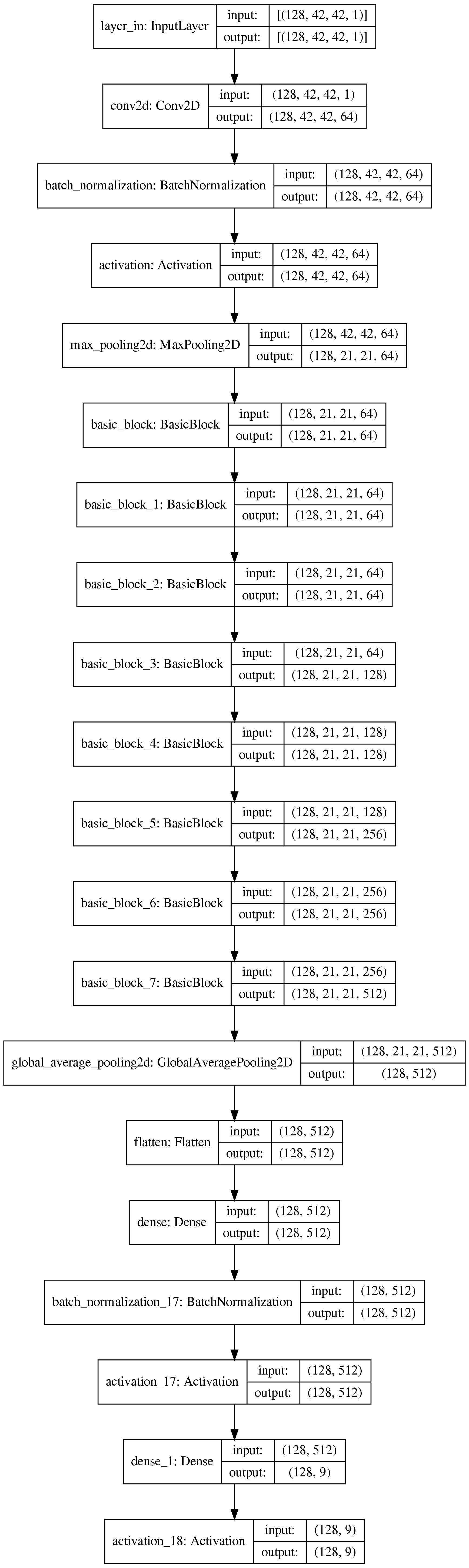}
    \caption{This diagram shows the whole architecture of our implementation of ResNet-18. 
The numbers in this diagram represent batchsize, height of image, width of image and channels. 
    Therefore they change depending on the dataset. 
    Current numbers correspond to MNIST-Positions. 
    For instance, the numbers on the top of the diagram means $(batch size, height, width, channels)=(128, 42, 42, 1)$. 
    Conv2D, Dense and BasicBlock mean a convolutional layer, a fully connected layer and a basic building block of ResNet, respectively.}
    \label{fig:archi}
\end{figure}

\begin{figure}[p]
    \centering
    \includegraphics[width=0.3\columnwidth]{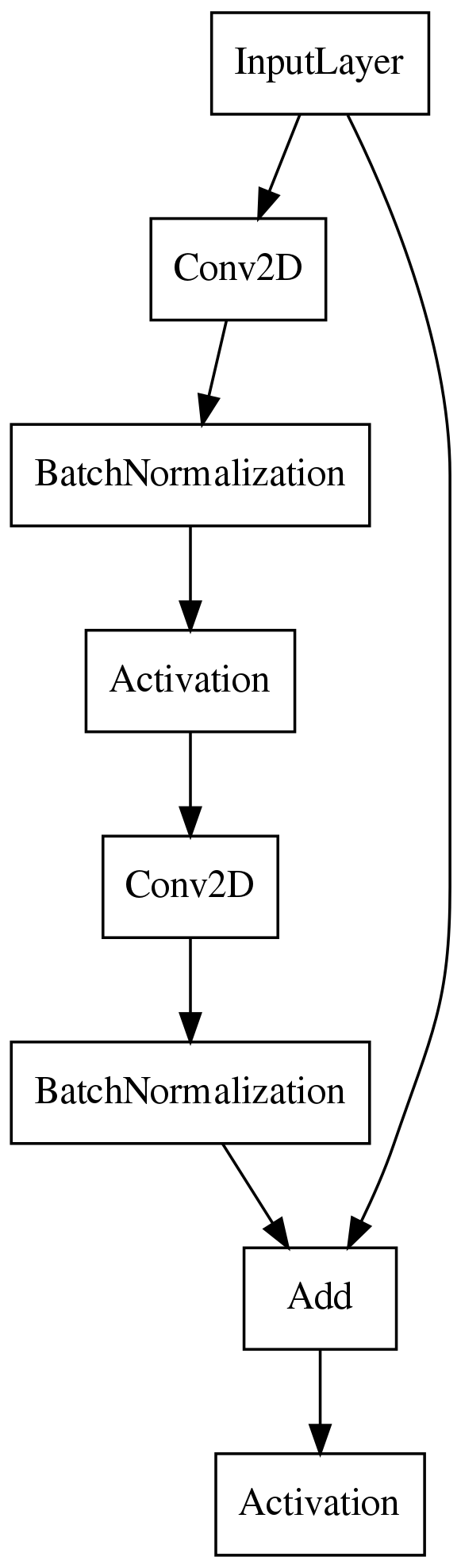}
    \caption{This diagram shows the architecture of BasicBlock in ResNet-18. Conv2D and Add mean a convolutional layser and a layer that simply add the two input values.}
    \label{fig:basic}
\end{figure}

\label{app:networks}

ResNet-18~\cite{He2016resnet} is adopted as the network for all experiments. 
The source codes are implemented based on Python v3.6.9, using TensorFlow v2.5.0 and NumPy v1.19.5. 
They are included in the zip file (/source\_code).
The whole network architecture is shown in Fig~\ref{fig:archi} and Fig~\ref{fig:basic}. 
All neurons 
employ the rectified linear function $g(z) = \max \{0, z  \}$ and satisfy $a^{nm}(\rvx) \ge 0 $. 
Glorot uniform initializer~\cite{Glorot10understandingthe} is adopted for the network weights initialization for all experiments.
We use BatchNormalization  to standardize the inputs to a layer for each mini-batch. We use it for stabilizing the learning process and reducing the number of training epochs.
We do not use any data augmentations. Invariance loss is applied to the last fully-connected layer “activation\_17" with $512$ neurons shown in Fig.~\ref{fig:archi}. Adam~\cite{Diederik2015Adam} is employed as the optimization algorithm. 
The cross-entropy loss is employed as the loss  $L$. 
The pixels of images are normalized within $0$ to $1$ as a preprocessing for all datasets.
The epoch size and batch size are confirmed to produce reasonable accuracy in the baseline case for each dataset and we employ the same values for all experiments with the same dataset.
For example, we use $100$ and $256$ as epoch size and batch size, respectively, for MNIST-Positions. 
The values of hyper-parameters 
are summarized in Table~\ref{tab:parameters}.
\begin{table}[ht]
\begin{center}
\caption{Hyper-parameters used for each dataset}
\begin{tabular}{lccccc}
\hline
dataset & epoch size & preprocessing & weights initialization & batch size\\
\hline\hline
MNIST-Positions & 100 & divide by 255 & Glorot uniform initializer & 256\\
iLab-Orientations & 100 & divide by 255 & Glorot uniform initializer & 256 \\
CarsCG-Orientations & 100 & divide by 255 & Glorot uniform initializer & 32\\
MiscGoods-Illuminations & 100 & divide by 255 & Glorot uniform initializer & 32\\
\hline
\end{tabular}
\label{tab:parameters}
\end{center}
\end{table}
We have employed four Tesla V100 GPUs for the experiments.
The preparation of training dataset $\gD_{\rm train}^{\rm (InD)}$, InD validation dataset $\gD^{(\rm InD)}_{\rm val}$, and OoD dataset $\gD^{(\rm OoD)}$ has been conducted as follows.

\begin{itemize}
\item {MNIST-Positions:} We use images of the original MNIST-Positions with image size of $42\times42$ pixels.
InD dataset and OoD dataset are prepared in the way described in Section~\ref{subsec:Creating spurious correlation}.The number of train dataset is $\#(\gD_{\rm train}^{\rm (InD)}) = 54000$. We use $\#(\gD^{(\rm InD)}_{\rm val}) = 8000$ for InD validation dataset. 
The number of OoD dataset is $\#(\gD^{(\rm OoD)}) =  8000$.

\item {iLab-Orientations: }We reize the images to $64 \times 64$ pixels.
InD dataset and OoD dataset are prepared in the way described in Section~\ref{subsec:Creating spurious correlation}.The number of train dataset is $\#(\gD_{\rm train}^{\rm (InD)}) = 18000$. We use $\#(\gD^{(\rm InD)}_{\rm val}) = 8000$ for InD validation dataset. 
The number of OoD dataset is $\#(\gD^{(\rm OoD)}) =  8000$.

\item {CarsCG-Orientations:} We resize the images to $224\times224$ pixels.  
InD dataset and OoD dataset are prepared in the way described in Section~\ref{subsec:Creating spurious correlation}.The number of train dataset is $\#(\gD_{\rm train}^{\rm (InD)}) = 3400$. We use $\#(\gD^{(\rm InD)}_{\rm val}) = 450$ for InD validation dataset. 
The number of OoD dataset is $\#(\gD^{(\rm OoD)}) =  800$.

\item {MiscGoods-Illuminations:} We resize the images to $224\times224$ pixels. 
InD dataset and OoD dataset are prepared in the way described in Section~\ref{subsec:Creating spurious correlation}.The number of train dataset is $\#(\gD_{\rm train}^{\rm (InD)}) = 800$. We use $\#(\gD^{(\rm InD)}_{\rm val}) = 200$ for InD validation dataset.
The number of OoD dataset is $\#(\gD^{(\rm OoD)}) =  400$.
\end{itemize} 
\section{Additional results of experiments}

\label{app:results}
\begin{figure*}[ht]
\centering
    \begin{tabular}{cccc}
        \subfigure[MNIST]{\label{fig:low_mnist_late_stop}
            \includegraphics[width=0.44\columnwidth]{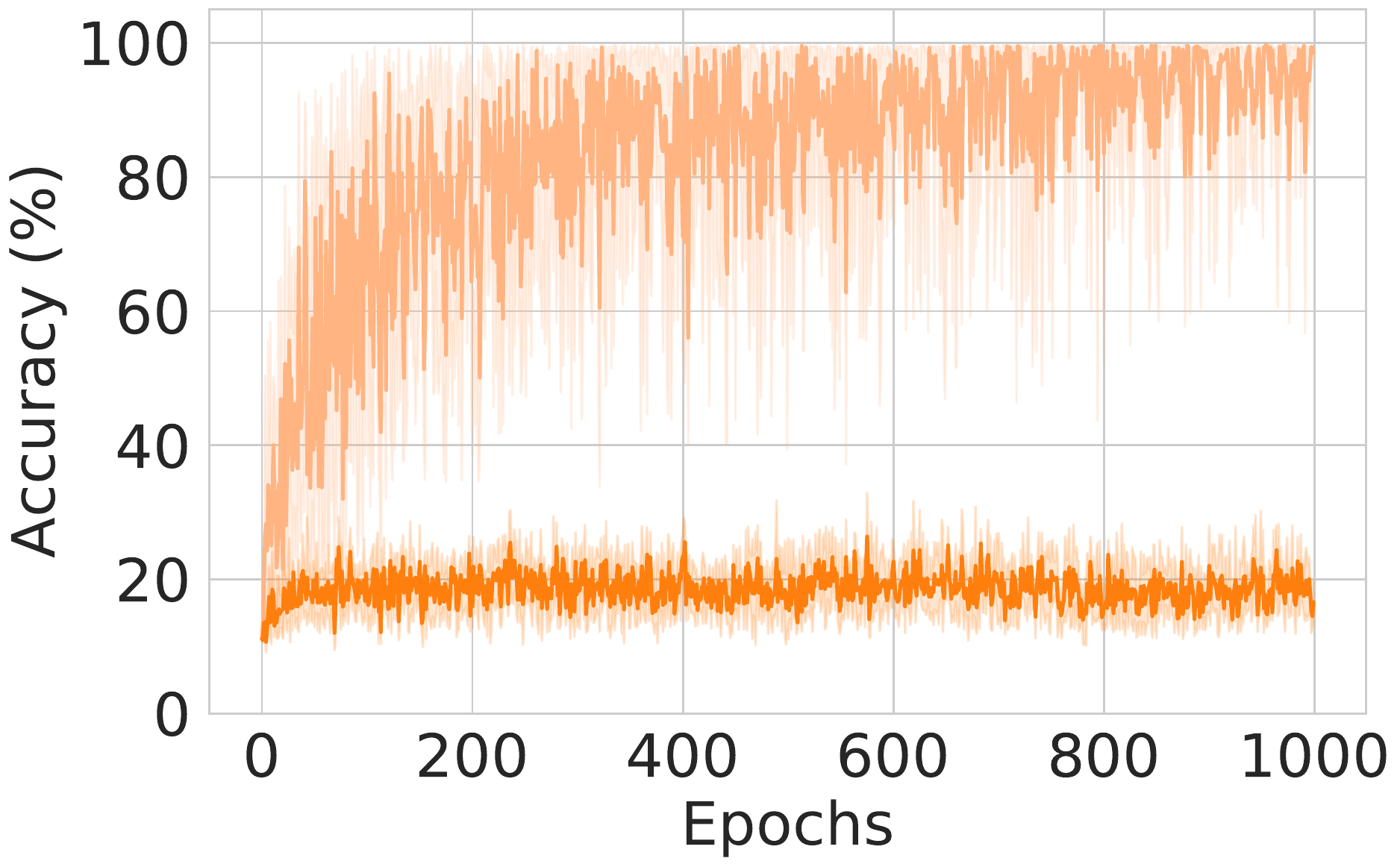}}
        \subfigure[iLab]{\label{fig:low_ilab_late_stop}
            \includegraphics[width=0.44\columnwidth]{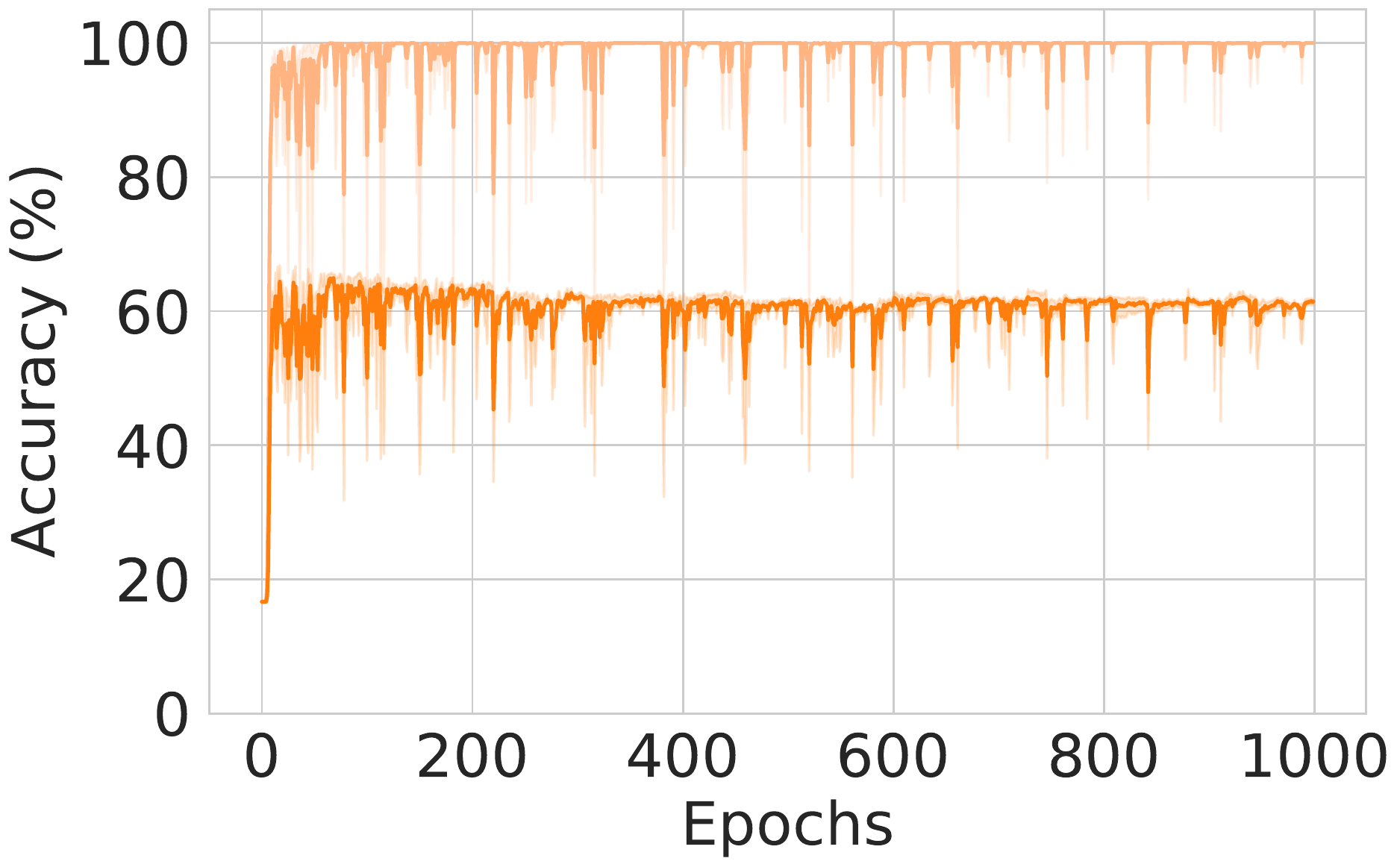}}\\
        \subfigure[CarsCG]{\label{fig:low_carcgs_late_stop}
            \includegraphics[width=0.44\columnwidth]{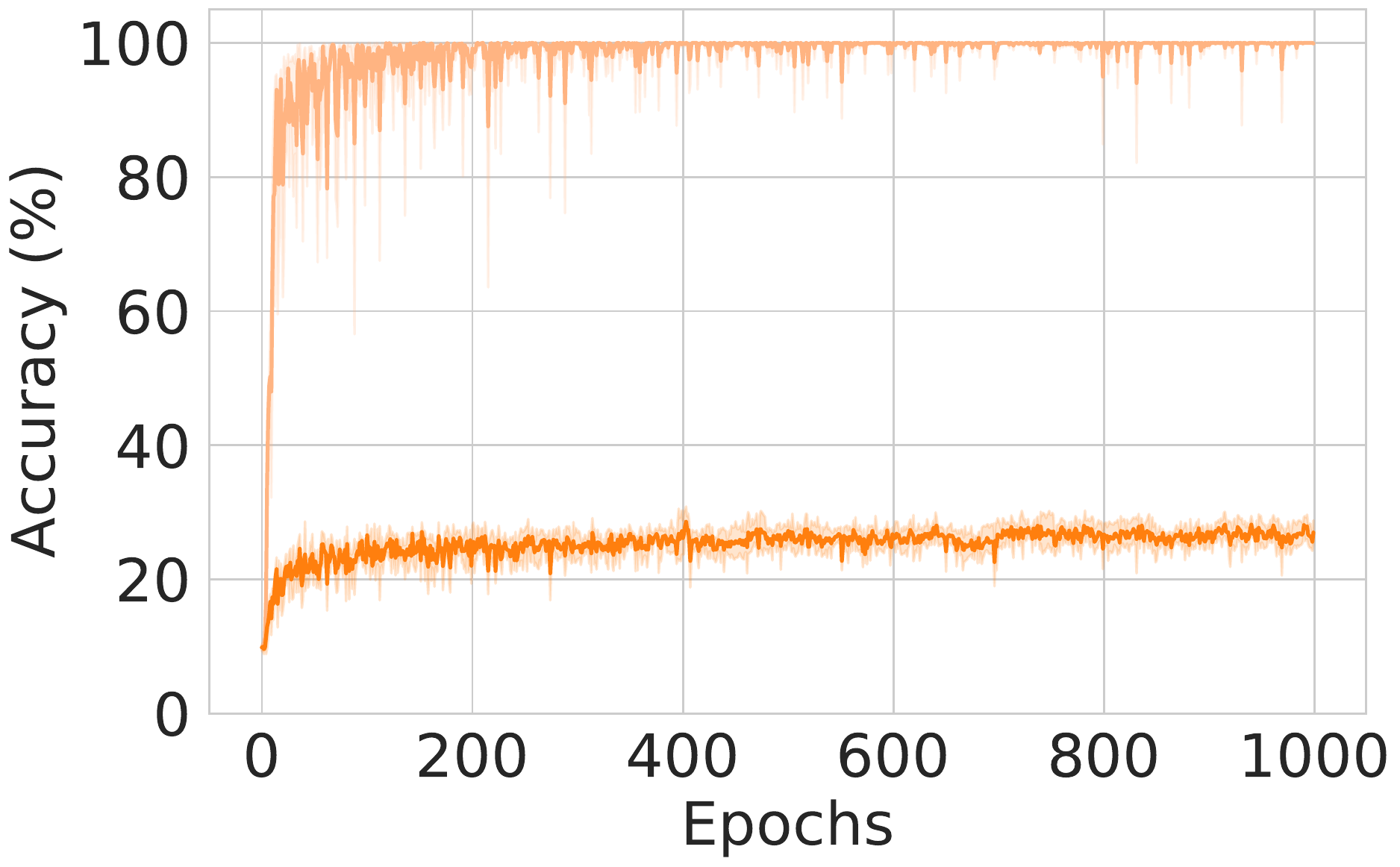}}
        \subfigure[MiscGoods]{\label{fig:low_daiso_late_stop}
            \includegraphics[width=0.44\columnwidth]{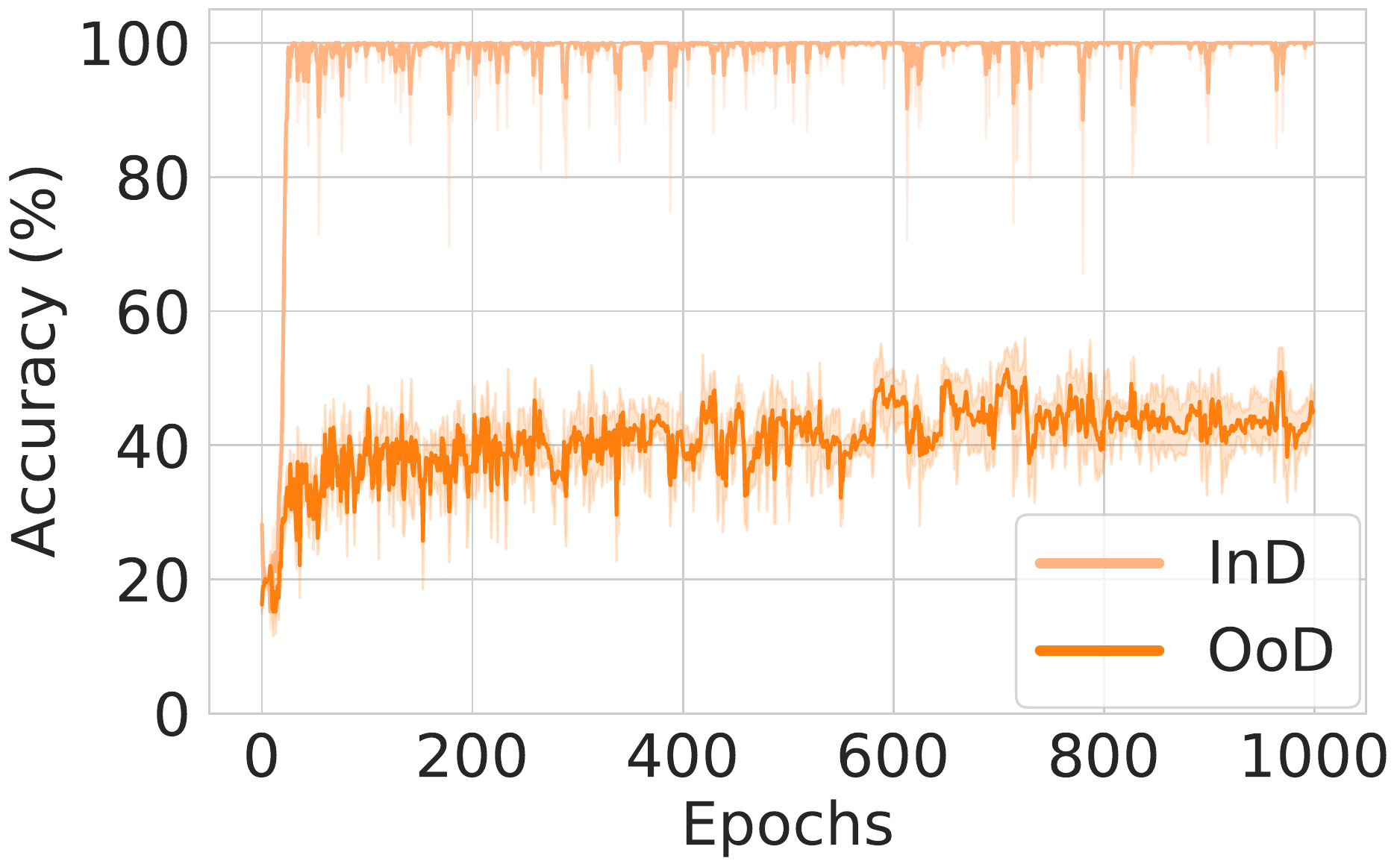}}
    \end{tabular}
\caption{Late-stopping with low InD data diversity}
\label{fig:low_late_stop_curves}
\end{figure*}

\begin{figure*}[ht]
    \begin{tabular}{cccc}
        \subfigure[MNIST]{\label{fig:medium_mnist_late_stop}
            \includegraphics[width=0.46\columnwidth]{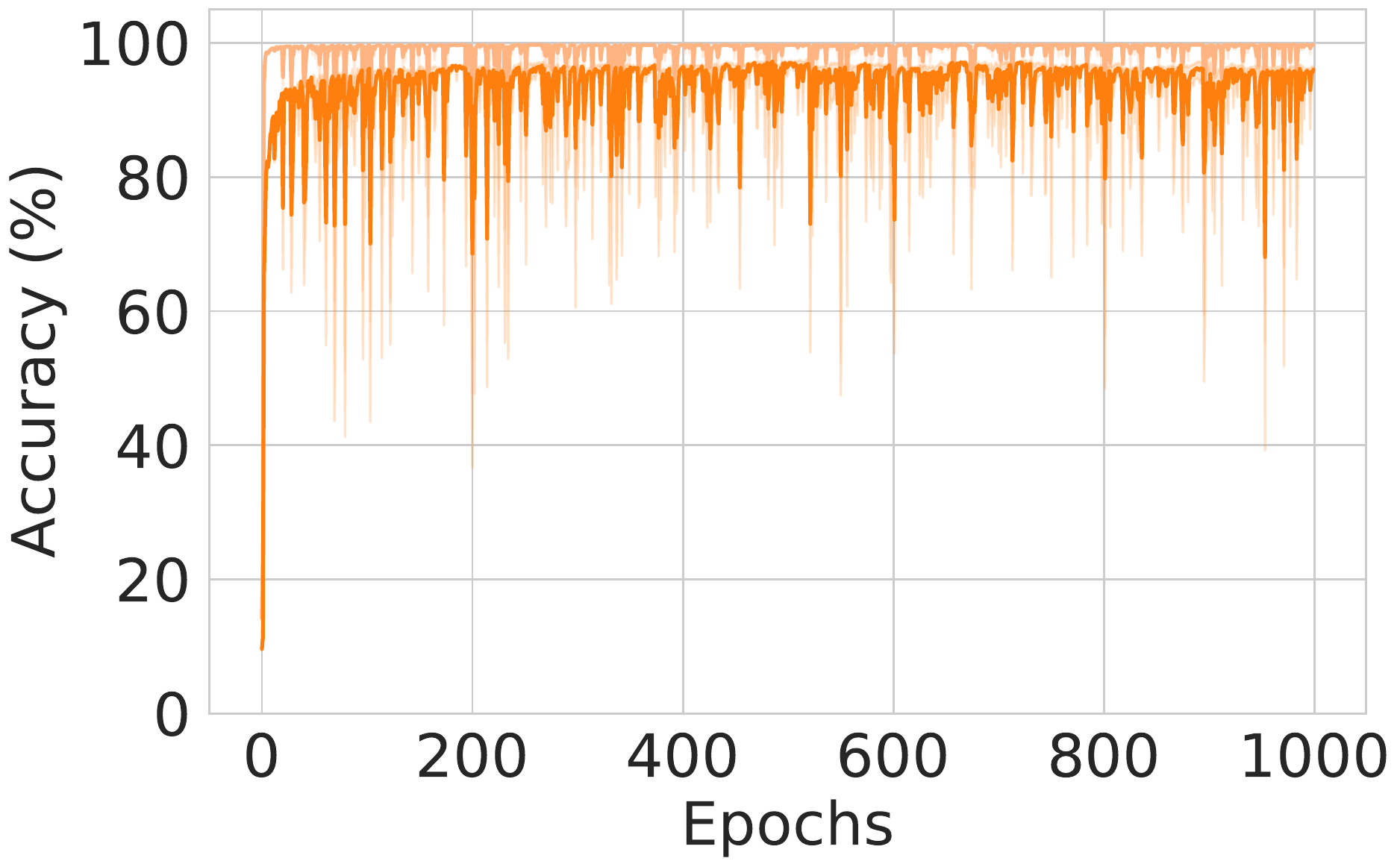}}
        \subfigure[iLab]{\label{fig:medium_ilab_late_stop}
            \includegraphics[width=0.46\columnwidth]{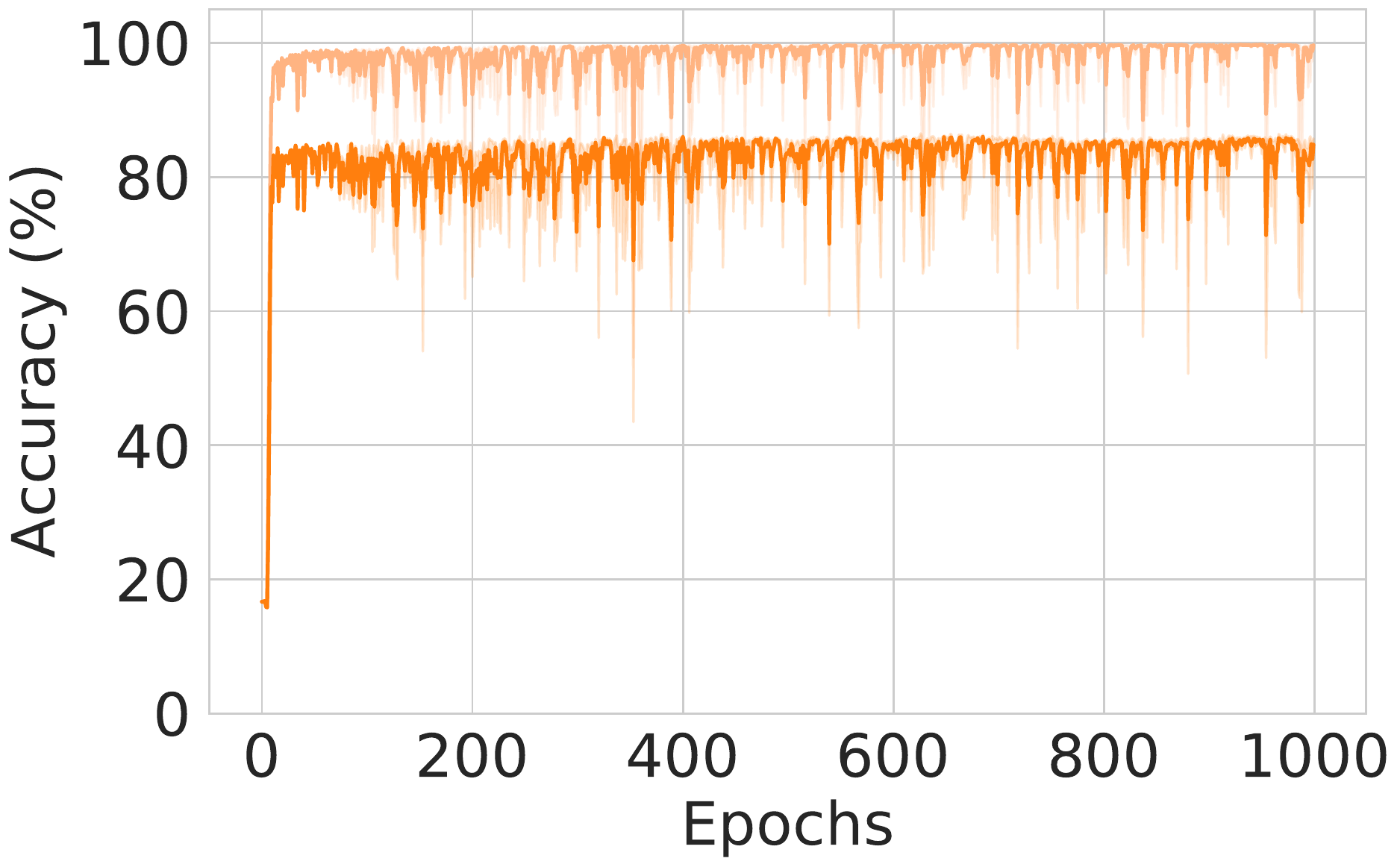}}\\
        \subfigure[CarsCG]{\label{fig:medium_carcgs_late_stop}
            \includegraphics[width=0.46\columnwidth]{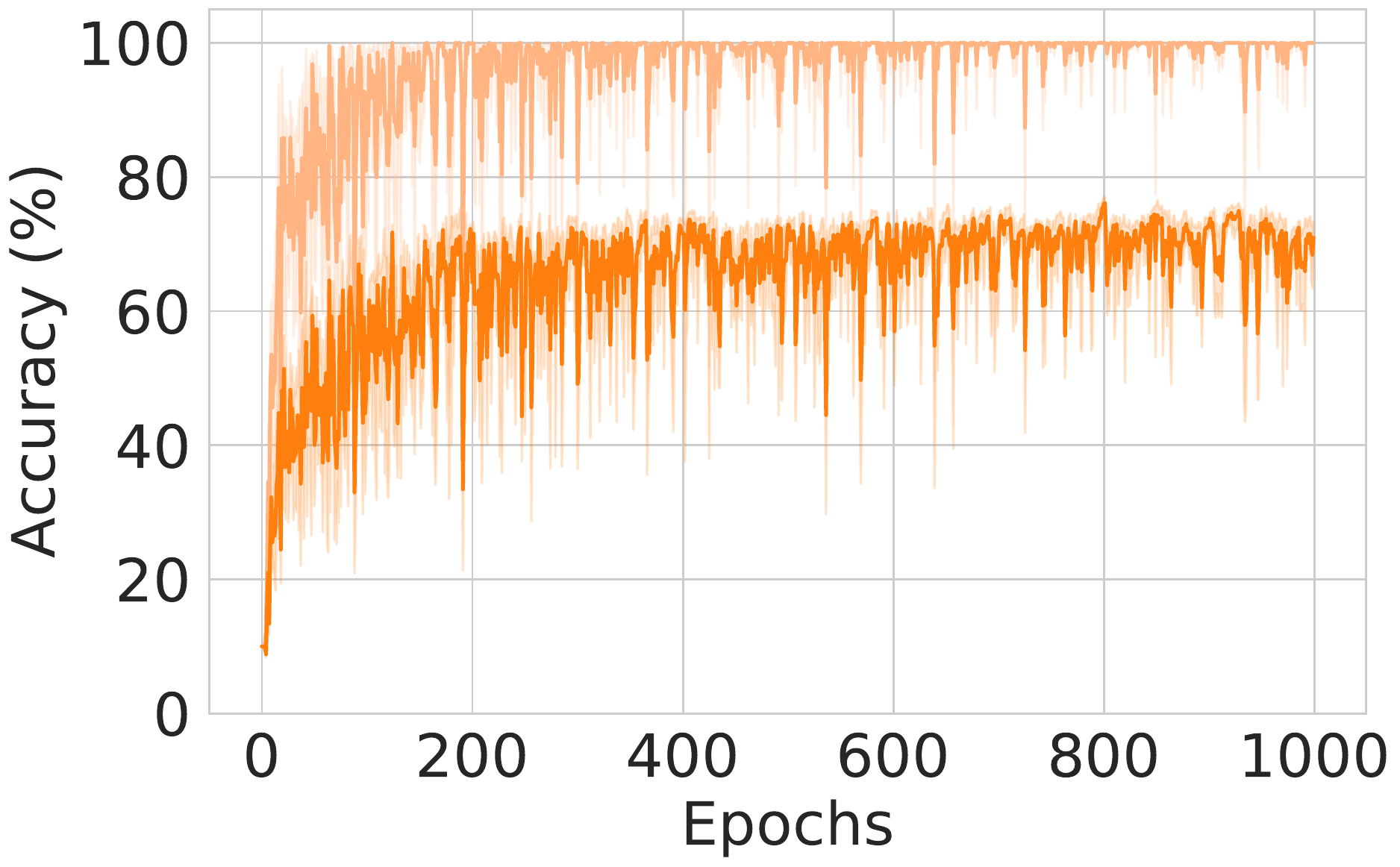}}
        \subfigure[MiscGoods]{\label{fig:medium_daiso_late_stop}
            \includegraphics[width=0.46\columnwidth]{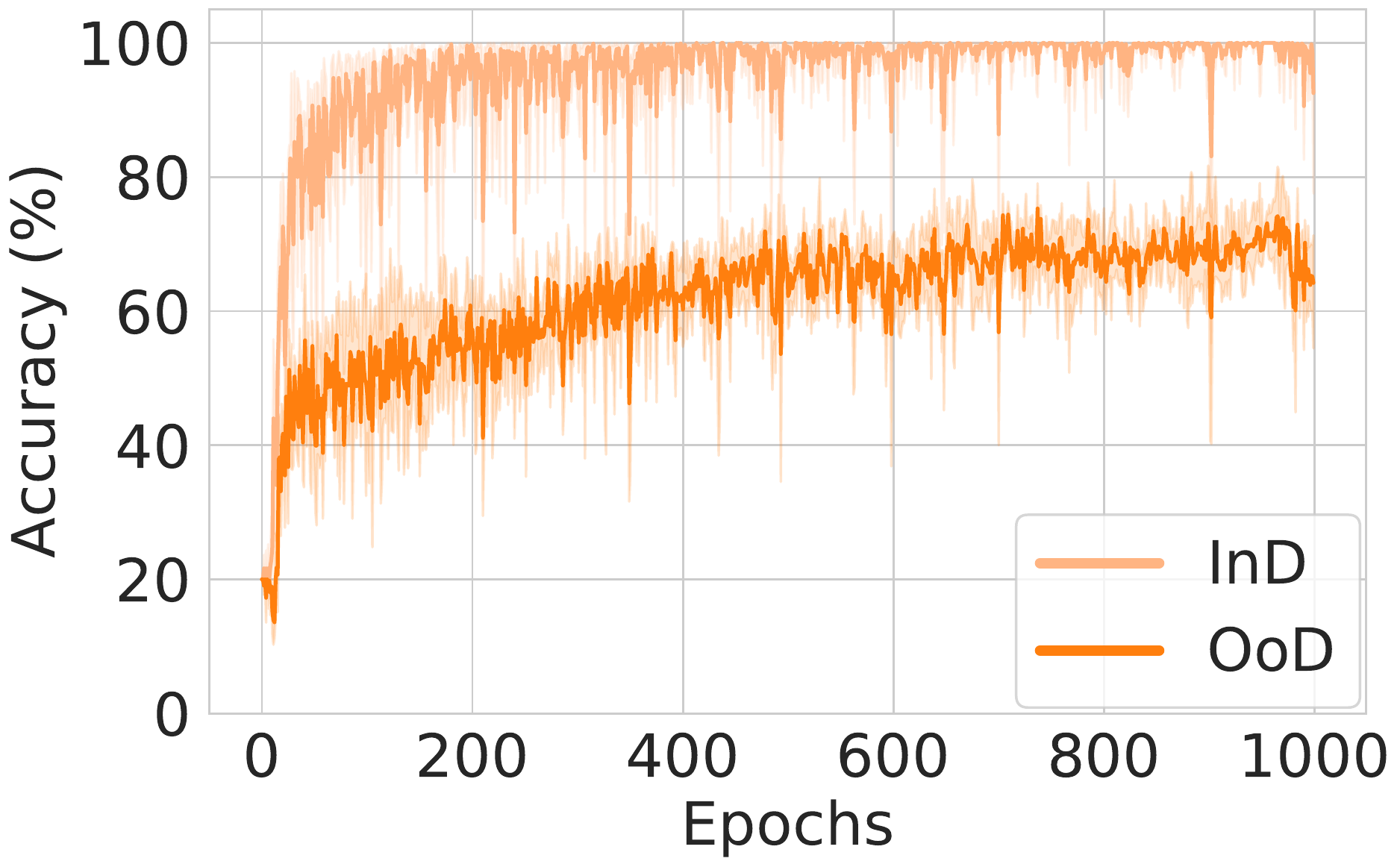}}
    \end{tabular}
\caption{Late-stopping with medium InD data diversity}
\label{fig:medium_late_stop_curves}
\end{figure*}

\begin{figure*}[ht]
    \begin{tabular}{cccc}
        \subfigure[MNIST]{\label{fig:high_mnist_late_stop}
            \includegraphics[width=0.46\columnwidth]{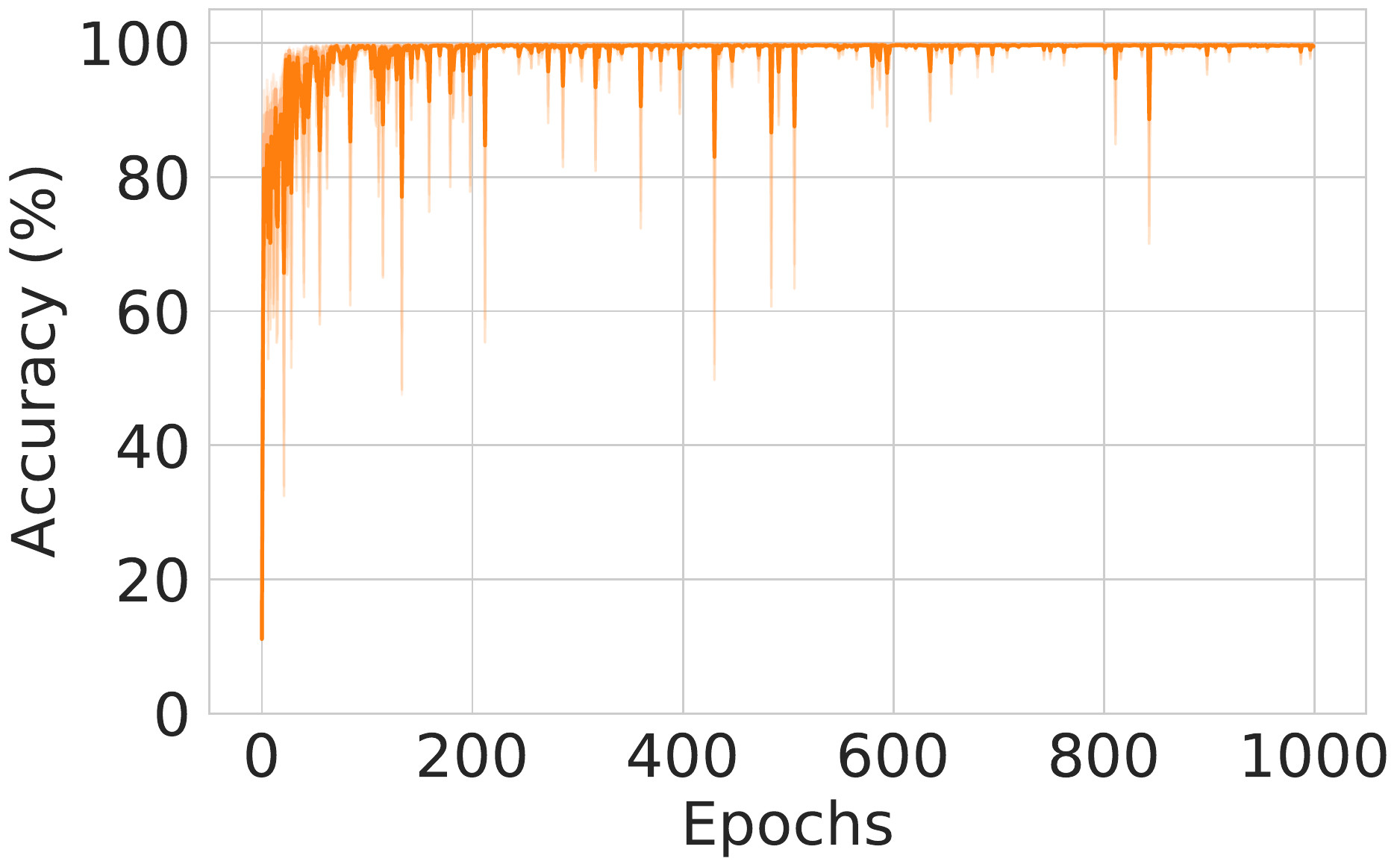}}
        \subfigure[iLab]{\label{fig:high_ilab_late_stop}
            \includegraphics[width=0.46\columnwidth]{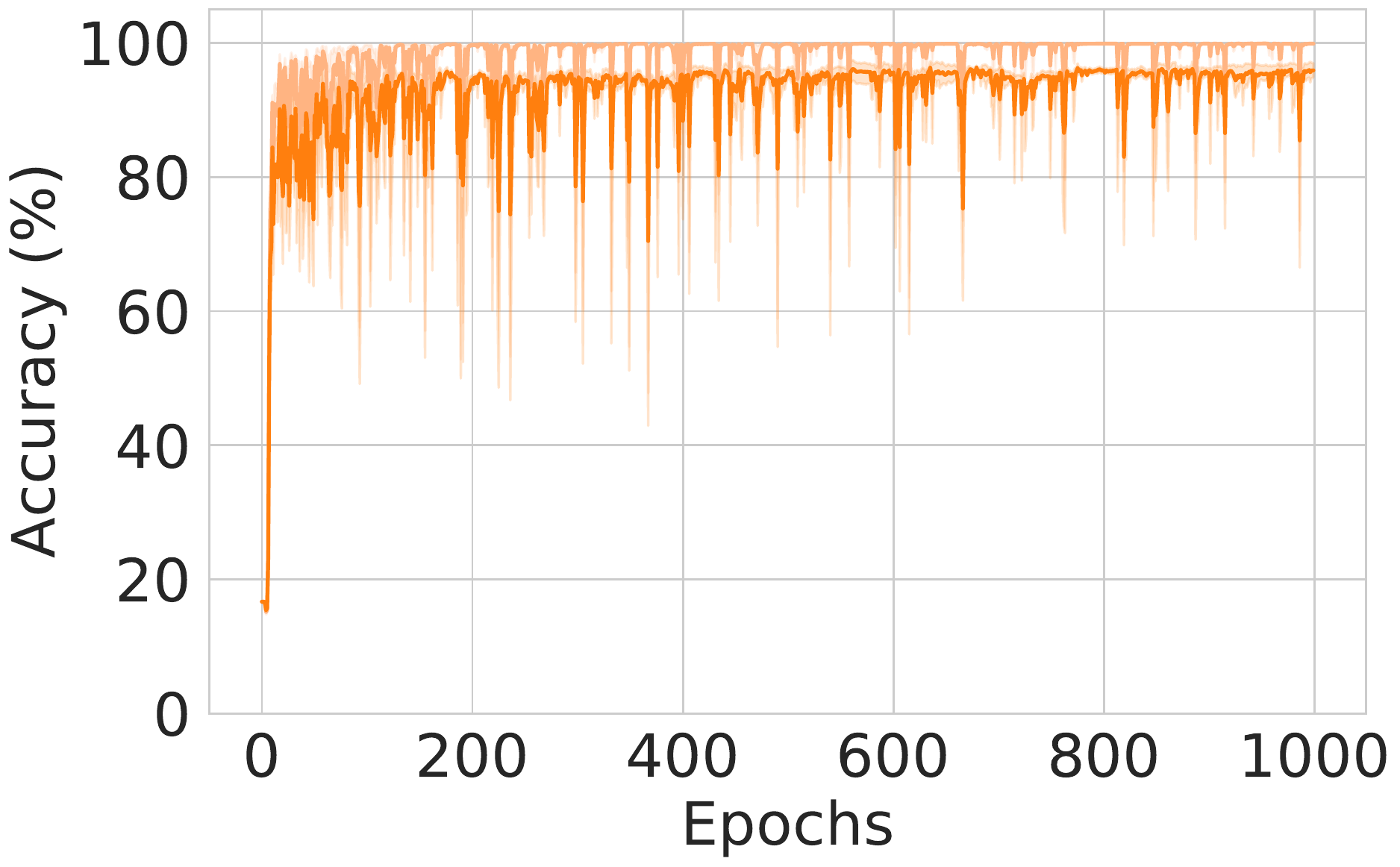}}\\
        \subfigure[CarsCG]{\label{fig:high_carcgs_late_stop}
            \includegraphics[width=0.46\columnwidth]{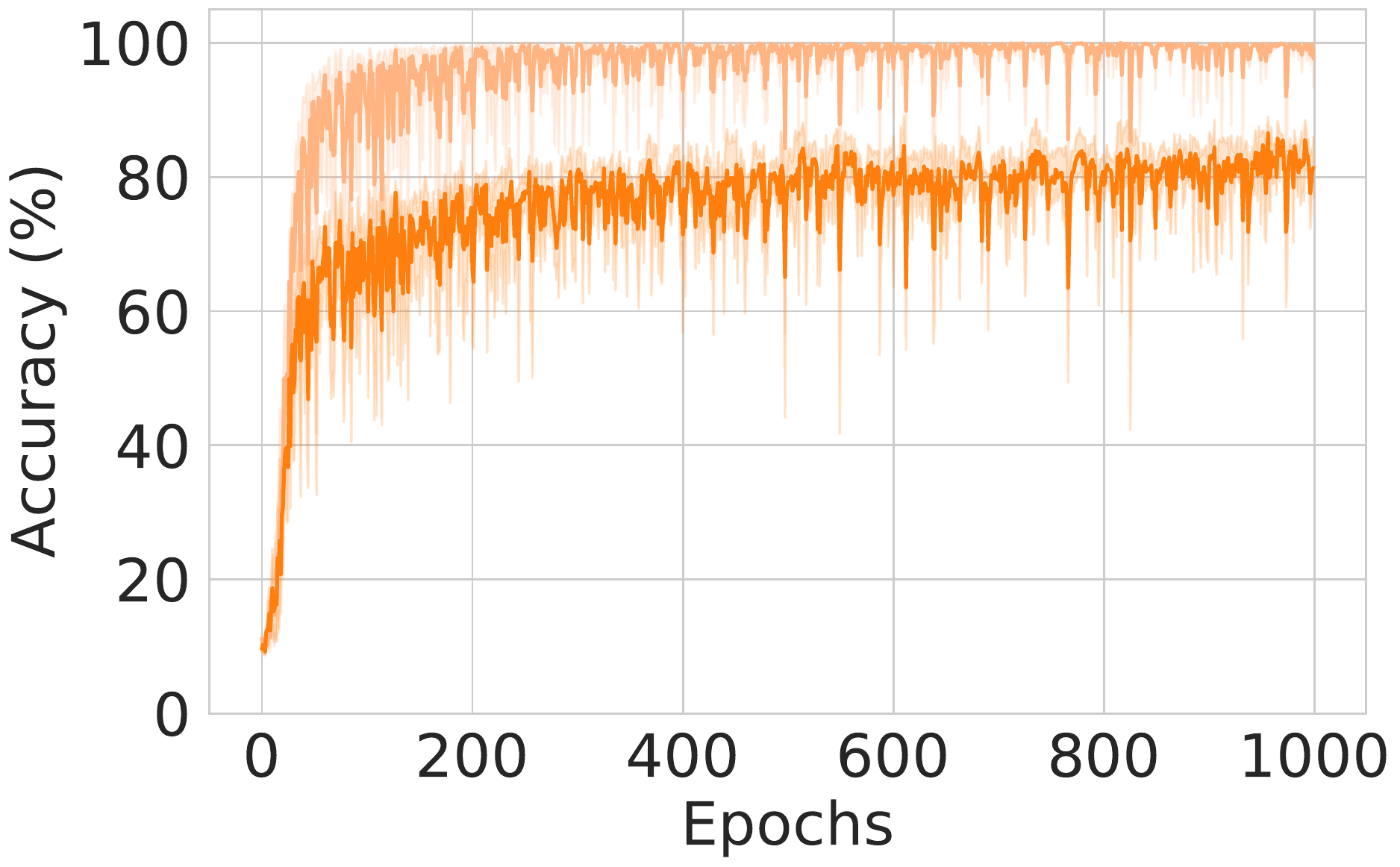}}
        \subfigure[MiscGoods]{\label{fig:high_daiso_late_stop}
            \includegraphics[width=0.46\columnwidth]{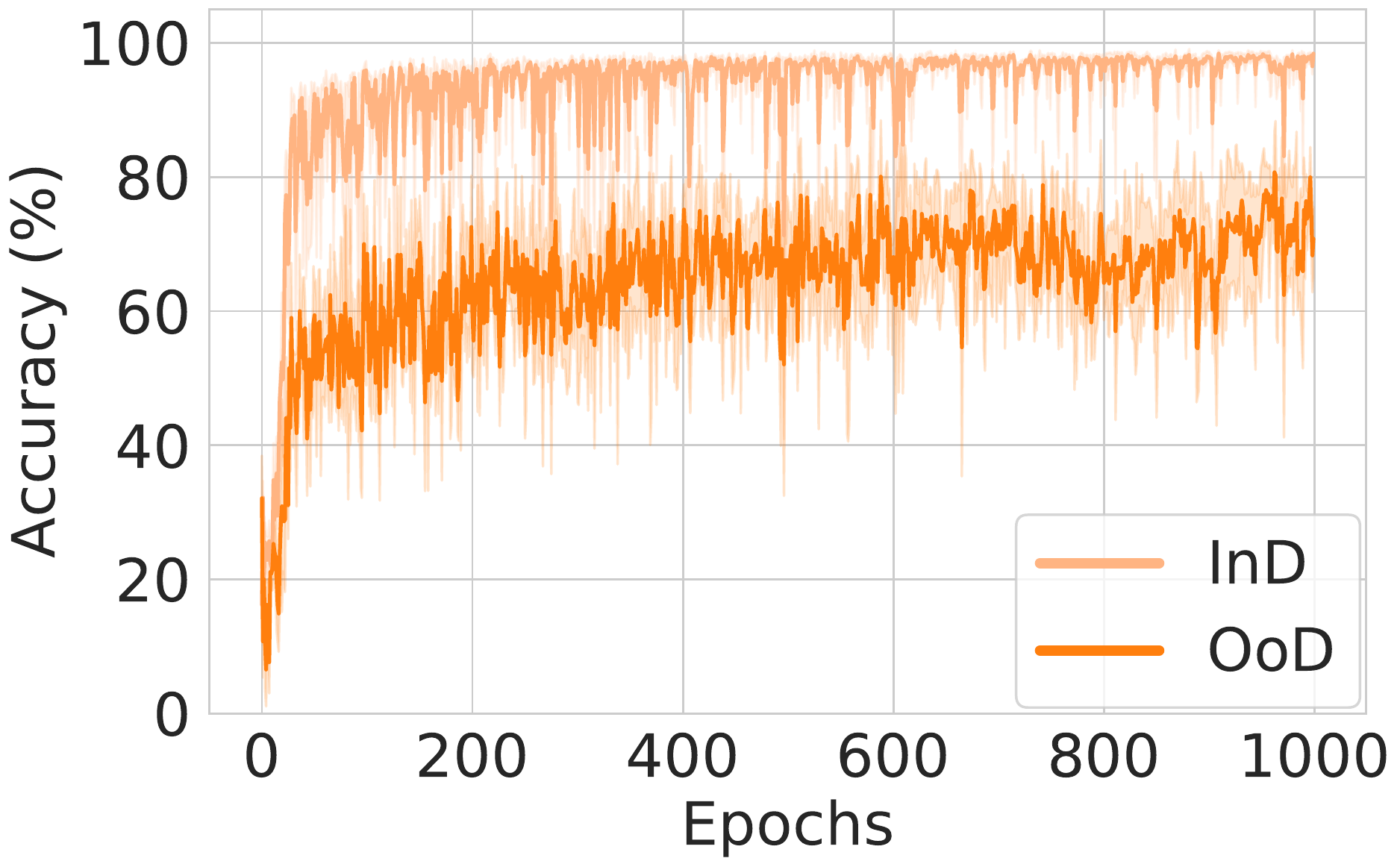}}
    \end{tabular}
\caption{Late-stopping with high InD data diversity}
\label{fig:high_late_stop_curves}
\end{figure*}

\begin{figure*}[ht]
\centering
    \begin{tabular}{cccccccc}
        \subfigure[Baseline MNIST]{\label{fig:low_mnist_baseline}
            \includegraphics[width=0.2\columnwidth]{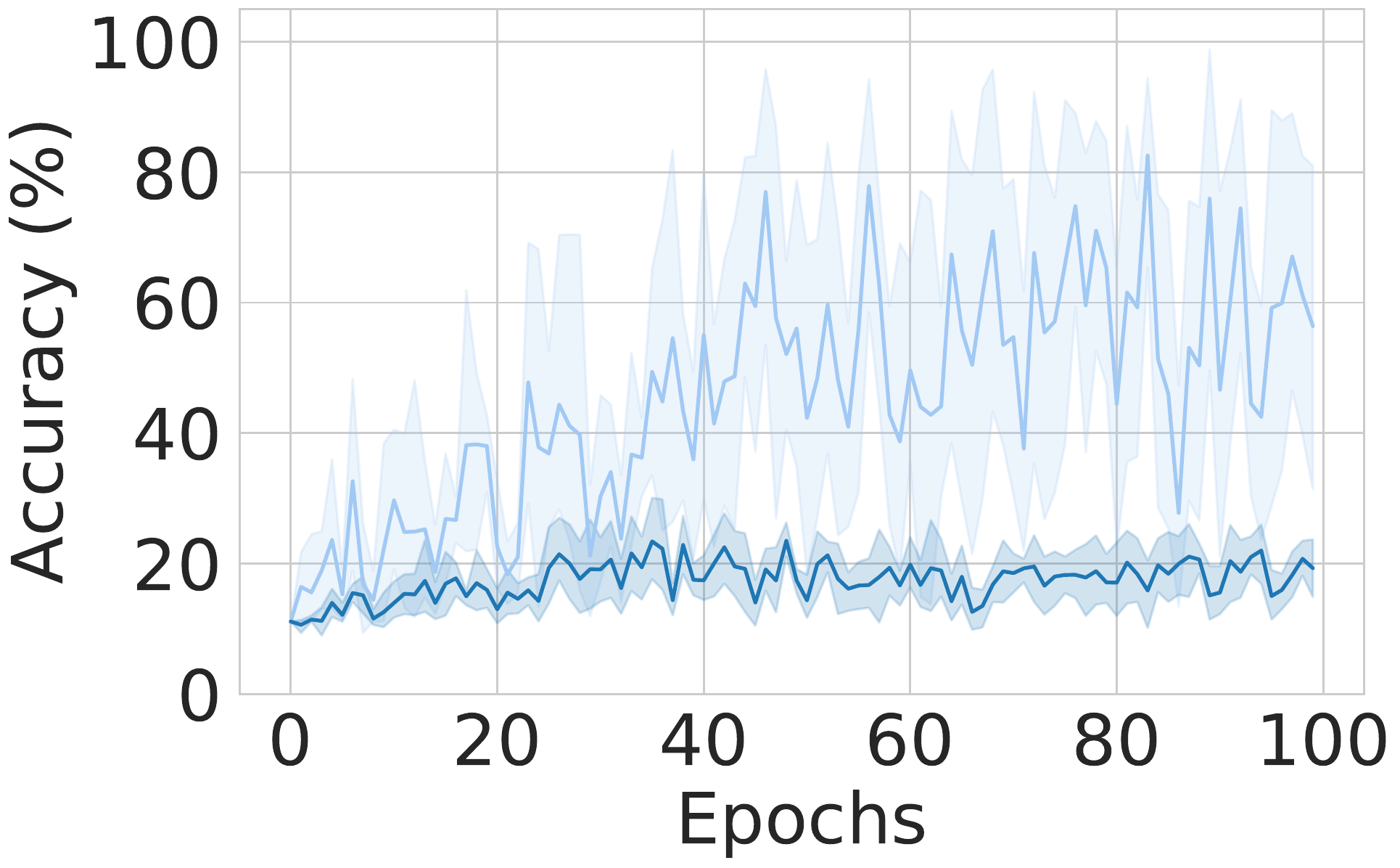}}
        \subfigure[Baseline iLab]{\label{fig:low_ilab_baseline}
            \includegraphics[width=0.2\columnwidth]{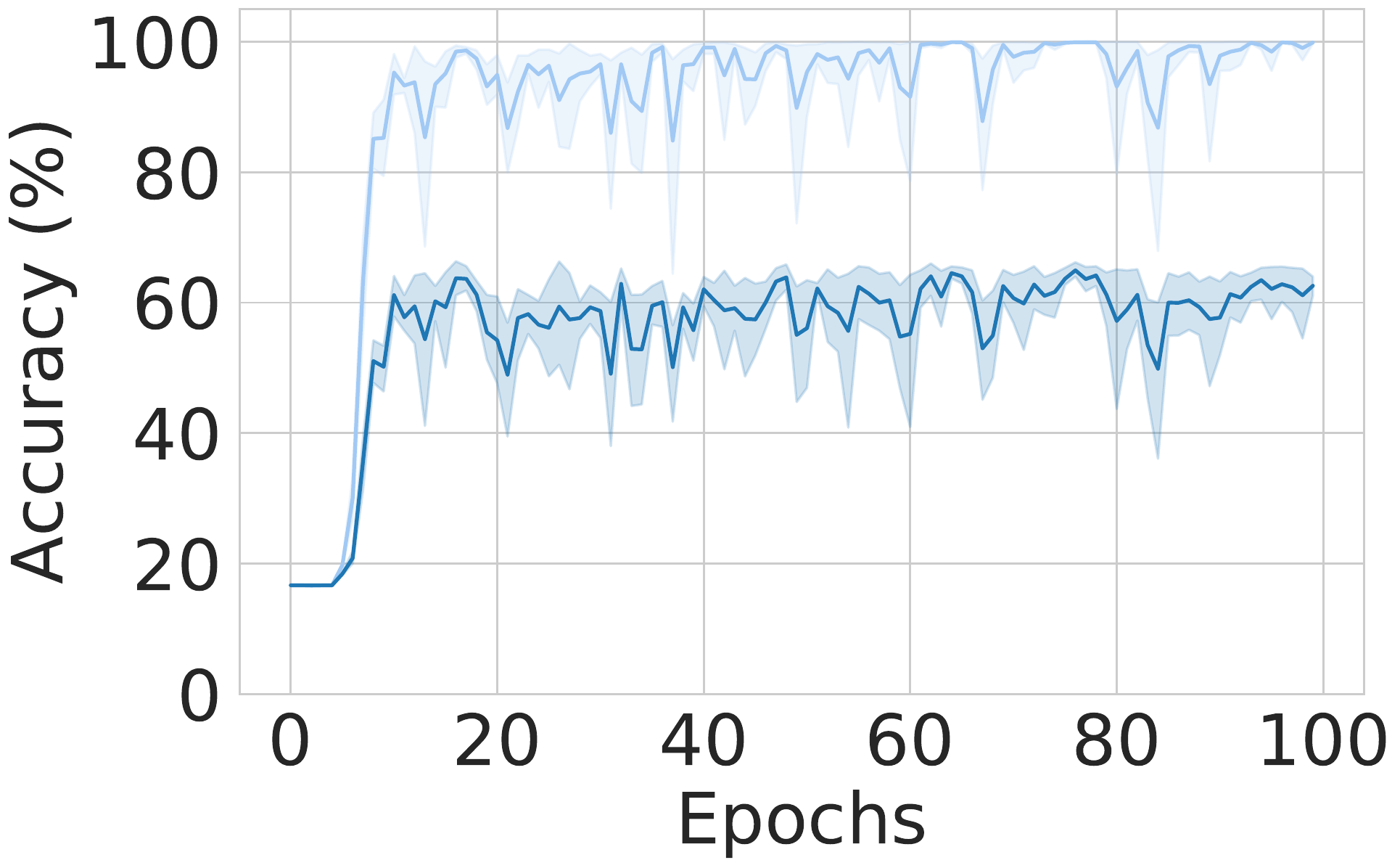}}
        \subfigure[Baseline CarsCG]{\label{fig:low_carcgs_baseline}
            \includegraphics[width=0.2\columnwidth]{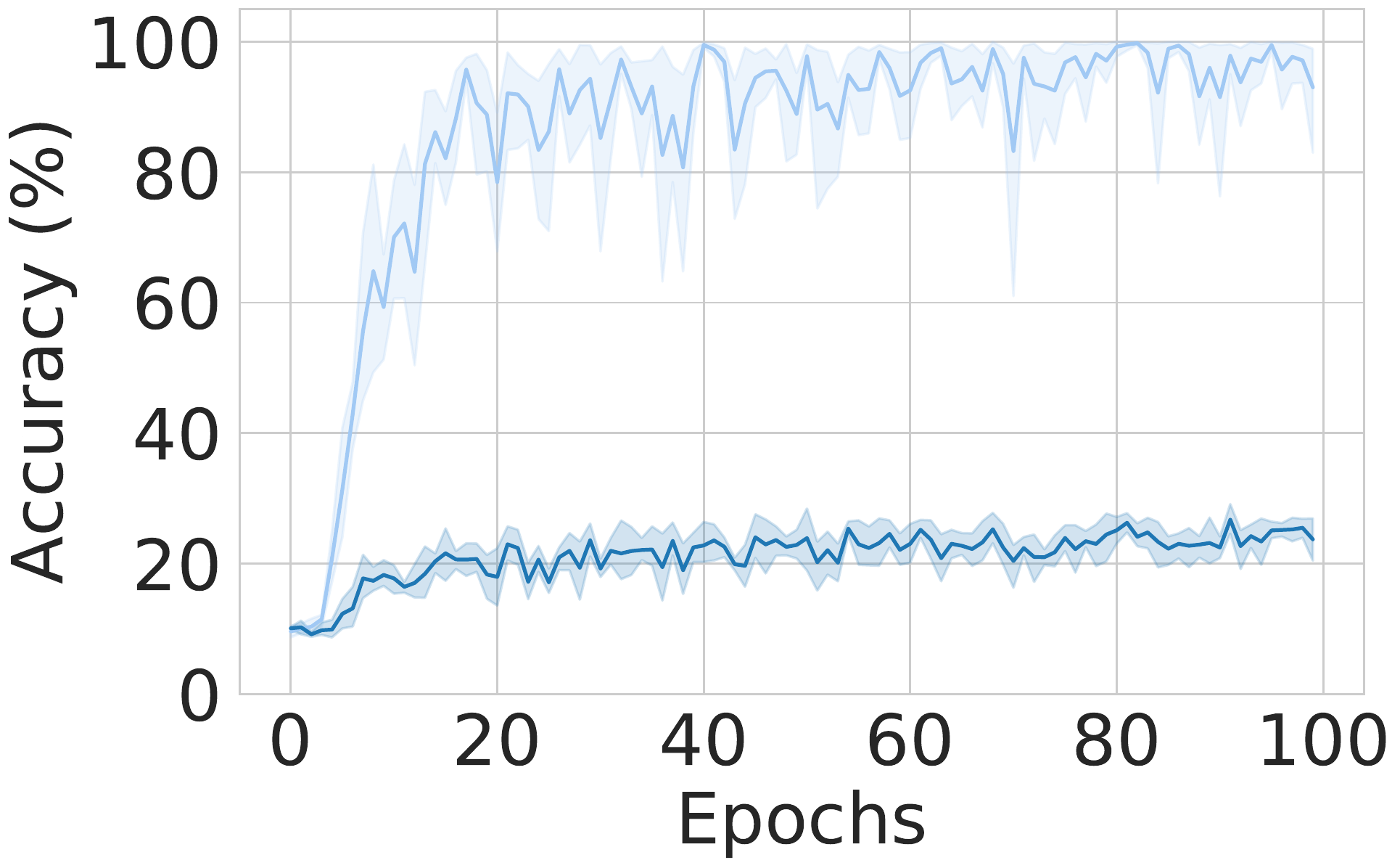}}
        \subfigure[Baseline MiscGoods]{\label{fig:low_daiso_baseline}
            \includegraphics[width=0.2\columnwidth]{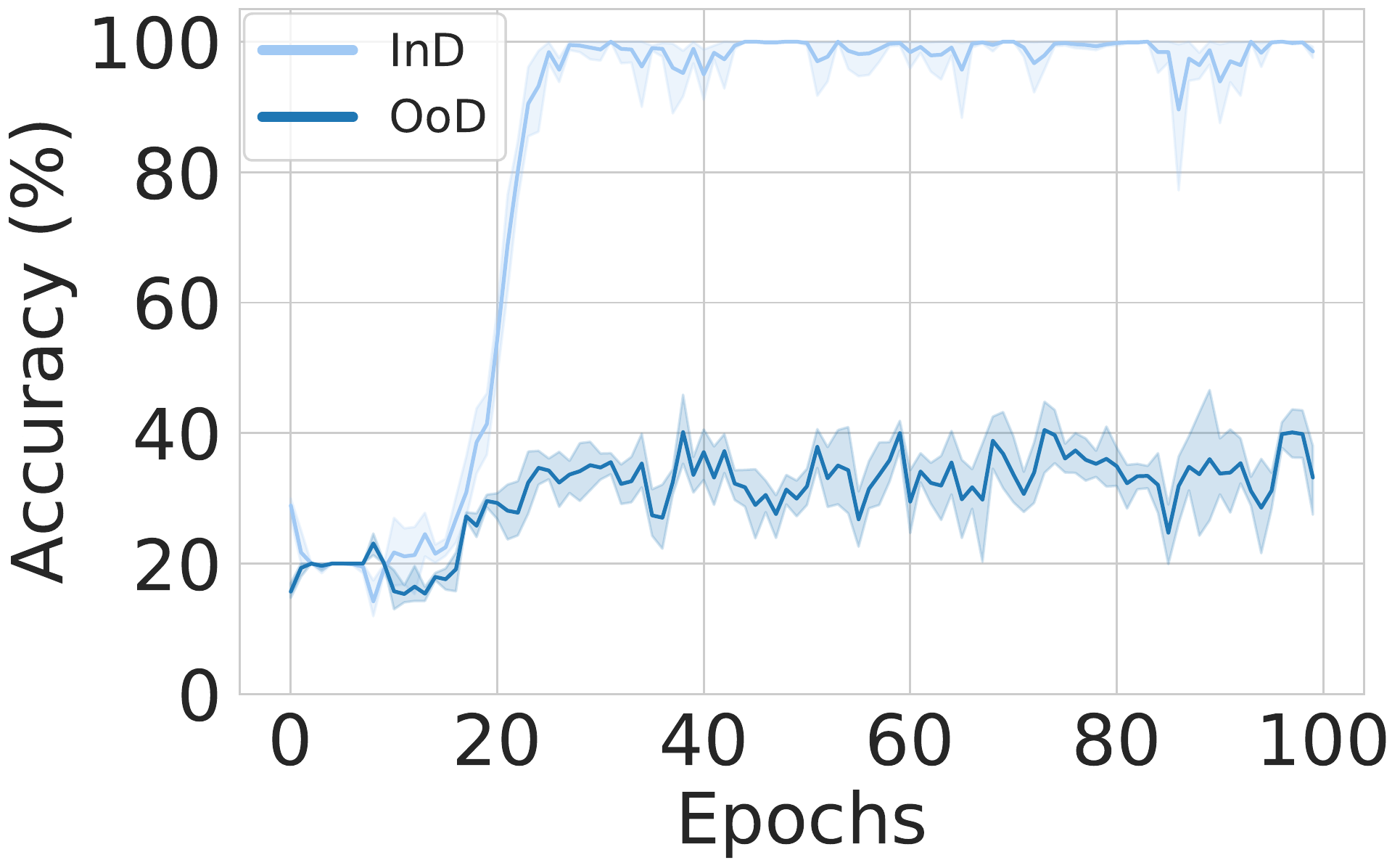}}\\
        \subfigure[BN momentum MNIST]{\label{fig:low_mnist_bnmomentum}
            \includegraphics[width=0.2\columnwidth]{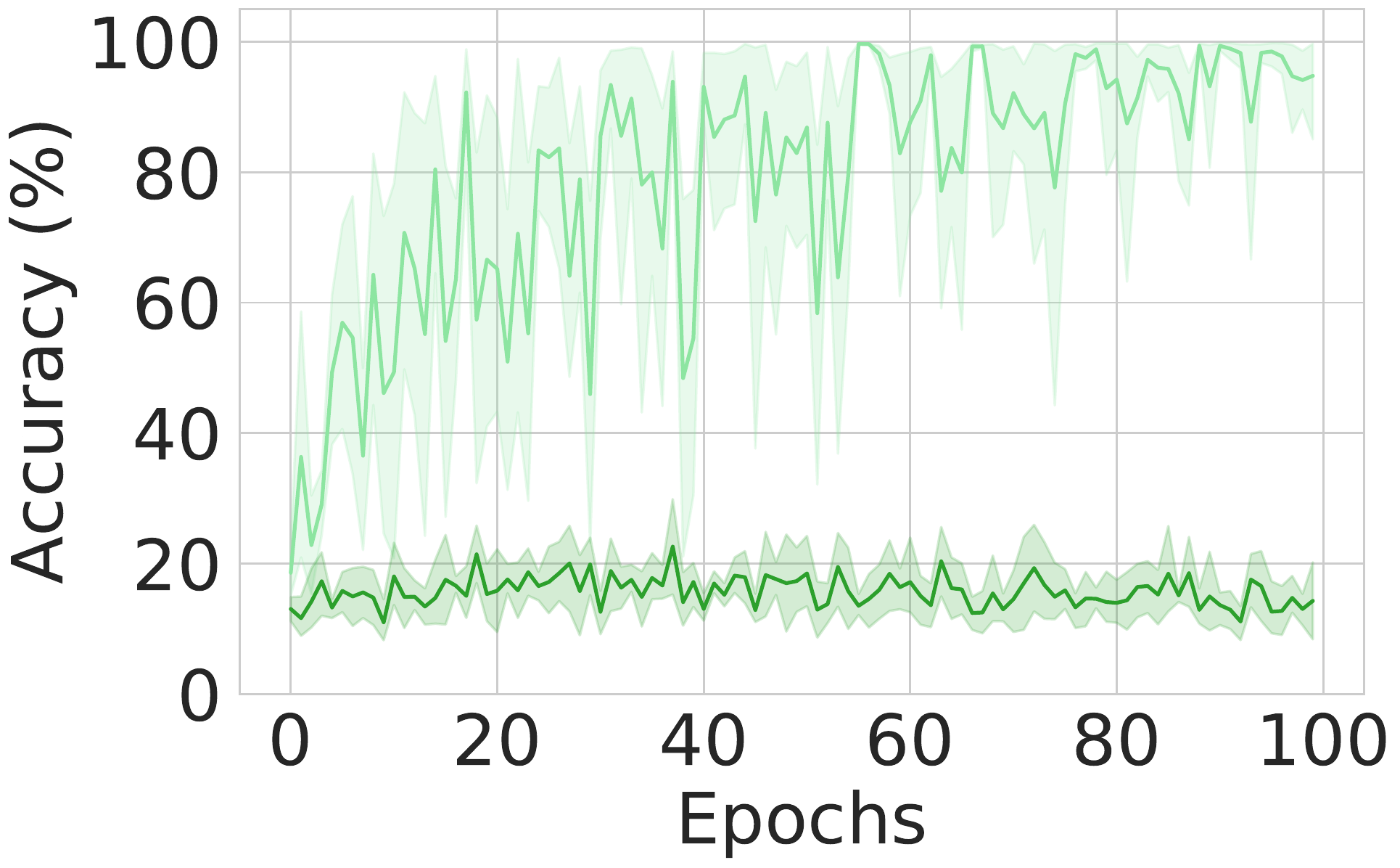}}
        \subfigure[BN momentum iLab]{\label{fig:low_ilab_bnmomentum}
            \includegraphics[width=0.2\columnwidth]{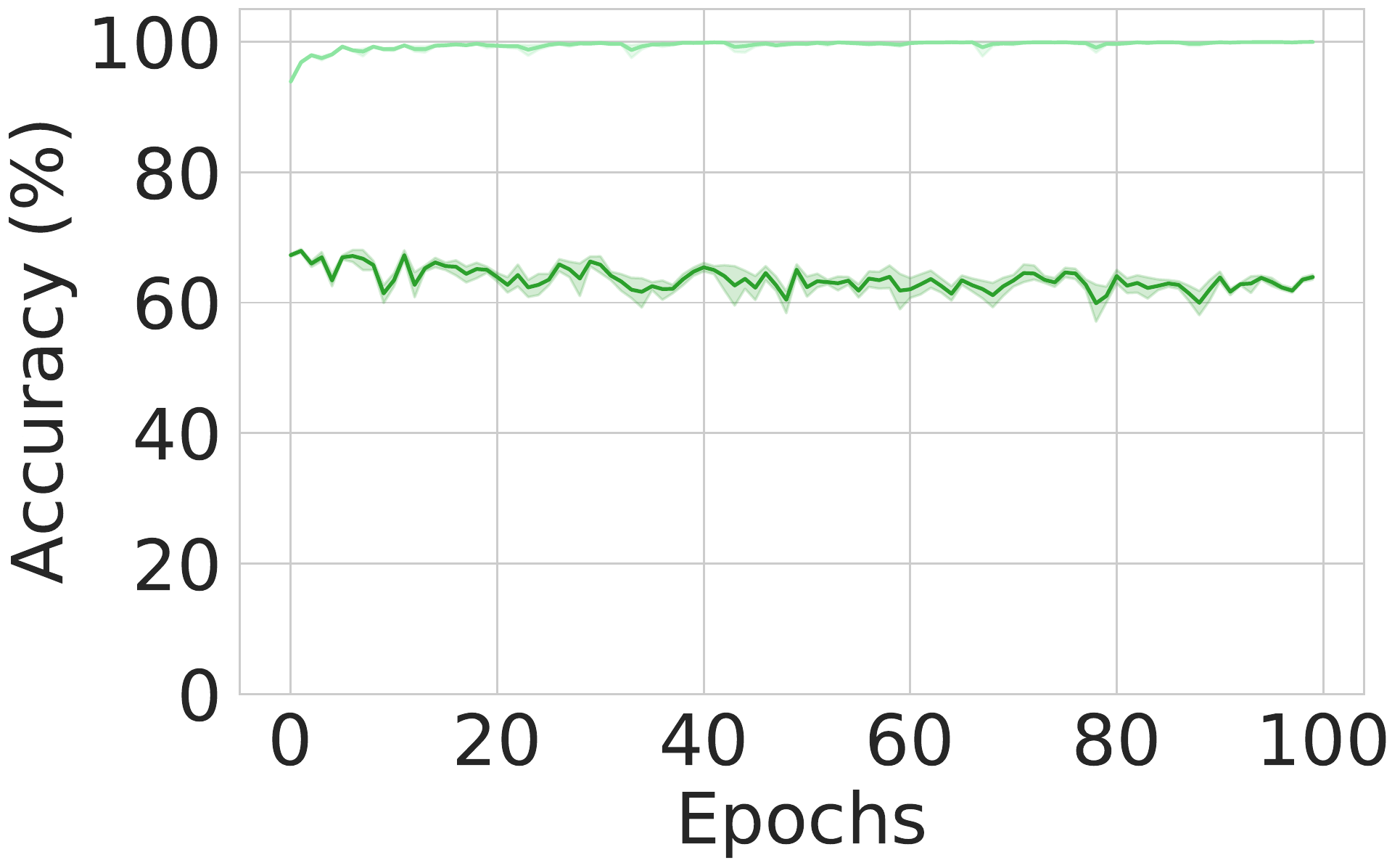}}
        \subfigure[BN momentum CarsCG]{\label{fig:low_carcgs_bnmomentum}
            \includegraphics[width=0.2\columnwidth]{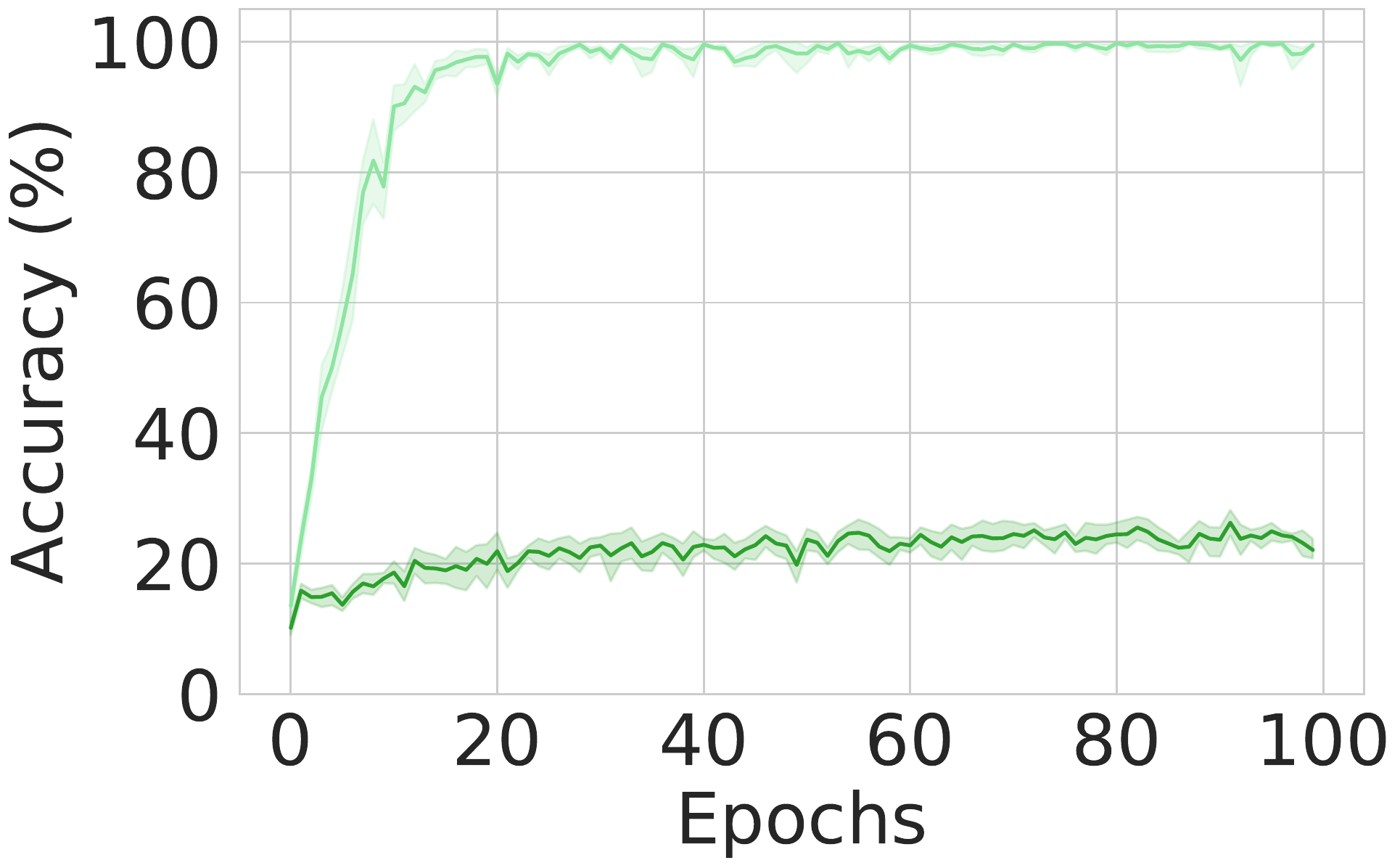}}
        \subfigure[BN momentum MiscGoods]{\label{fig:low_daiso_bnmomentum}
            \includegraphics[width=0.2\columnwidth]{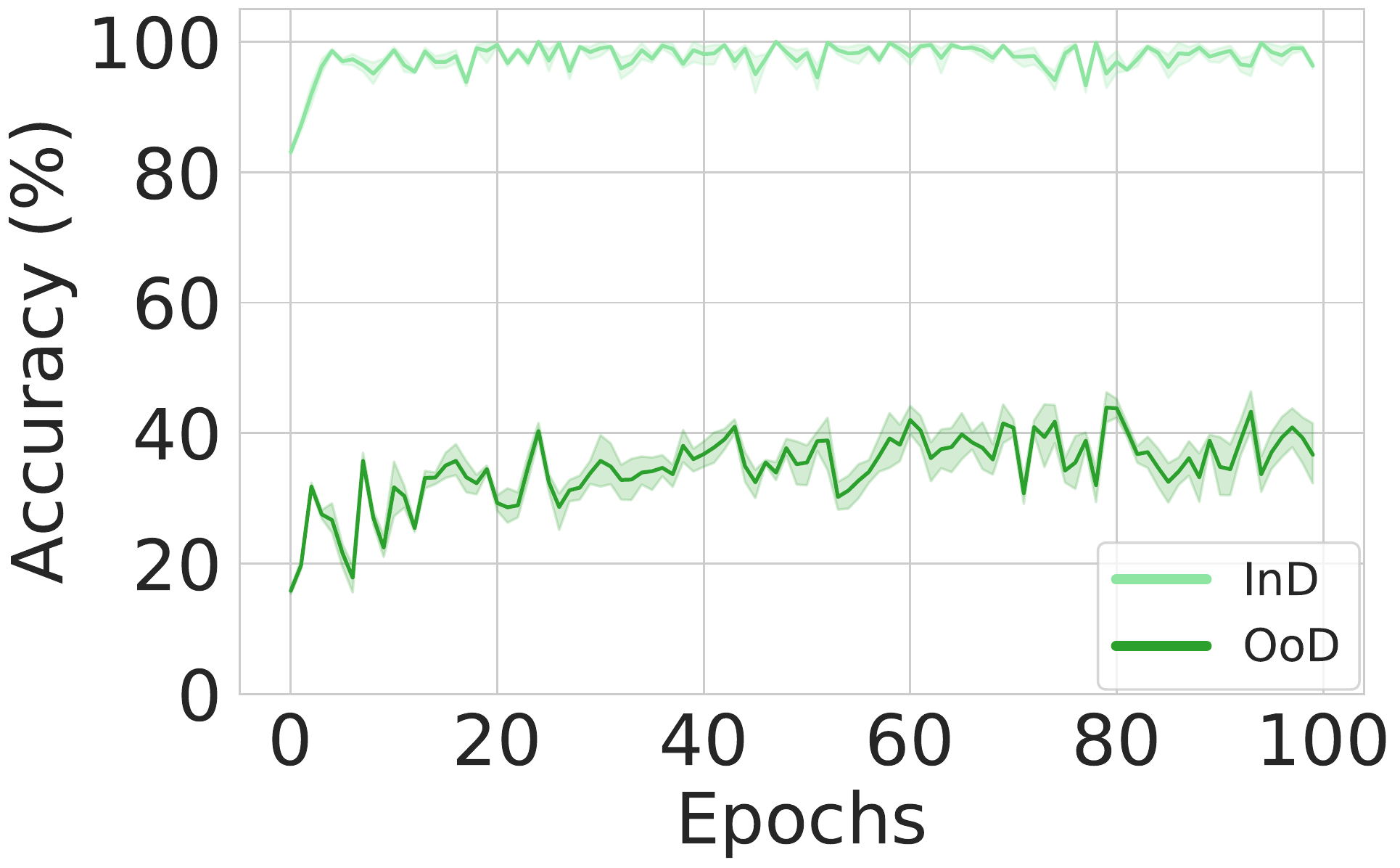}}
    \end{tabular}
\caption{Baseline and BN momentum with low InD data diversity}
\label{fig:low_baseline_bnm_curves}
\end{figure*}

\begin{figure*}[ht]
\centering
    \begin{tabular}{cccccccc}
        \subfigure[Baseline MNIST]{\label{fig:medium_mnist_baseline}
            \includegraphics[width=0.2\columnwidth]{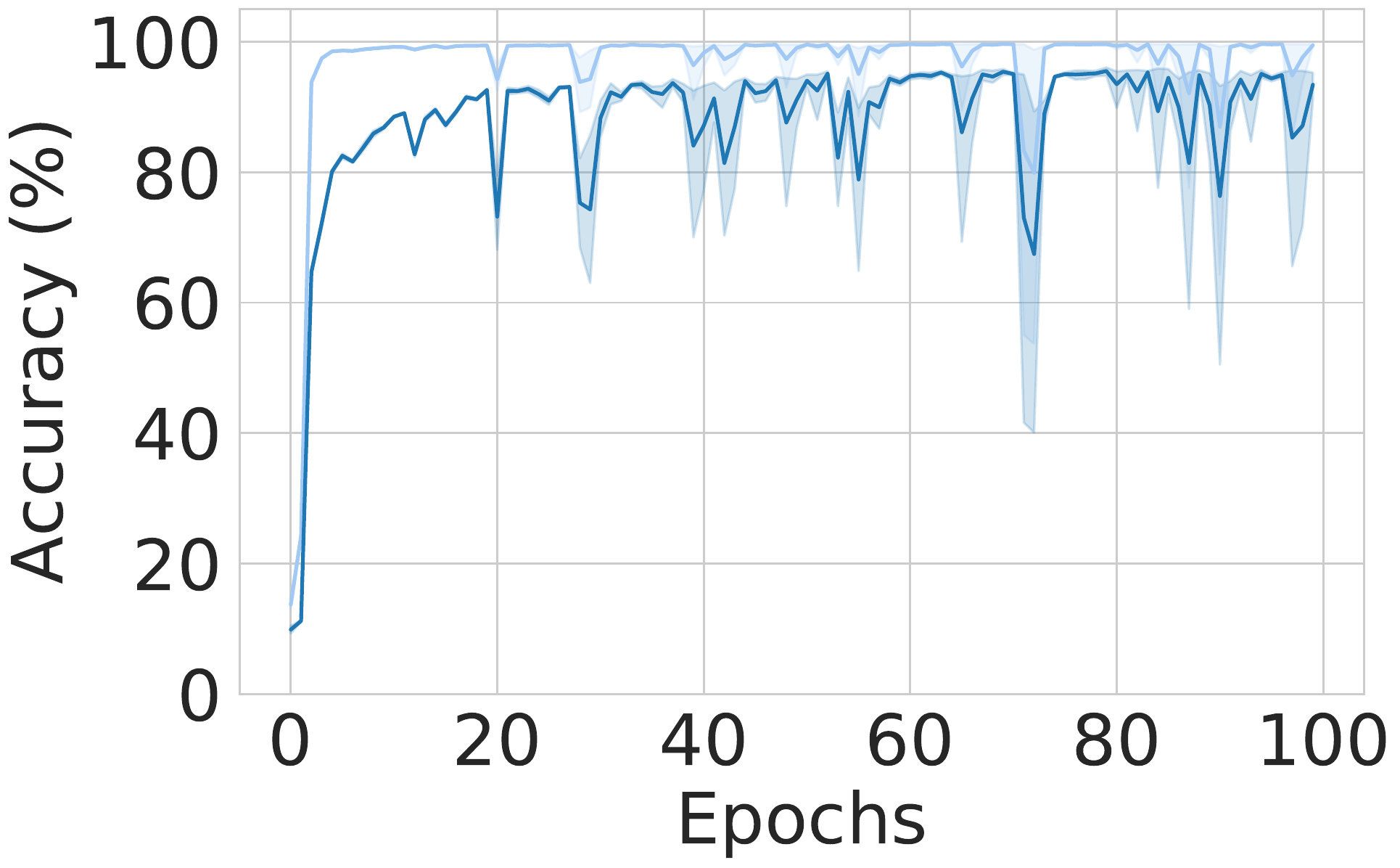}}
        \subfigure[Baseline iLab]{\label{fig:medium_ilab_baseline}
            \includegraphics[width=0.2\columnwidth]{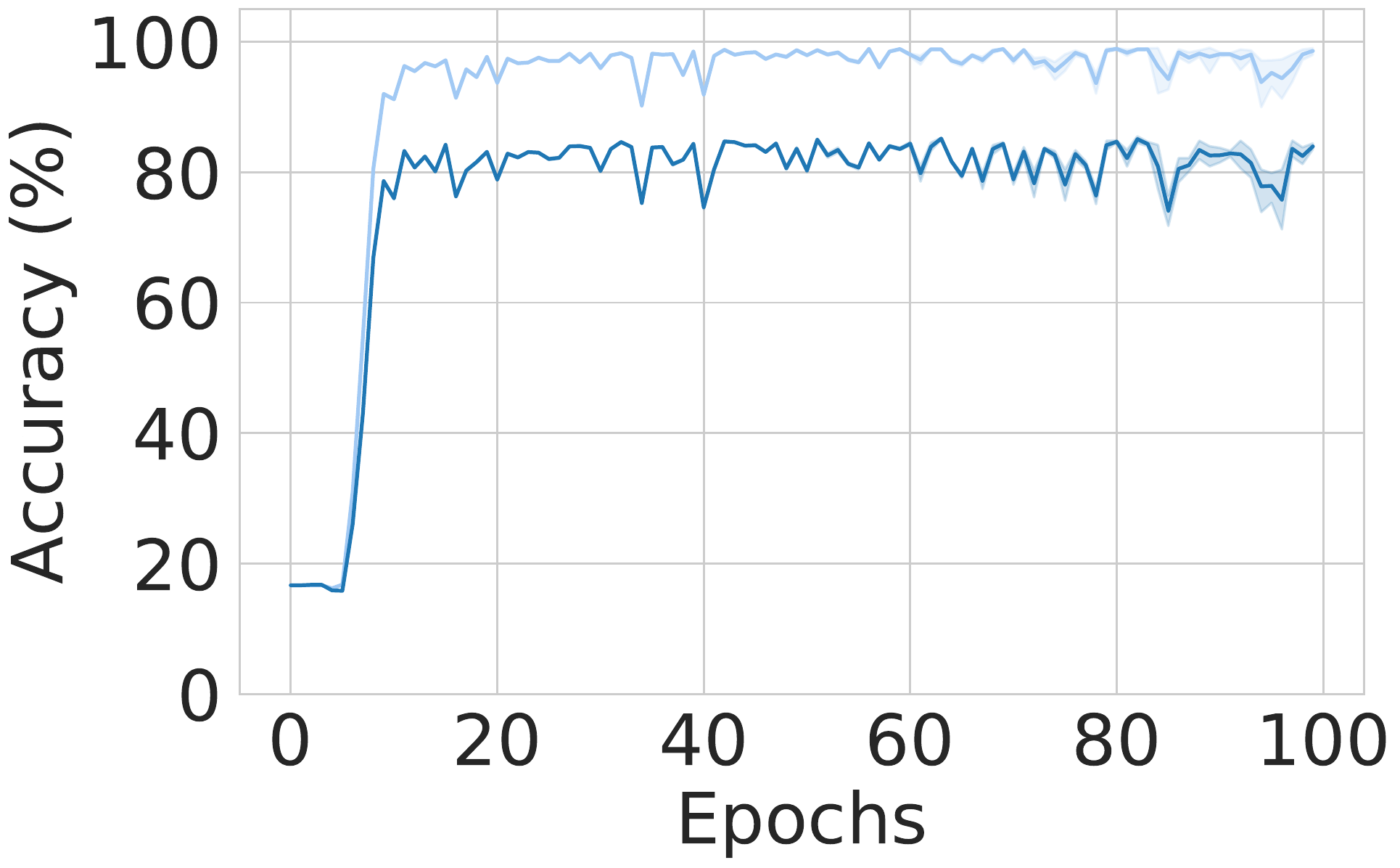}}
        \subfigure[Baseline CarsCG]{\label{fig:medium_carcgs_baseline}
            \includegraphics[width=0.2\columnwidth]{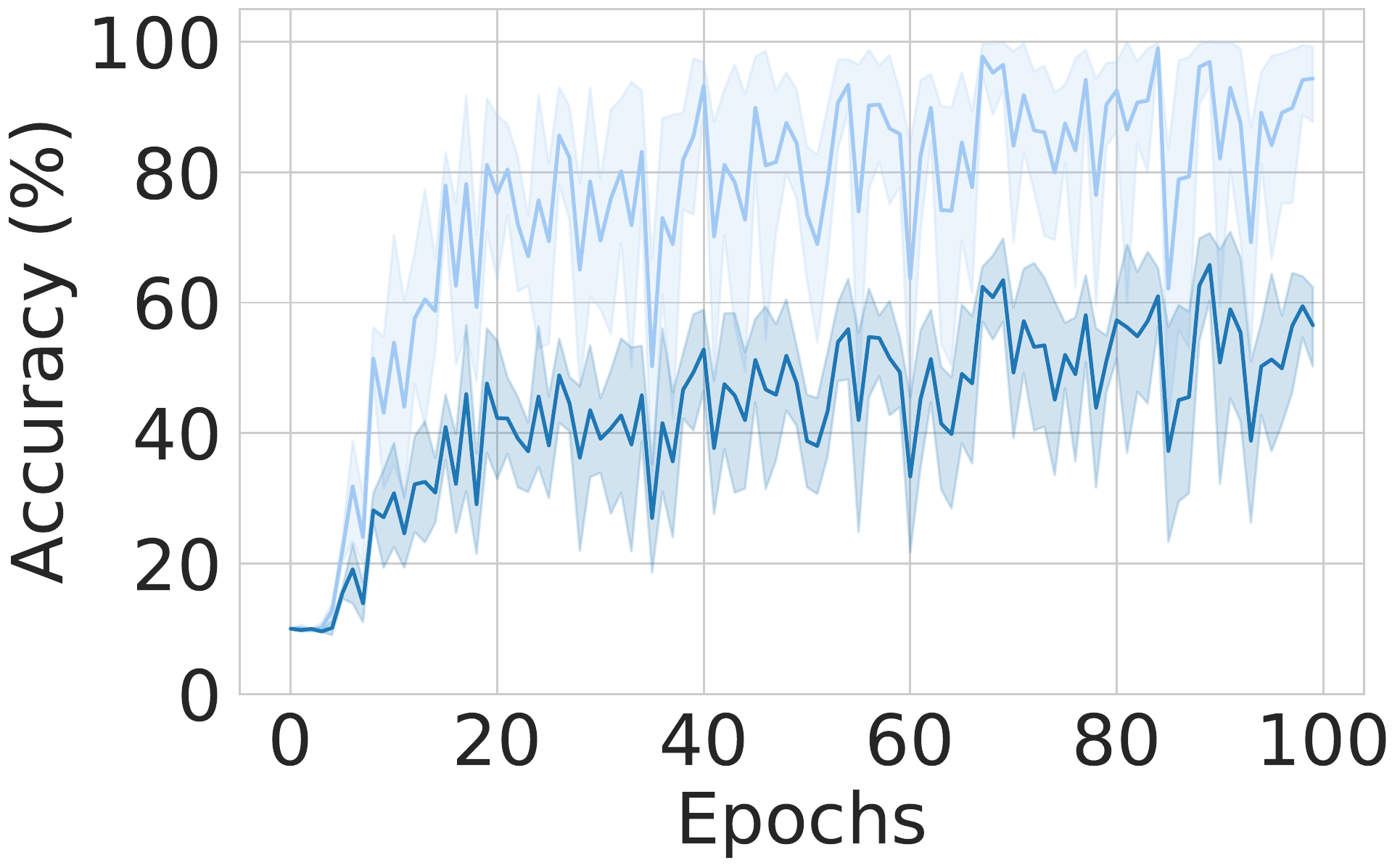}}
        \subfigure[Baseline MiscGoods]{\label{fig:medium_daiso_baseline}
            \includegraphics[width=0.2\columnwidth]{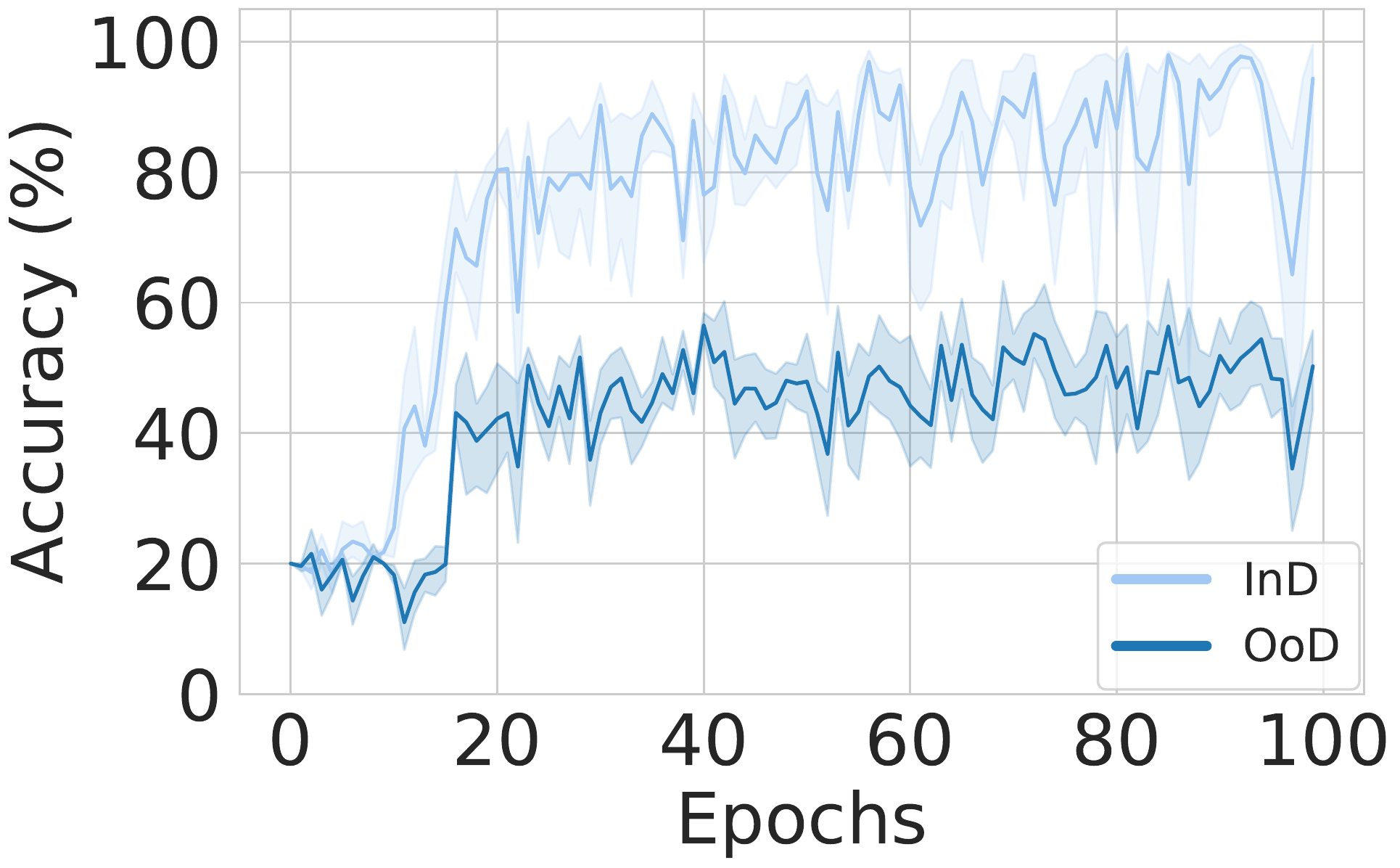}}\\
        \subfigure[BN momentum MNIST]{\label{fig:medium_mnist_bnmomentum}
            \includegraphics[width=0.2\columnwidth]{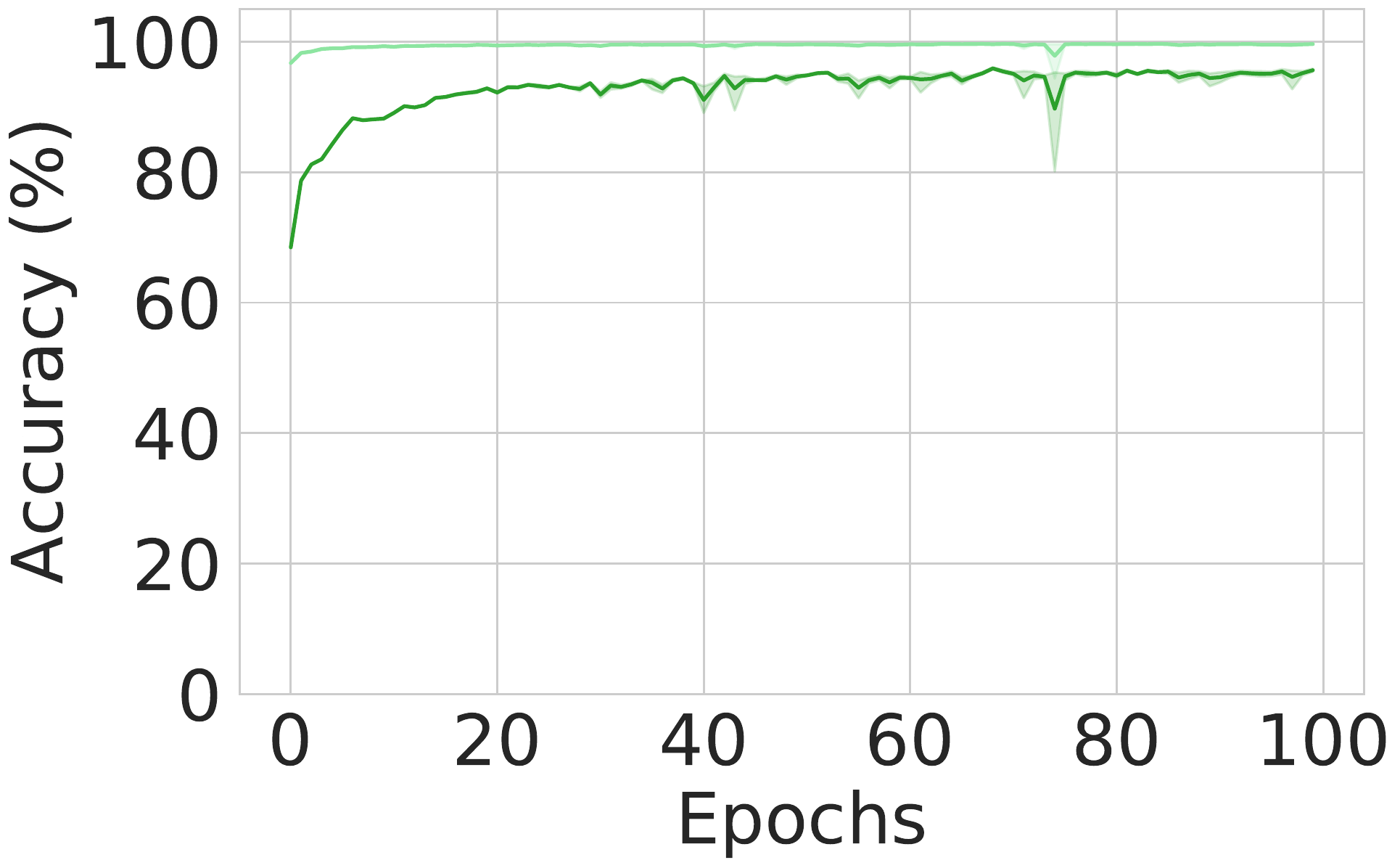}}
        \subfigure[BN momentum iLab]{\label{fig:medium_ilab_bnmomentum}
            \includegraphics[width=0.2\columnwidth]{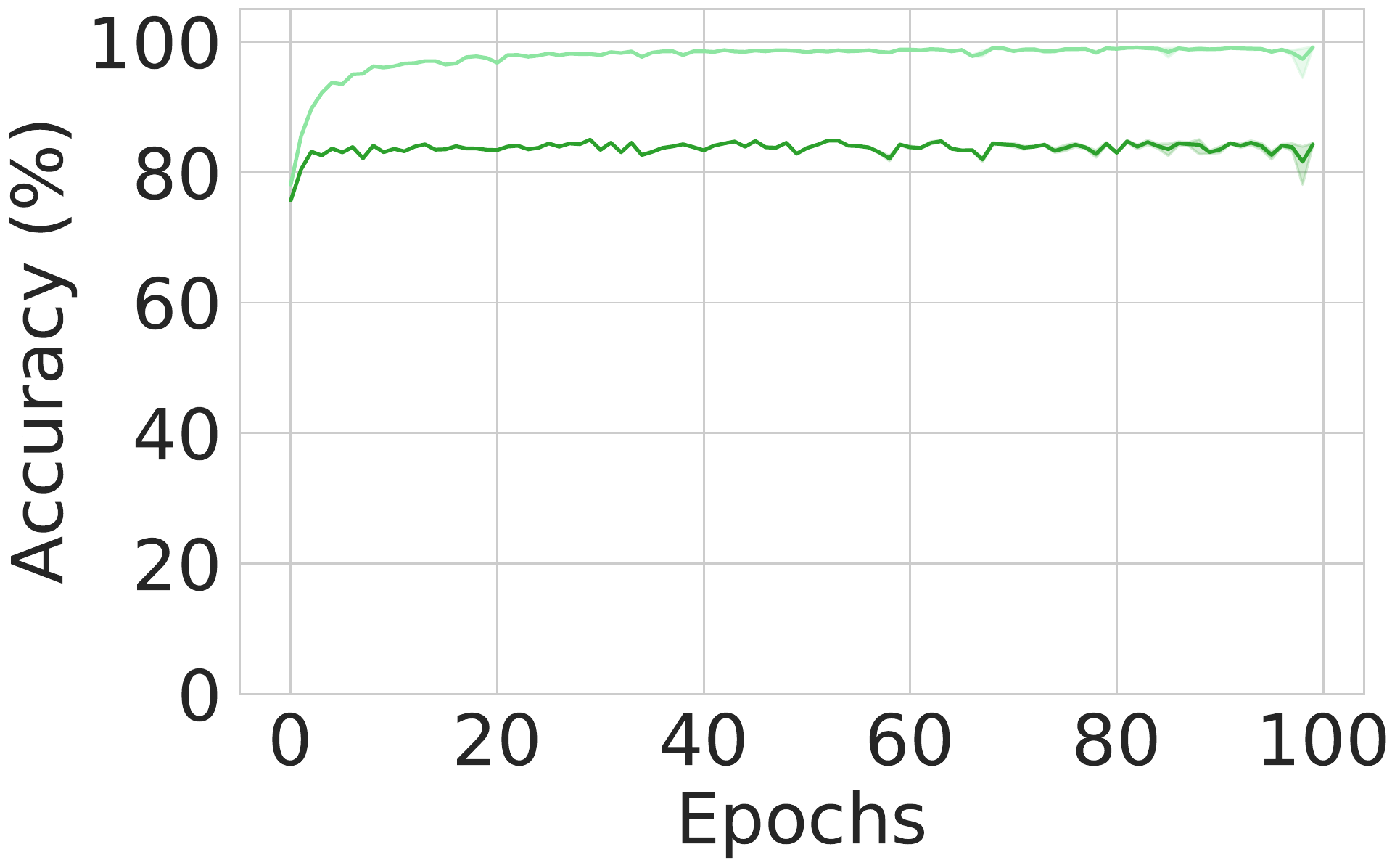}}
        \subfigure[BN momentum CarsCG]{\label{fig:medium_carcgs_bnmomentum}
            \includegraphics[width=0.2\columnwidth]{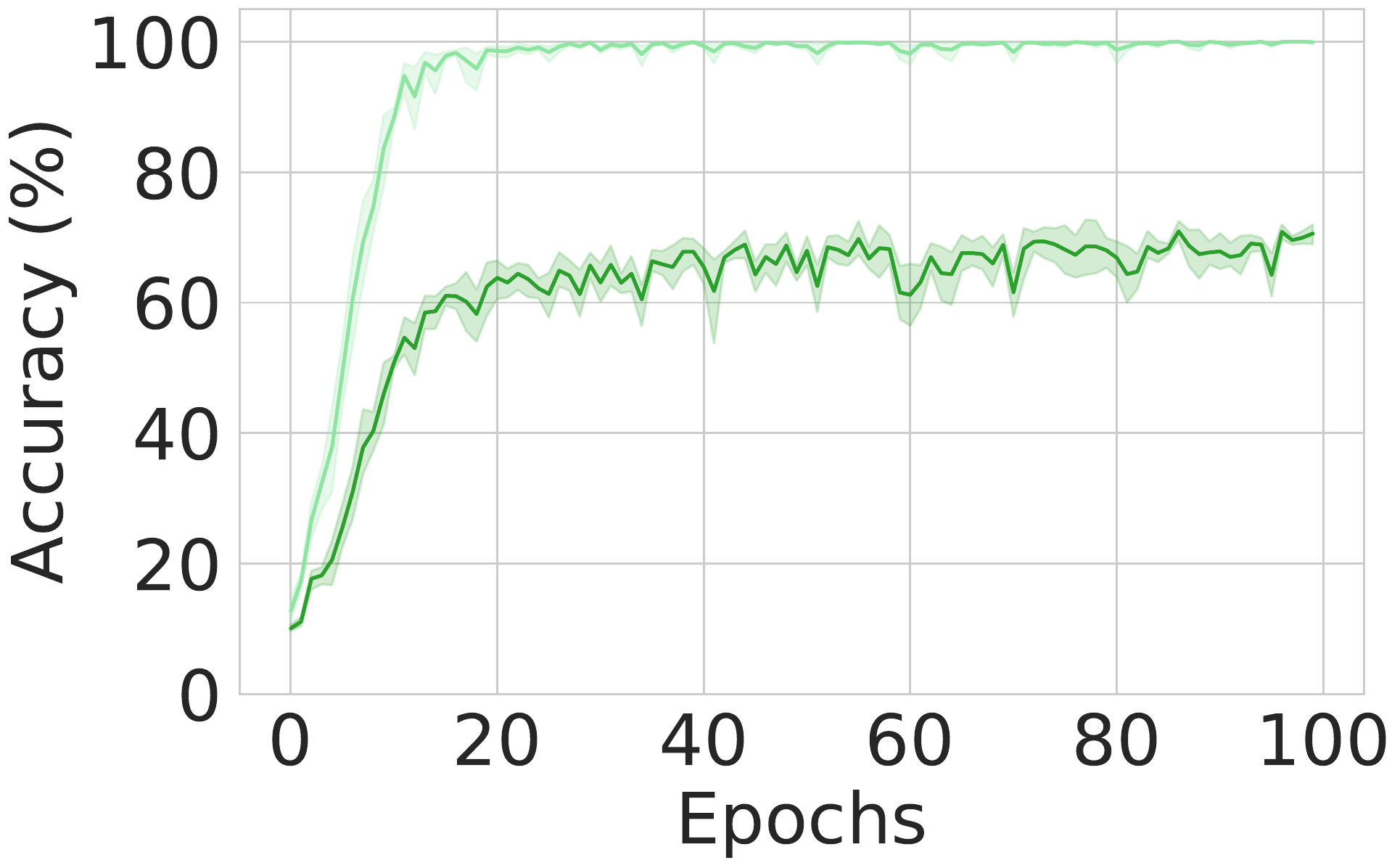}}
        \subfigure[BN momentum MiscGoods]{\label{fig:medium_daiso_bnmomentum}
            \includegraphics[width=0.2\columnwidth]{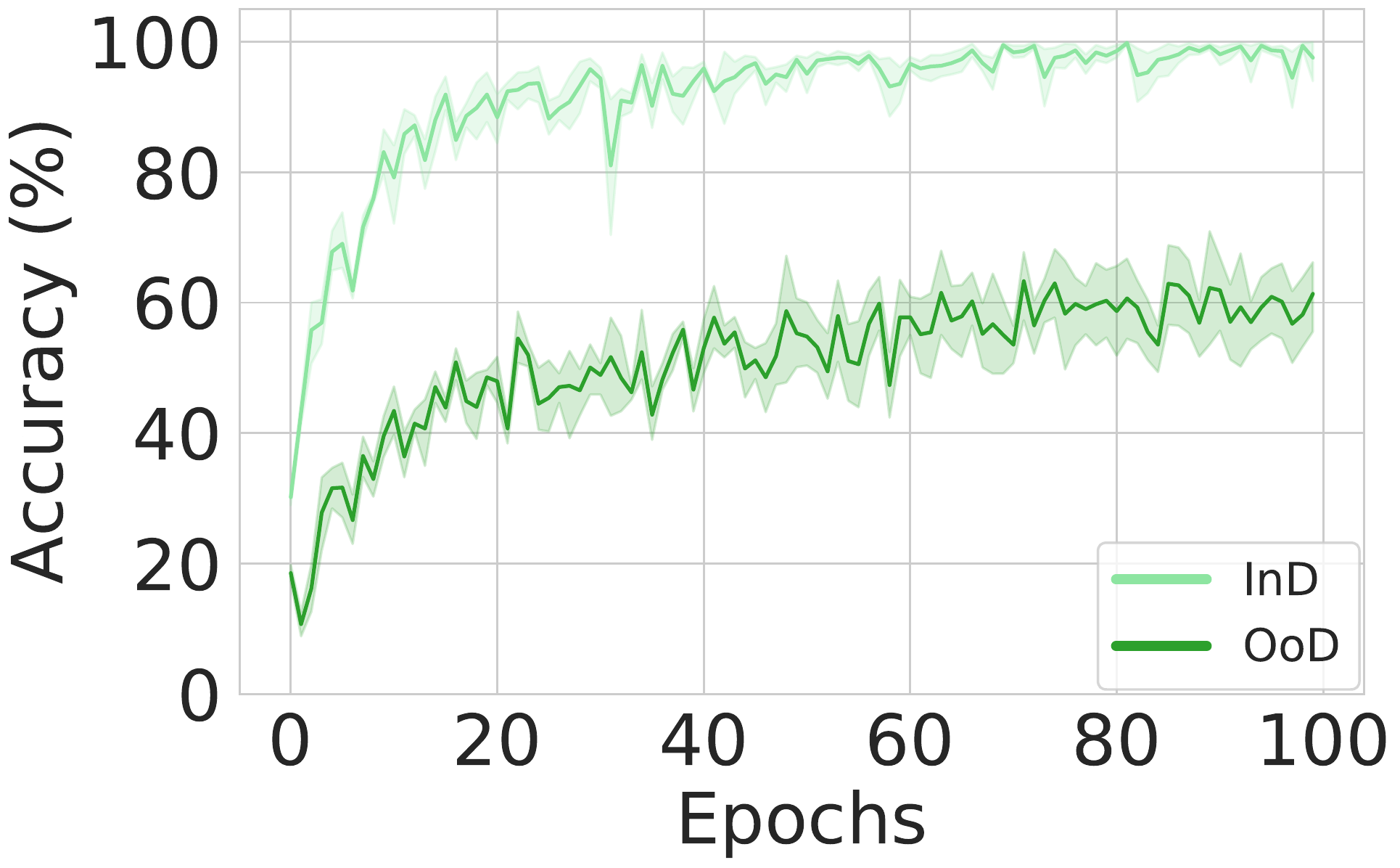}}
    \end{tabular}
\caption{Baseline and BN momentum with medium InD data diversity}
\label{fig:medium_baseline_bnm_curves}
\end{figure*}

\begin{figure*}[ht]
\centering
    \begin{tabular}{cccccccc}
        \subfigure[Baseline MNIST]{\label{fig:high_mnist_baseline}
            \includegraphics[width=0.2\columnwidth]{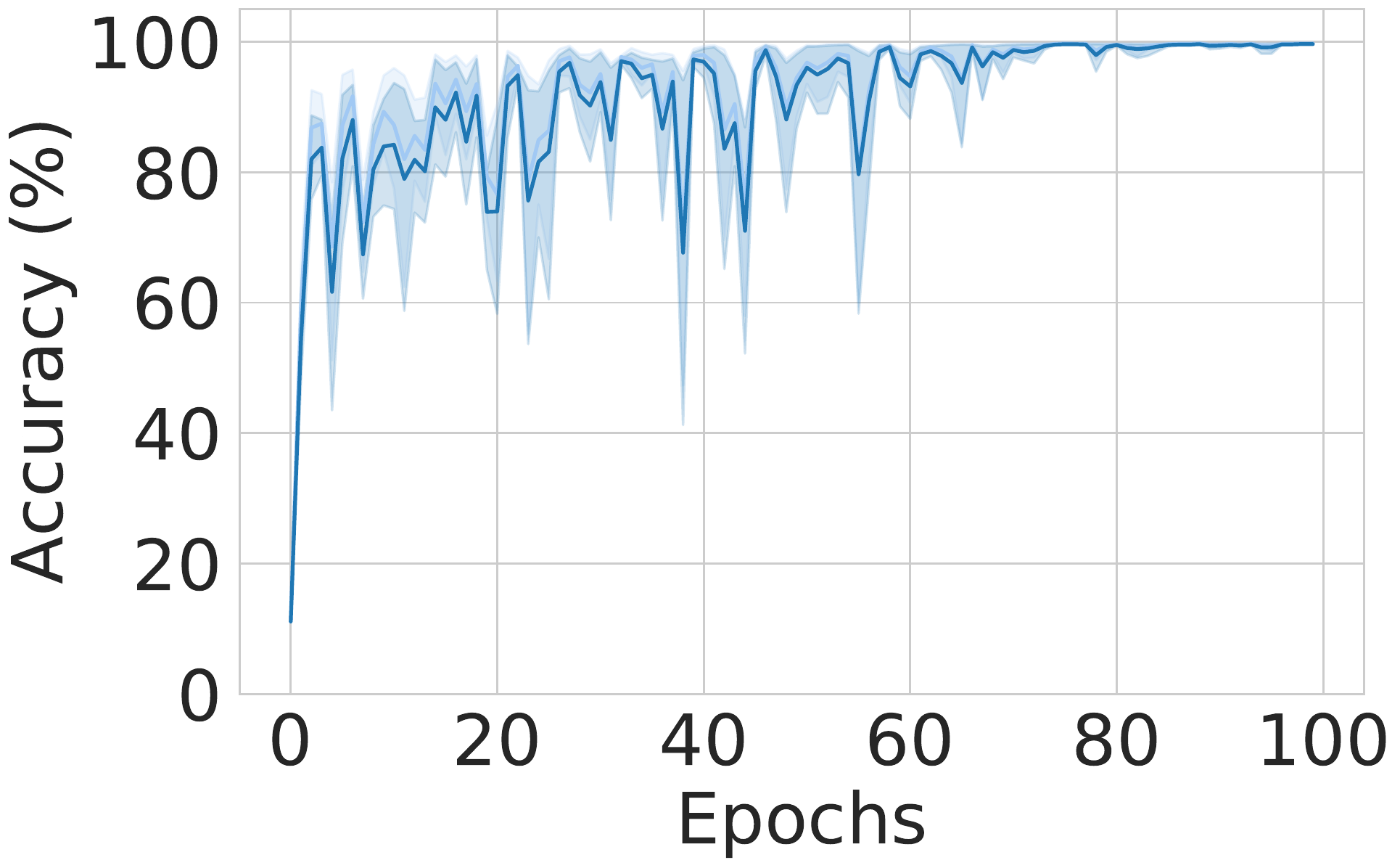}}
        \subfigure[Baseline iLab]{\label{fig:high_ilab_baseline}
            \includegraphics[width=0.2\columnwidth]{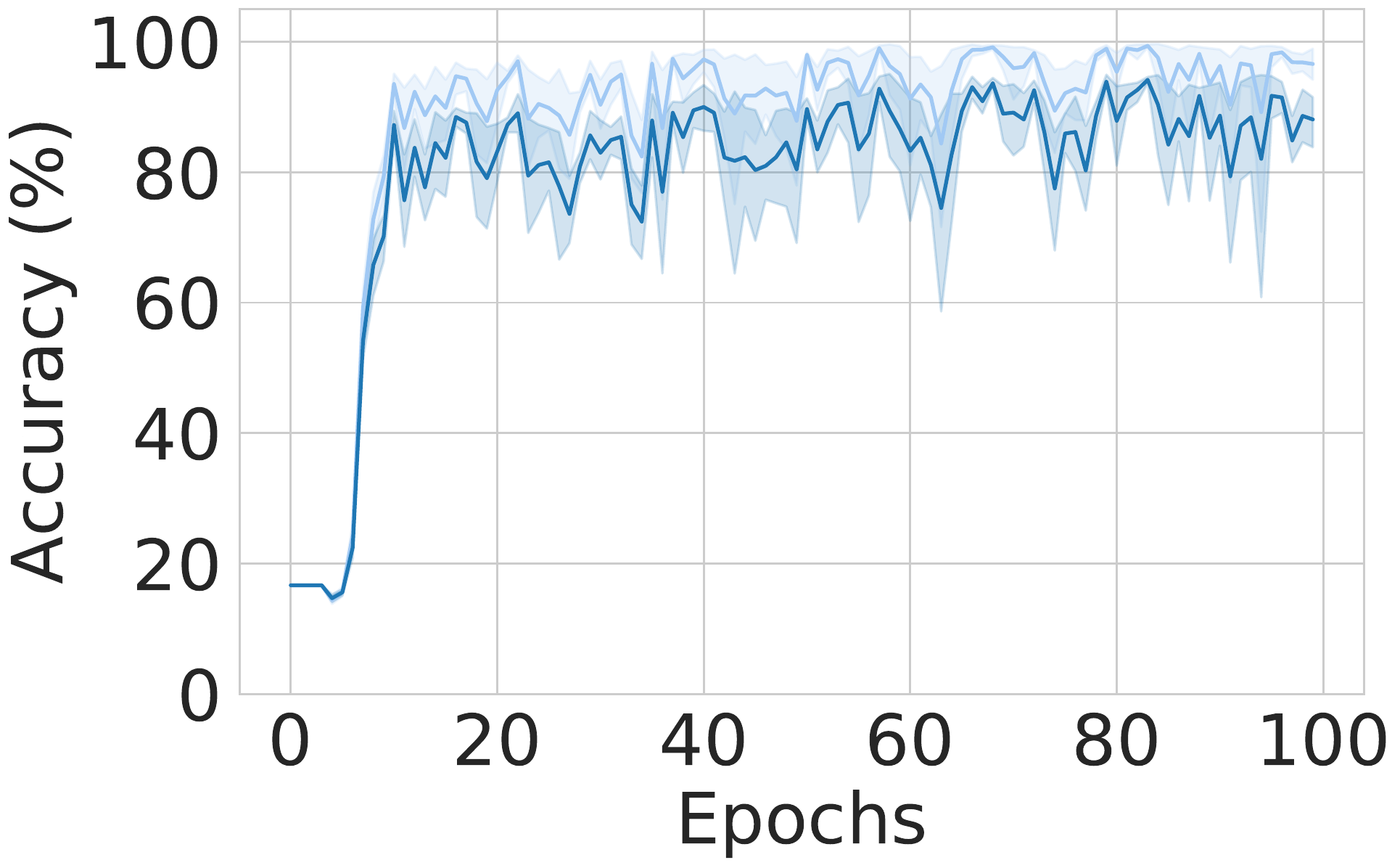}}
        \subfigure[Baseline CarsCG]{\label{fig:high_carcgs_baseline}
            \includegraphics[width=0.2\columnwidth]{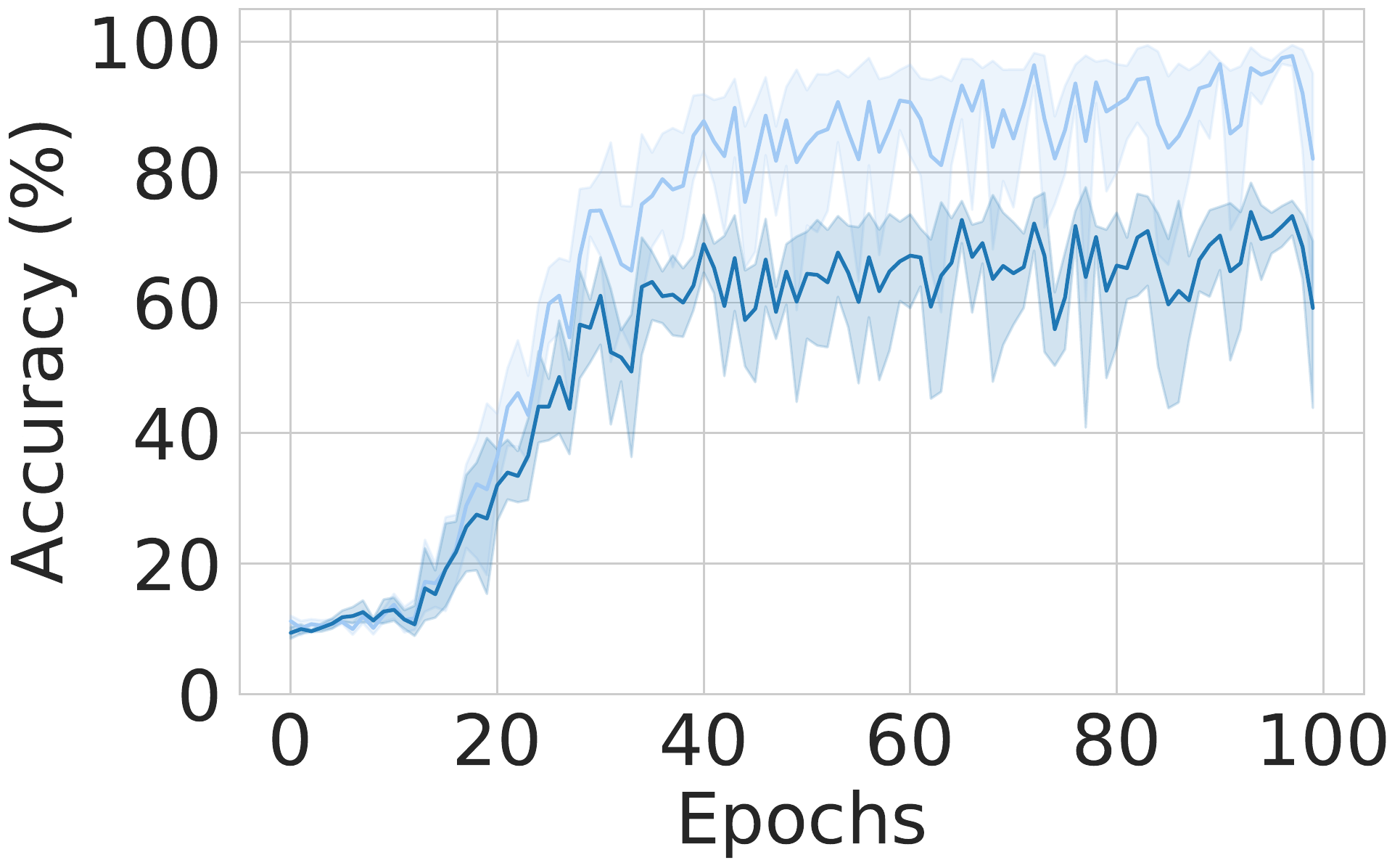}}
        \subfigure[Baseline MiscGoods]{\label{fig:high_daiso_baseline}
            \includegraphics[width=0.2\columnwidth]{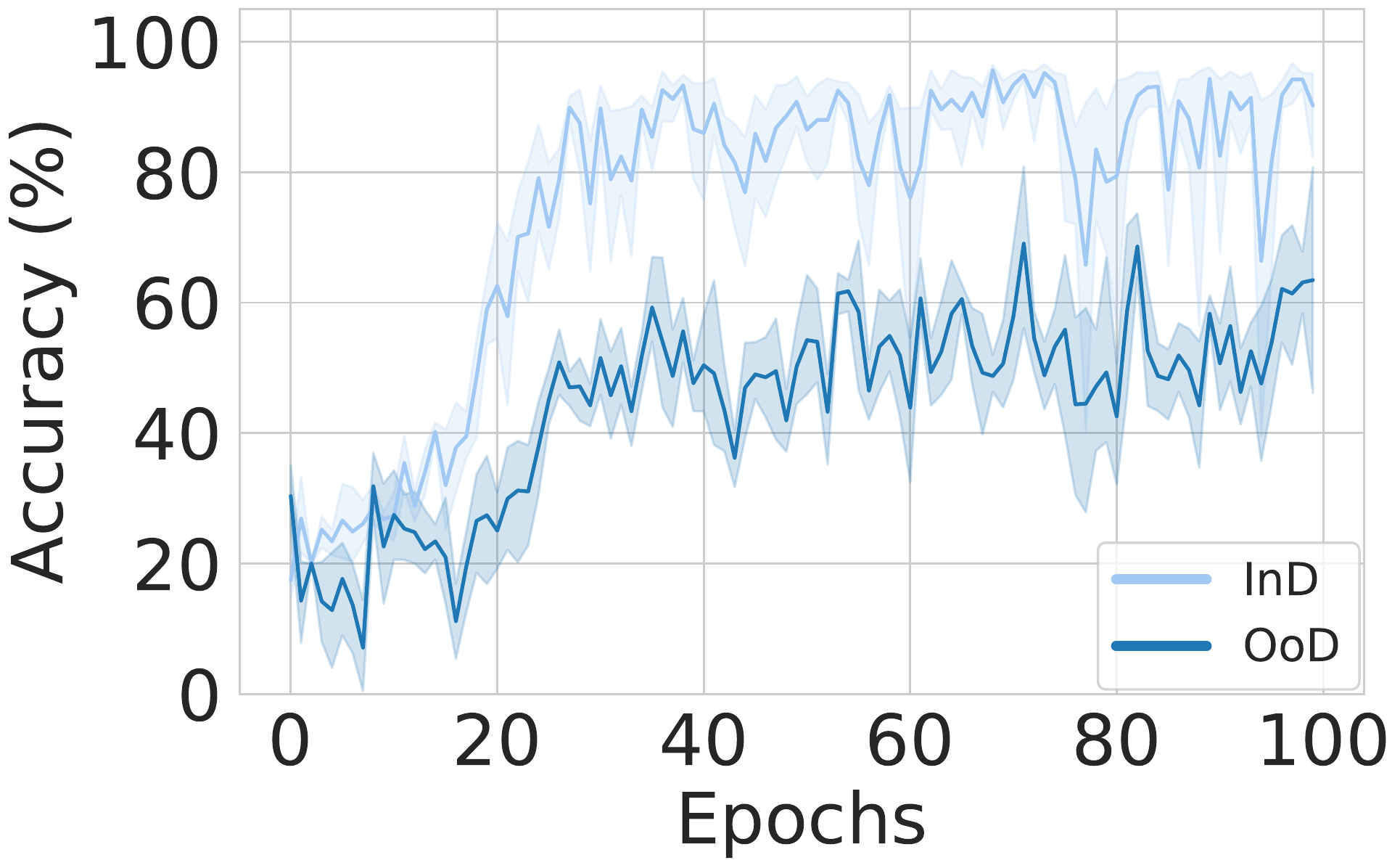}}\\
        \subfigure[BN momentum MNIST]{\label{fig:high_mnist_bnmomentum}
            \includegraphics[width=0.2\columnwidth]{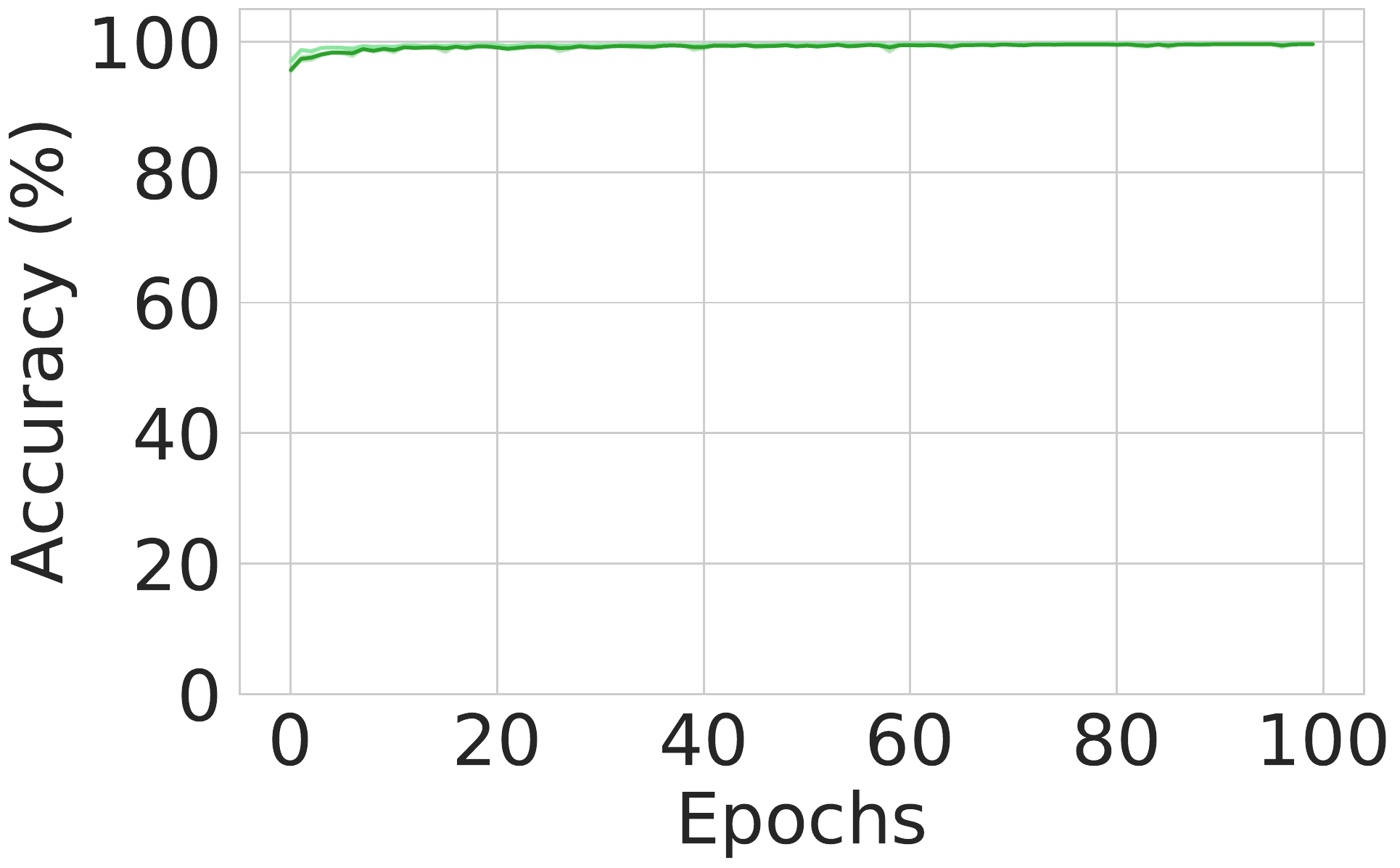}}
        \subfigure[BN momentum iLab]{\label{fig:high_ilab_bnmomentum}
            \includegraphics[width=0.2\columnwidth]{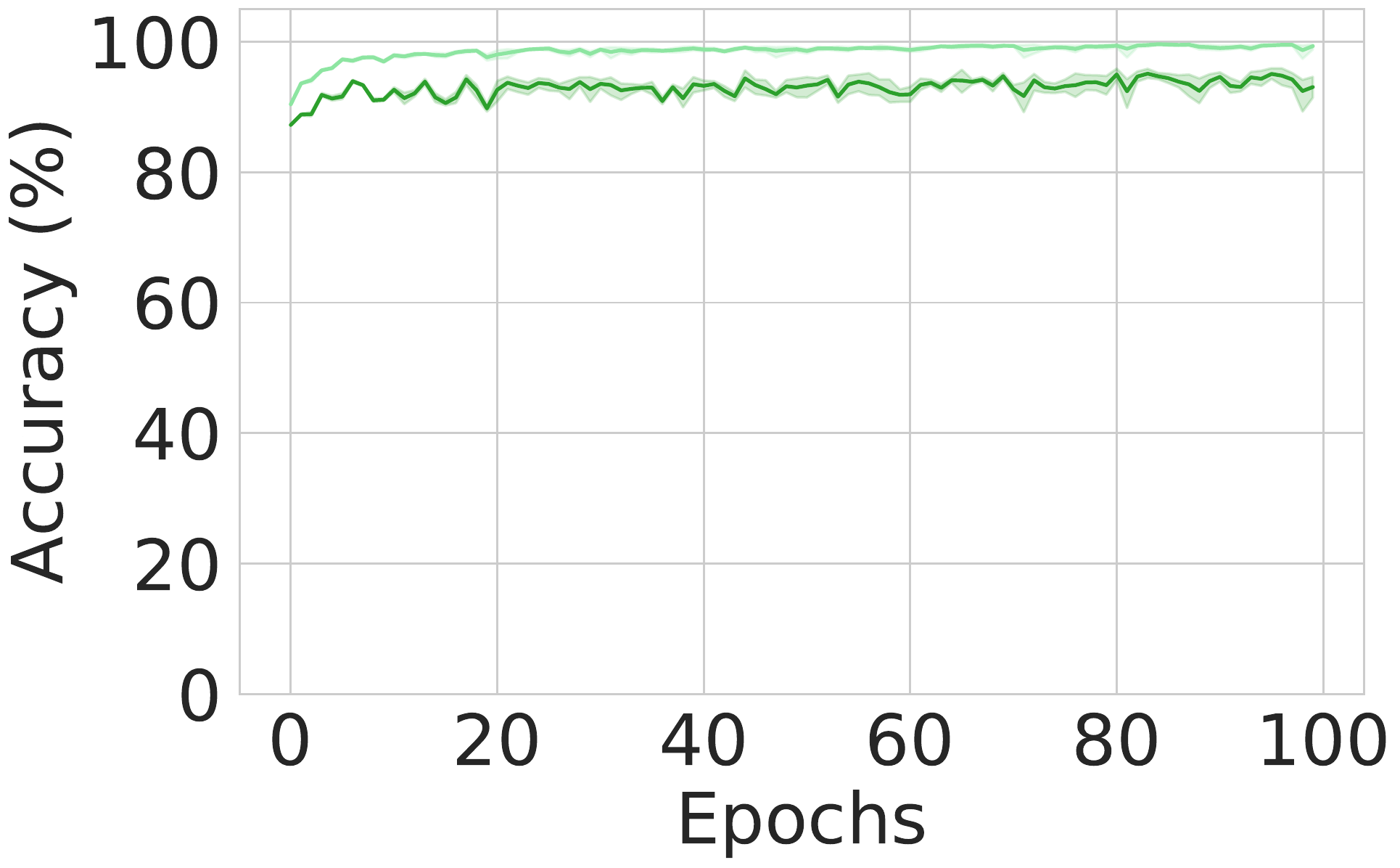}}
        \subfigure[BN momentum CarsCG]{\label{fig:high_carcgs_bnmomentum}
            \includegraphics[width=0.2\columnwidth]{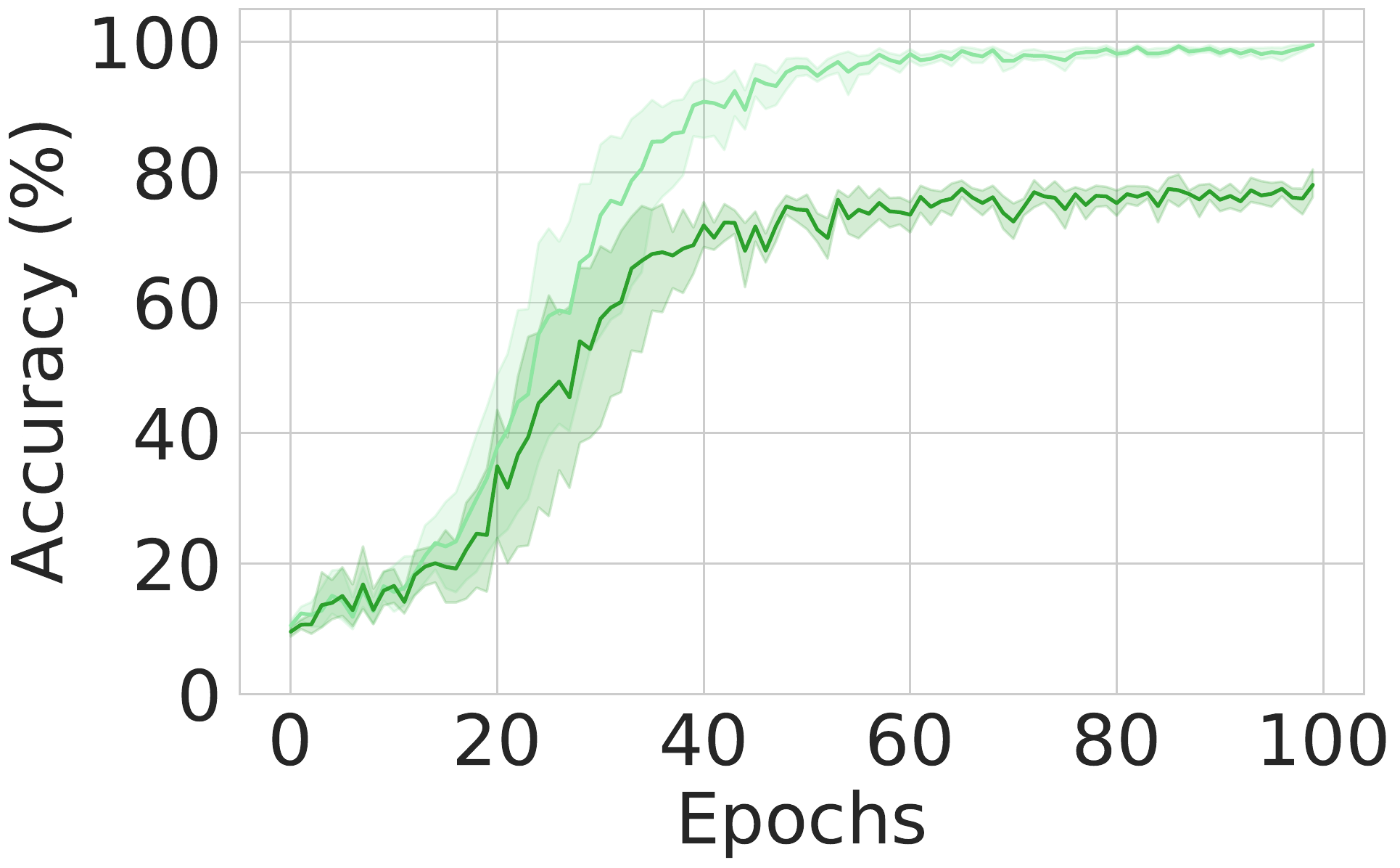}}
        \subfigure[BN momentum MiscGoods]{\label{fig:high_daiso_bnmomentum}
            \includegraphics[width=0.2\columnwidth]{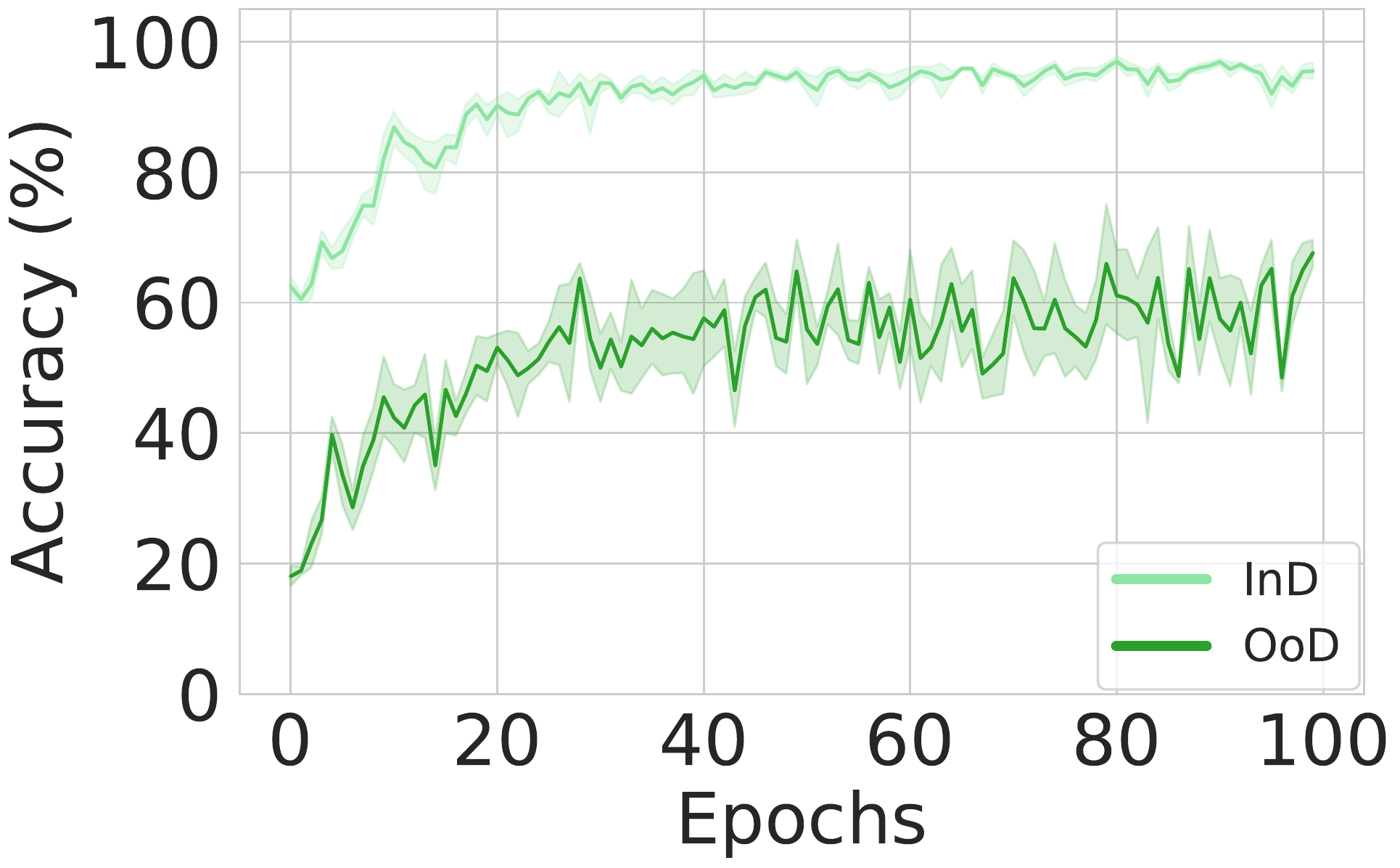}}
    \end{tabular}
\caption{Baseline and BN momentum with high InD data diversity}
\label{fig:high_baseline_bnm_curves}
\end{figure*}

\begin{figure*}[t]
\centering
    \begin{tabular}{ccc}
        \subfigure[Low InD data diversity]{\label{fig:low_seen}
            \includegraphics[width=0.3\columnwidth]{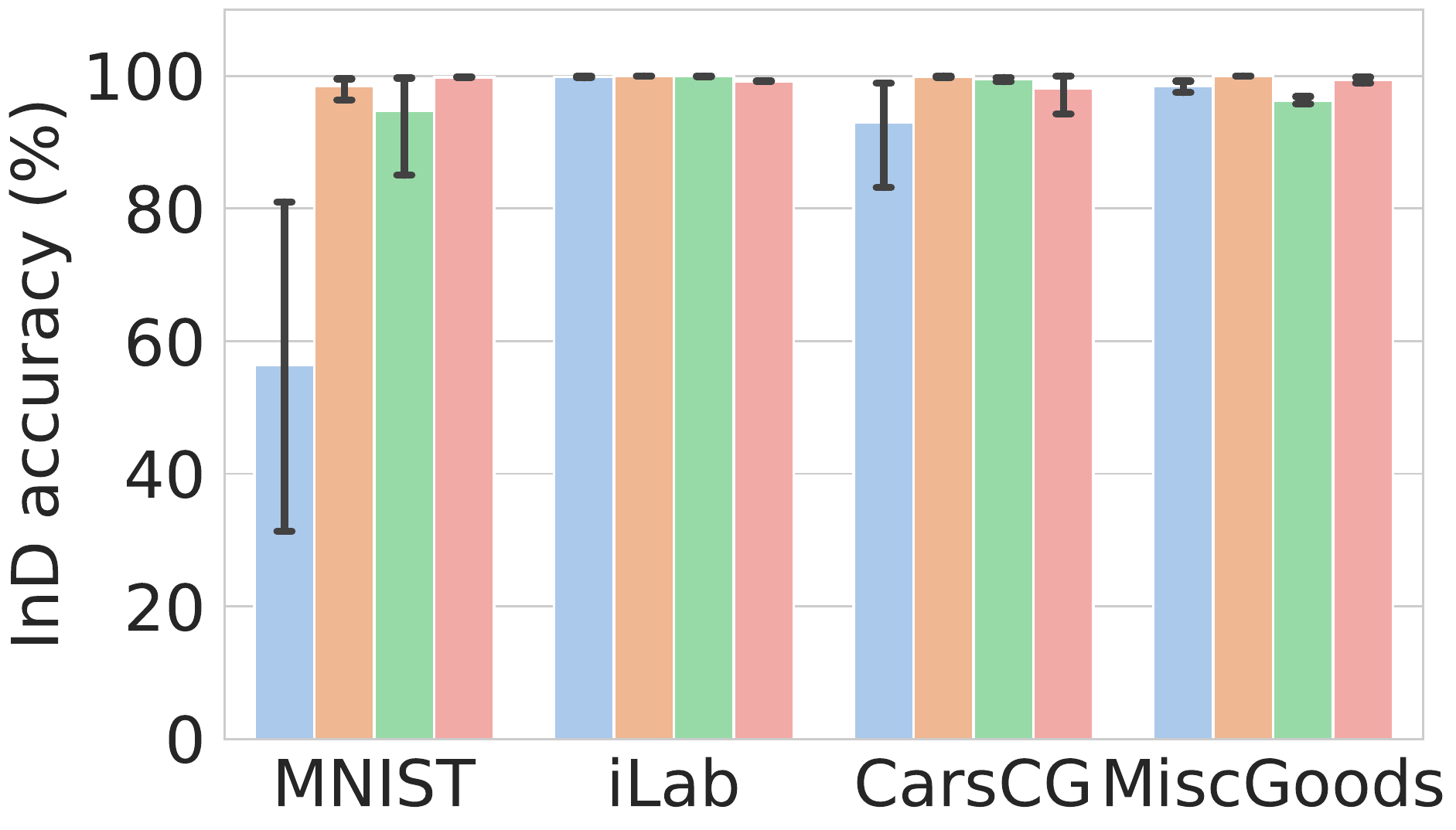}}
        \subfigure[Medium InD data diversity]{\label{fig:medium_seen}
            \includegraphics[width=0.3\columnwidth]{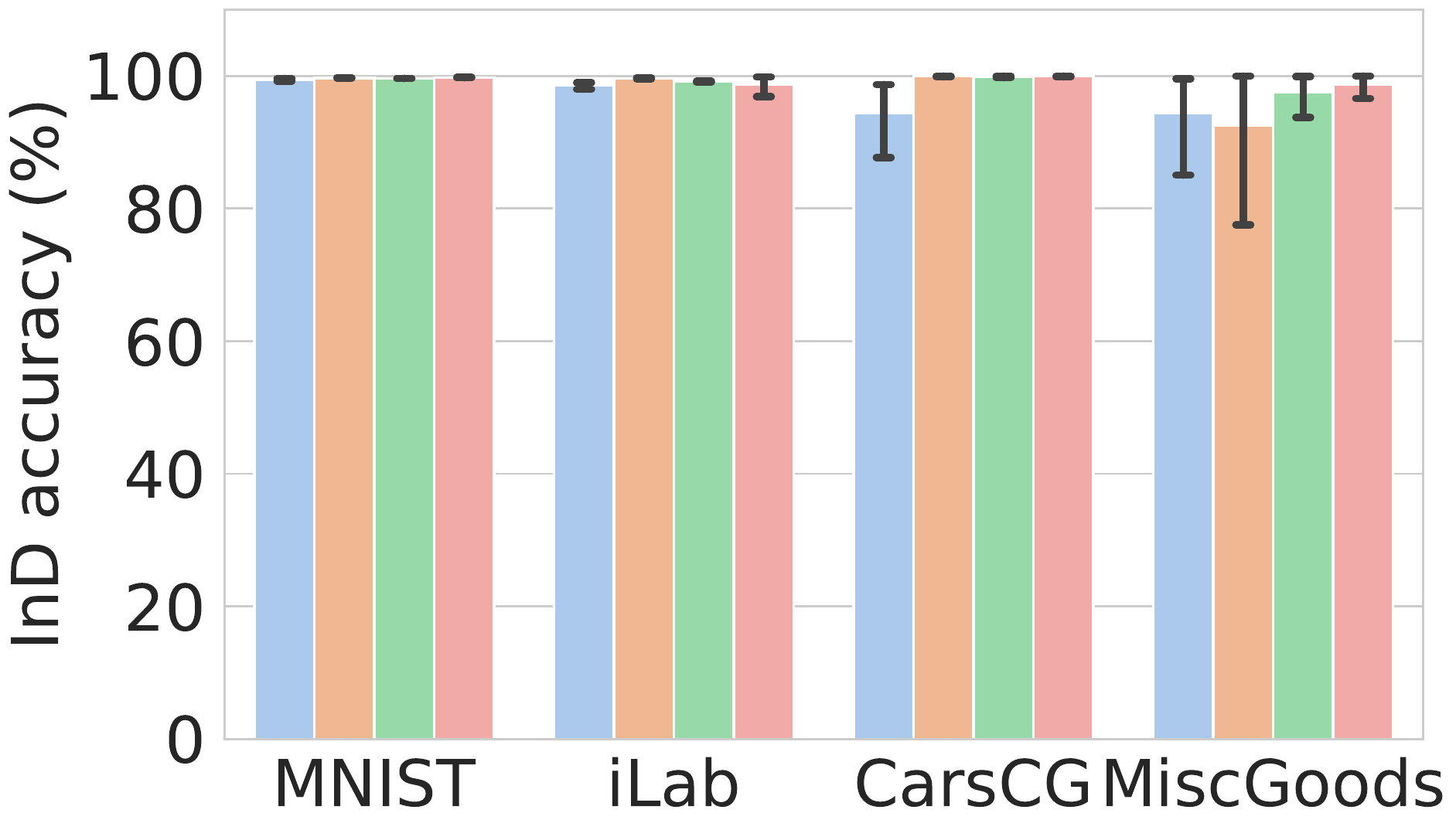}}
        \subfigure[High InD data diversity]{\label{fig:high_seen}
            \includegraphics*[width=0.3\columnwidth]{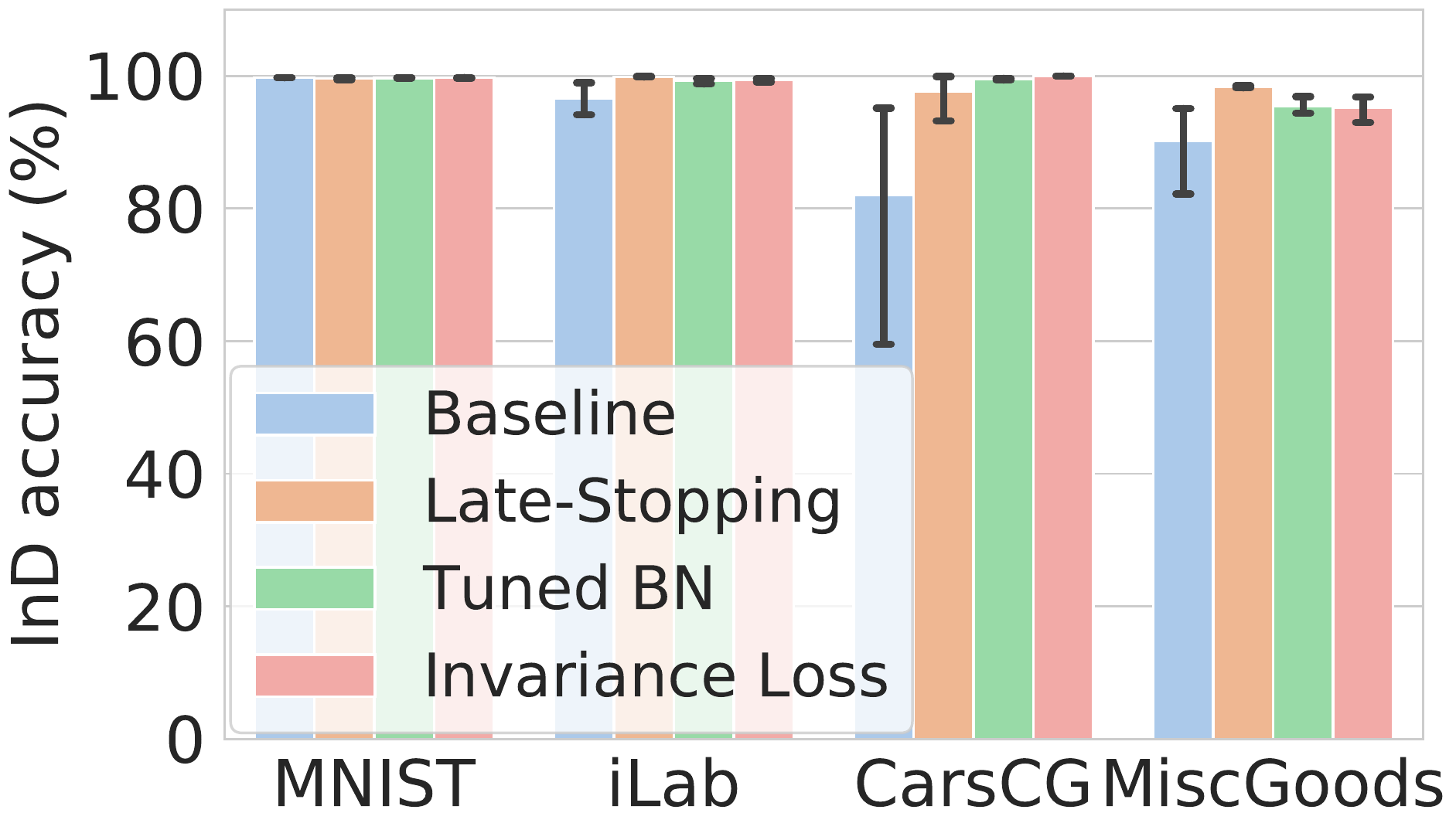}}
    \end{tabular}
\caption{Performance improvement in mean InD accuracy}
\label{fig:seen_compare_performance}
\end{figure*}

\begin{figure*}[t]
\centering
    \begin{tabular}{ccc}
        \subfigure[Low InD data diversity]{\label{fig:low_seen_diff}
            \includegraphics[width=0.3\columnwidth]{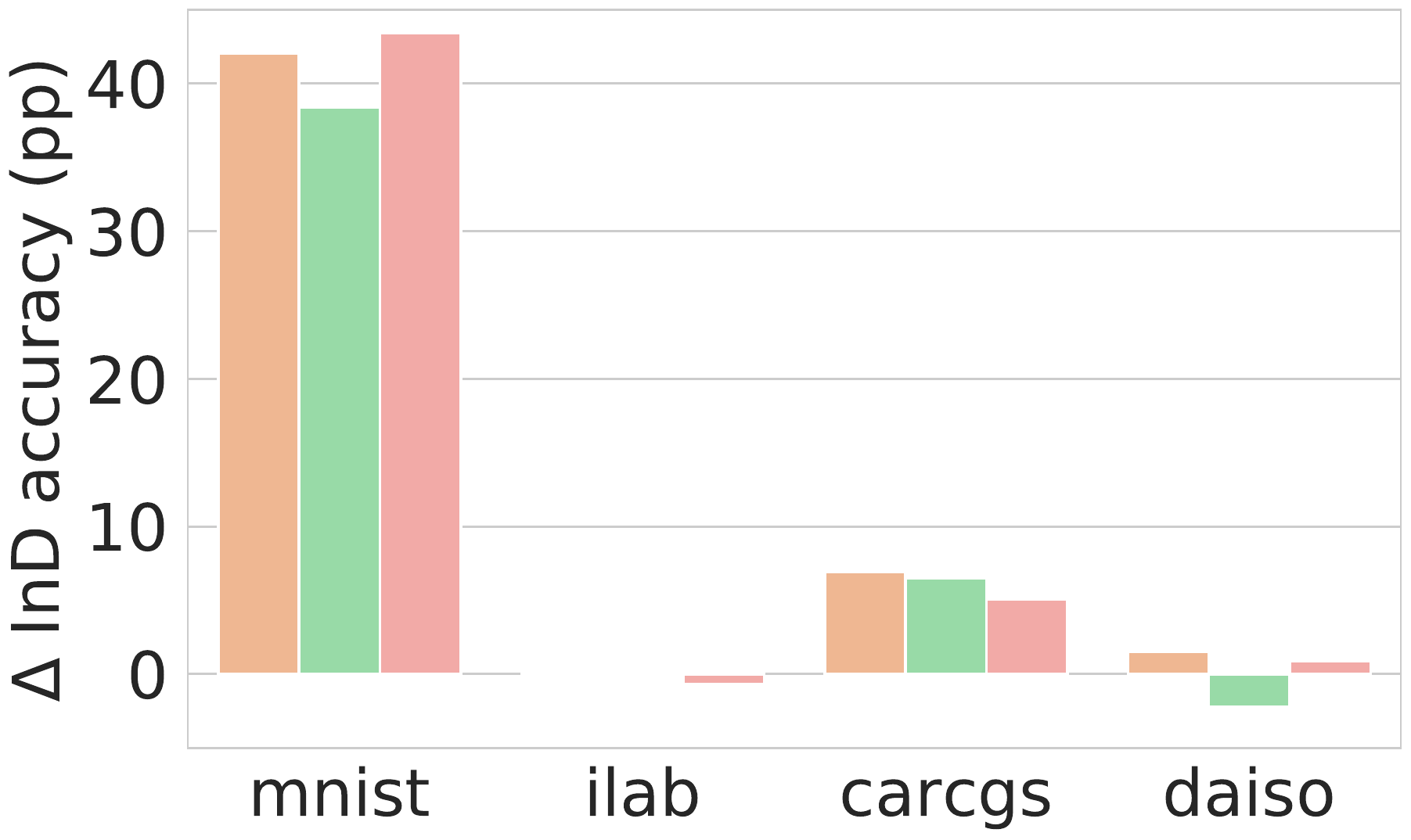}}
        \subfigure[Medium InD data diversity]{\label{fig:medium_seen_diff}
            \includegraphics[width=0.3\columnwidth]{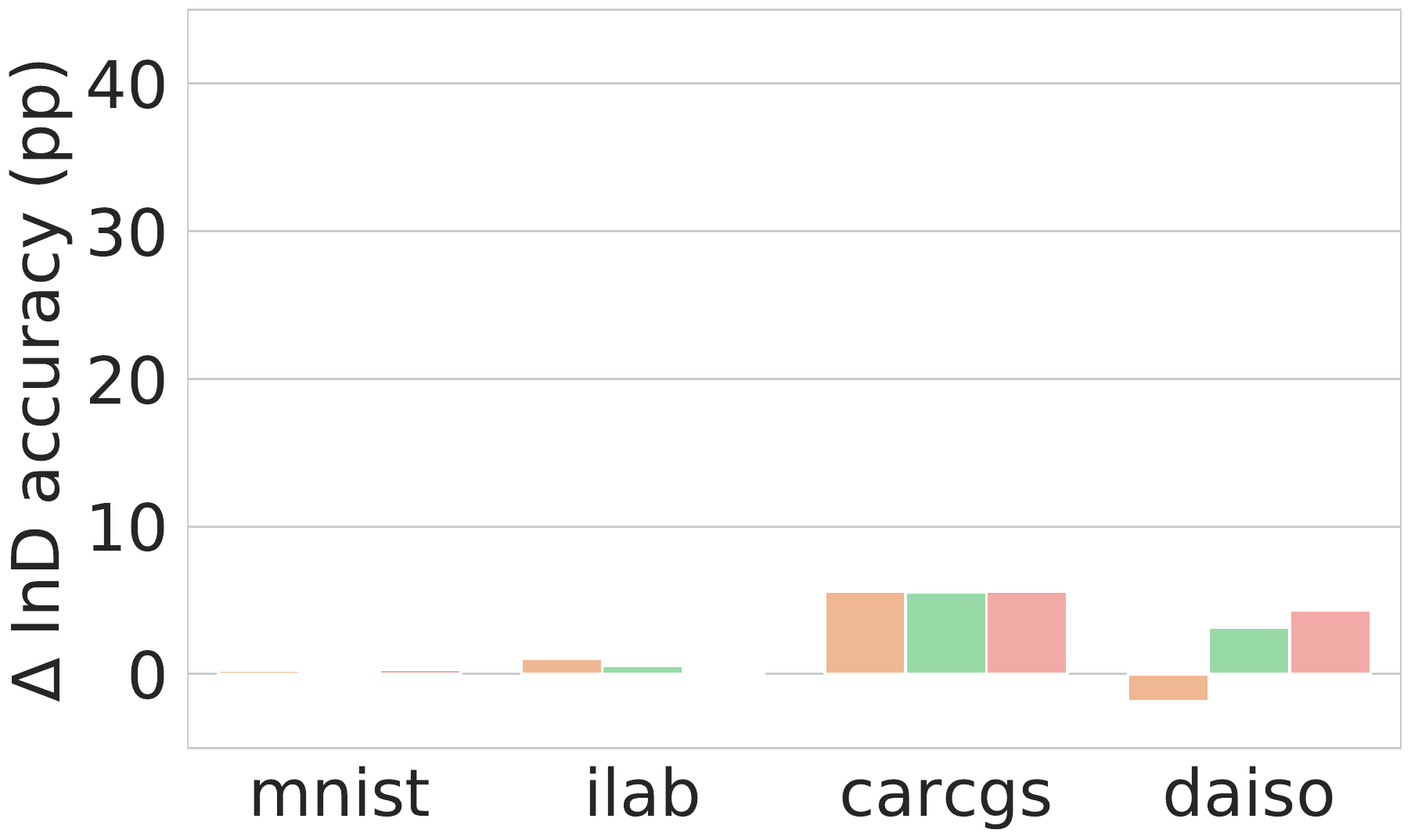}}
        \subfigure[High InD data diversity]{\label{fig:high_seen_diff}
            \includegraphics*[width=0.3\columnwidth]{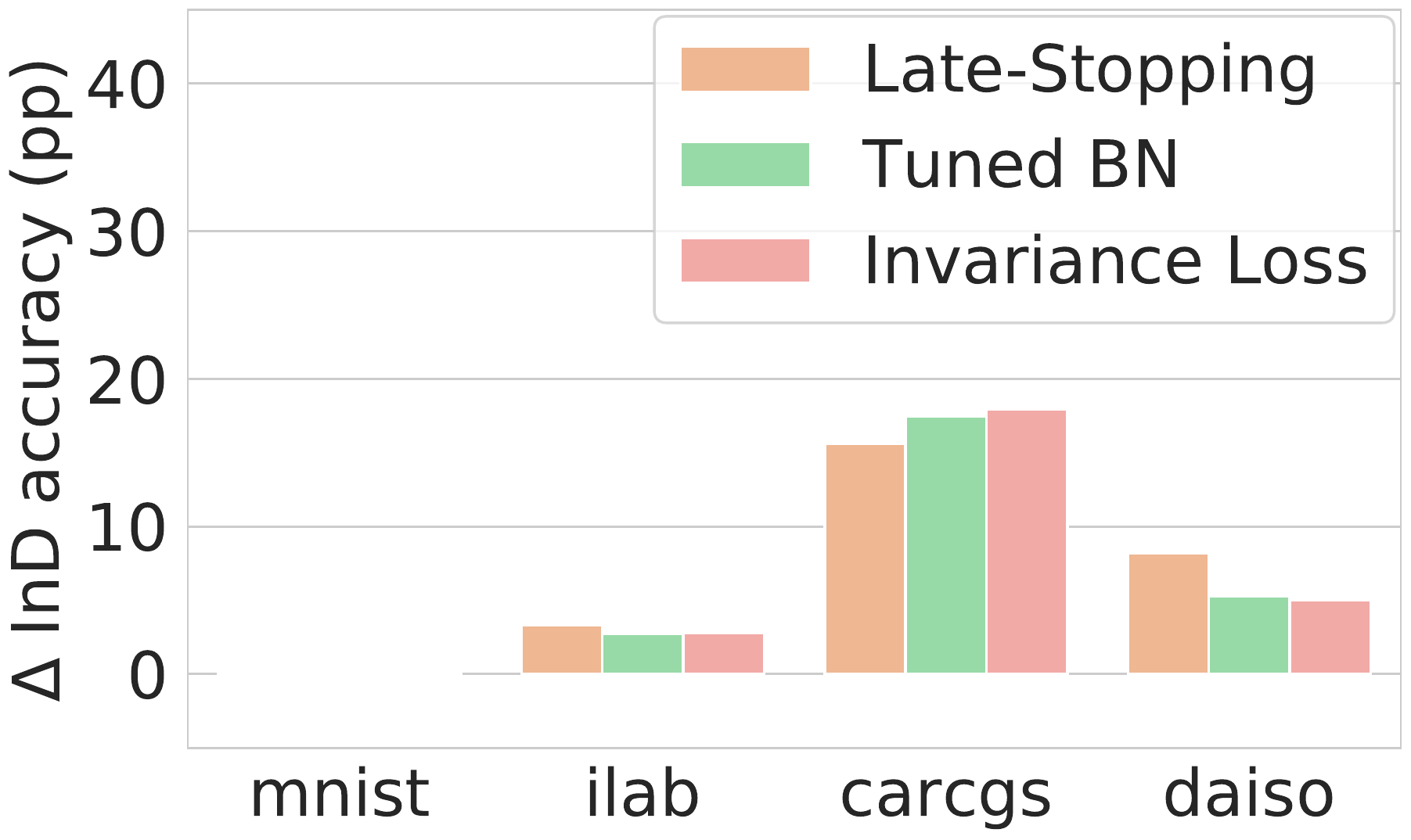}}
    \end{tabular}
\caption{Performance improvement of difference from Baseline in mean InD accuracy}
\label{fig:seen_compare_perform_diff}
\end{figure*}

InD accuracy and OoD accuracy learning curves with all dataset and all diversity corresponding to Fig.~\ref{fig:daiso_late-stopping} are available in Fig.~\ref{fig:low_late_stop_curves}~\ref{fig:medium_late_stop_curves}~\ref{fig:high_late_stop_curves}.
Furthermore InD accuracy and OoD accuracy learning curves with all dataset and all diversity corresponding to Fig.~\ref{fig:carscg_bnmomentum} are available in Fig.~\ref{fig:low_baseline_bnm_curves}~\ref{fig:medium_baseline_bnm_curves}~\ref{fig:high_baseline_bnm_curves}.
InD accuracy corresponding to Fig.~\ref{fig:low_unseen}~\ref{fig:medium_unseen}~\ref{fig:high_unseen} is available in Fig.~\ref{fig:seen_compare_performance}.
InD accuracy of difference from baseline corresponding to Fig.~\ref{fig:low_unseen_diff}~\ref{fig:medium_unseen_diff}~\ref{fig:high_unseen_diff} is also available in Fig.~\ref{fig:seen_compare_perform_diff}
The experiments for measuring accuracy are exactly same as what we reported in the main body of the paper.

\begin{figure*}[ht]
    \begin{tabular}{ccc}
        \subfigure[
$\scriptstyle \mbox{Low InD data diversity}$]{
            \includegraphics[width=0.31\textwidth]{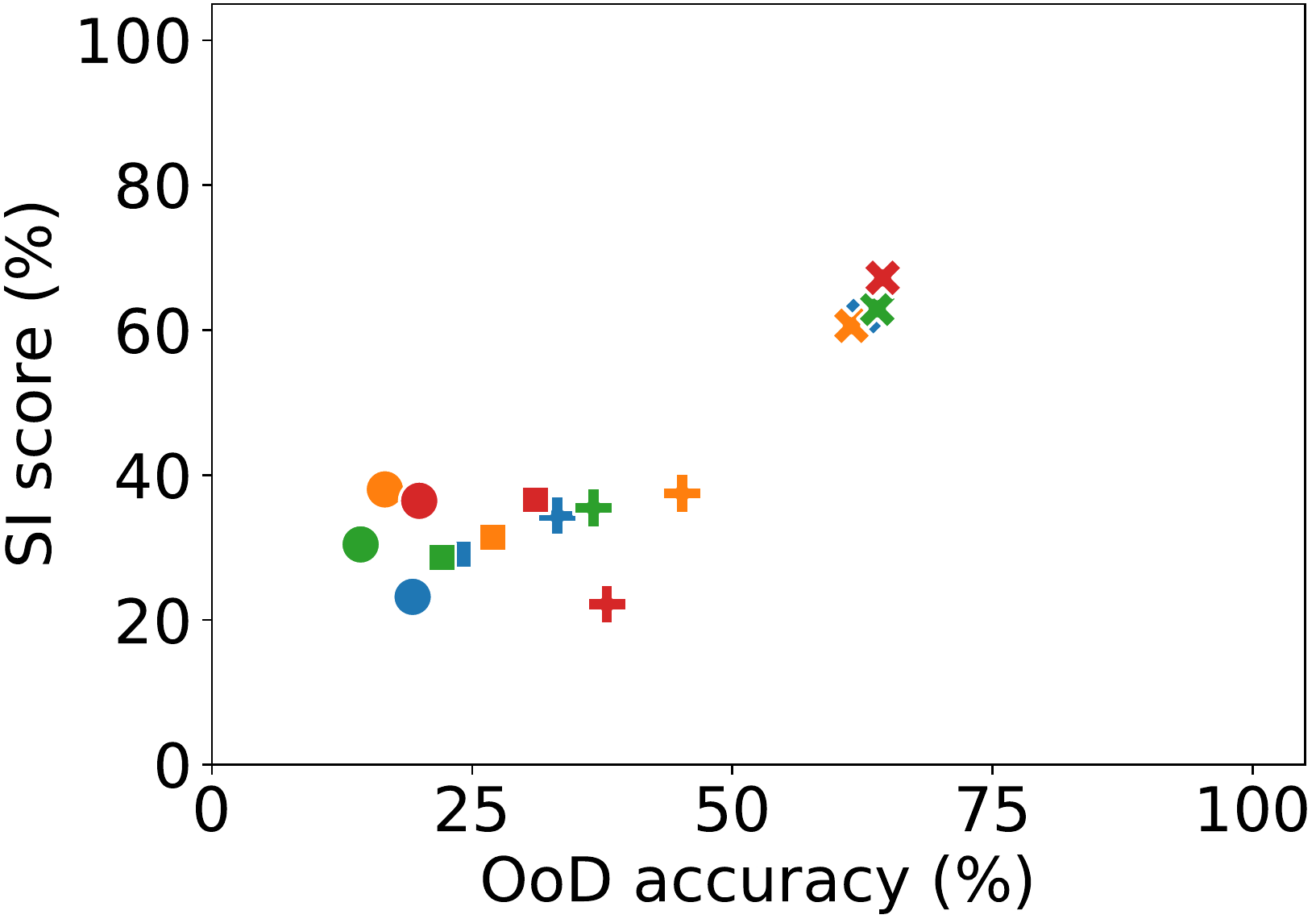}}
        \subfigure[$\scriptstyle \mbox{Medium InD data diversity}$]{
            \includegraphics[width=0.31\textwidth]{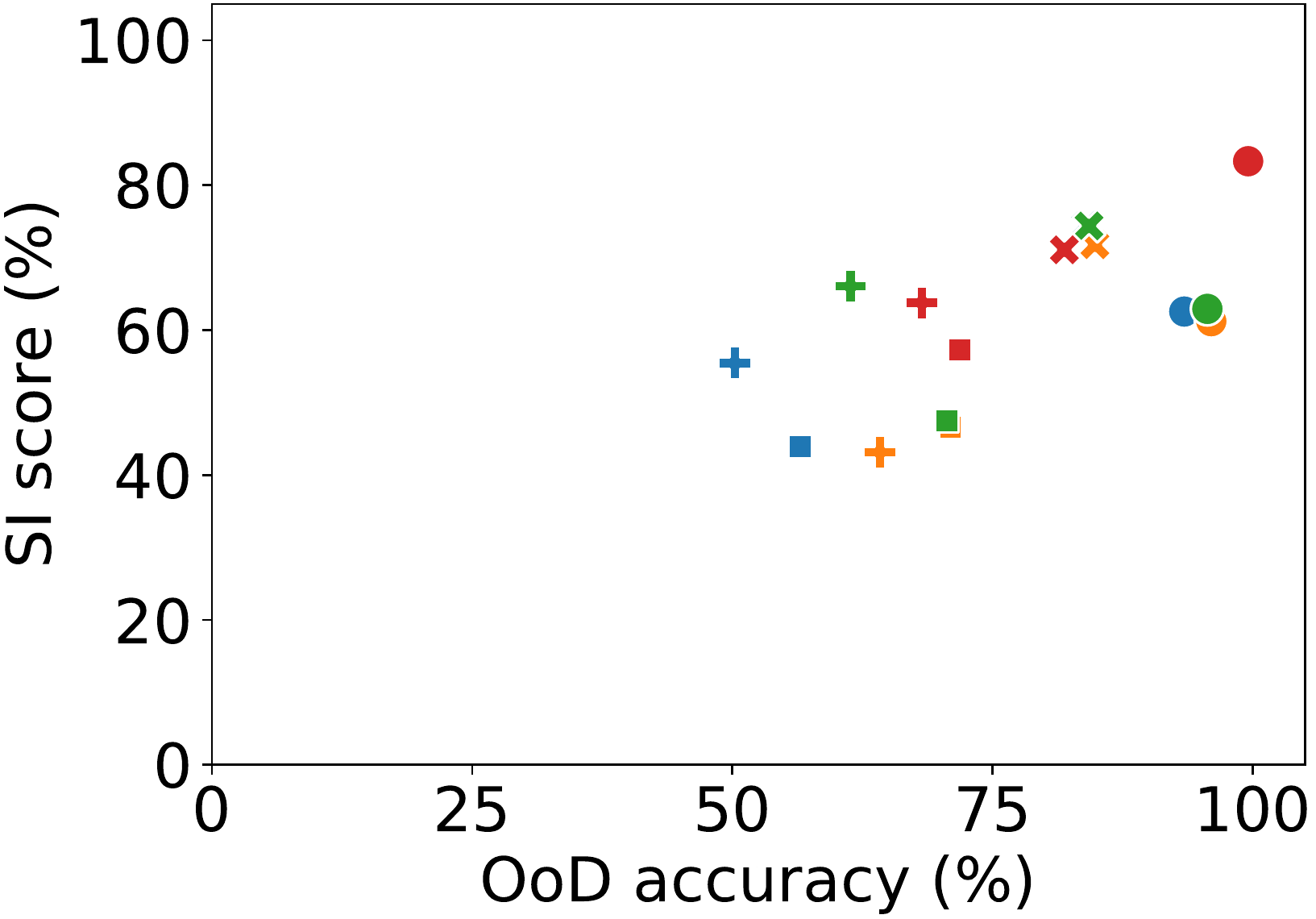}}
        \subfigure[$\scriptstyle \mbox{High InD data diversity}$]{
            \includegraphics*[width=0.31\textwidth]{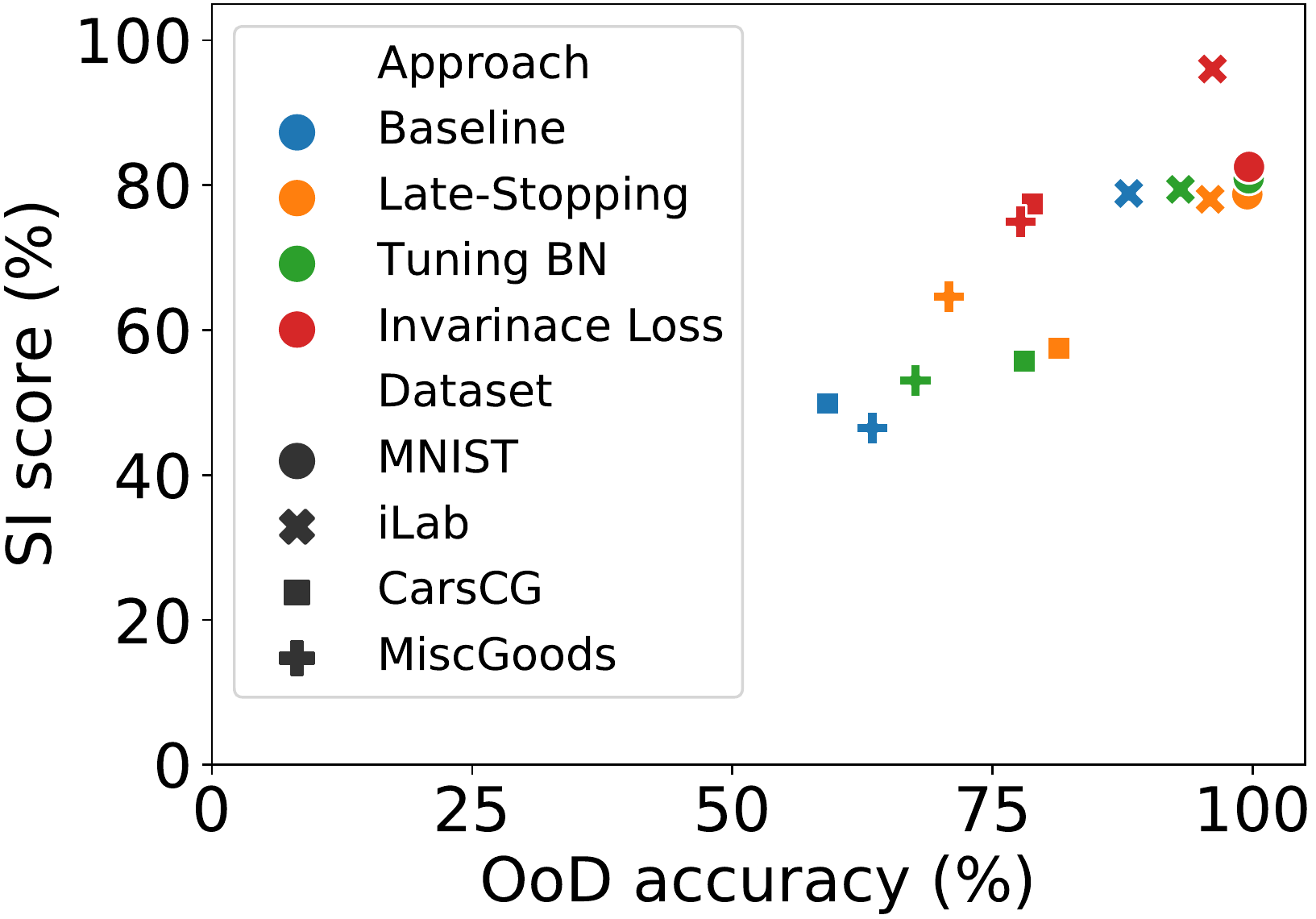}} \\
    \end{tabular}
\caption{\emph{Overall results between IS scores and all combinations of (data diversity, dataset, approaches)}}
\label{fig:compare_col_analysis}
\end{figure*}

\setcounter{figure}{0}
\setcounter{table}{0}
\setcounter{equation}{0}

\section{Details of combined method}
\label{app:combined}
For the best of three approaches, we choose the one with the highest OoD accuracy calculated on the data set for hyperparameter tuning among late stopping, tuning batch normalization momentum, and invariance loss.
The combined three approaches employ the hyper-parameters that determined in the Section~\ref{SecExperiments}: epoch size for longer epochs, momentum parameter $\beta$ for tuning batch normalization momentum, and learning rate, paring interval and the value $\lambda$ for invariance loss. If the InD accuracy drops to a chance, the learning rate is multiplied  by $0.1$ and the training is re-started. 
\section{Abbreviation list}
\begin{itemize}
    \item \ood: out-of-distribution
    \item \id: in-distribution
    \item DNN: deep neural network
    \item BN: batch normalization
    \item MNIST dataset: modified national institute of standards and technology dataset
    \item CarsCG dataset: cars computer graphics dataset
    \item  MiscGoods dataset: miscellaneous goods dataset
    \item 3D: three dimension 
    \item EIIL: environment inference for invariant learning 
    \item SI score: selectivity and invariance score
    \item ResNet: residual network
    \item ReLU: rectified linear unit
    \item CC: creative commons
\end{itemize} 
\end{document}